\documentclass[aic,crcready]{iosart2x}

\usepackage{dcolumn}

\newcolumntype{d}[1]{D{.}{.}{#1}}

\usepackage{lmodern} 
\usepackage[utf8]{inputenc} 
\usepackage[T1]{fontenc}    
\usepackage{booktabs}       
\usepackage{amsfonts}       
\usepackage{nicefrac}       
\usepackage{microtype}      
\usepackage{lipsum}		
\usepackage{graphicx}
\usepackage{natbib}
\usepackage{bbm}

\usepackage{amsthm}
\newtheorem{definition}{Definition}
\newtheorem{corollary}{Corollary}
\newtheorem{lemma}{Lemma}

\newtheorem{example}{Example}
\usepackage{mathrsfs}
\usepackage{etoolbox}
\usepackage{tikz}
\usepackage{tkz-euclide}
\usepackage{color}
\usepackage{pgfgantt}
\usepackage{siunitx}
\usetikzlibrary{shapes,arrows}
\usetikzlibrary{arrows,automata}
\usetikzlibrary{positioning}
\usetikzlibrary{automata,decorations.pathreplacing,shapes}
\usepackage{pgfplots}
\pgfplotsset{compat=newest}
\usetikzlibrary{calc}
\usepackage{ifthen}
\usepackage[space]{grffile}
\usetikzlibrary{arrows,decorations.markings}
\usepackage{amsmath}
\usepackage{lineno}
\usepackage{tikz}
\usepackage{algorithm}
\usepackage[noend]{algpseudocode}
\algnewcommand{\LineComment}[1]{\State \(\triangleright\) #1}
\usepackage{color}
\usetikzlibrary{shapes,arrows}
\usetikzlibrary{arrows,automata}
\usetikzlibrary{positioning}
\usepackage{caption}
\usepackage{subfigure}
\usepackage{amsmath}
\DeclareMathOperator*{\argmax}{arg\,max}

\usepackage{pgfplots}
\pgfplotsset{compat=newest}
\usetikzlibrary{calc}
\usetikzlibrary{decorations.pathreplacing,shapes}
\usepackage{ifthen}
\usepackage{bm}
\usepackage{amsfonts}
\usepackage{stackengine}
\usepackage{fancyheadings}

\usetikzlibrary{arrows,decorations.markings}
\usepackage[british]{babel}
\usepackage{hyphenat}

\newcommand{\ifequals}[3]{\ifthenelse{\equal{#1}{#2}}{#3}{}}
\pubyear{0000}
\volume{0}
\firstpage{1}
\lastpage{1}
\usepackage{fancyhdr}

\begin{document}

\begin{frontmatter}

\pretitle{Accepted for publication at AI Communications}
\title{Lifetime policy reuse and the importance of task capacity}
\runtitle{Lifetime policy reuse and the importance of task capacity}

\begin{aug}
\author[A]{\inits{D.M.}\fnms{David M.} \snm{Bossens}\ead[label=e1]{bossensdm@cfar.a-star.edu.sg}%
\thanks{Corresponding author. \printead{e1}. Current address: Centre for Frontier AI Research, A*STAR, 1 Fusionopolis Way, \#16--16 Connexis (North Tower), Singapore 138632}}
\author[A,B]{\inits{A.J.}\fnms{Adam J.} \snm{Sobey}\ead[label=e2]{ajs502@soton.ac.uk}}
\address[A]{Maritime Engineering group, \orgname{University of Southampton},\cny{United Kingdom}}
\address[B]{Marine and Maritime Group, \orgname{The Alan Turing Institute},\cny{United Kingdom}\printead[presep={\\}]{e1,e2}}
\end{aug}

\begin{abstract}
A long-standing challenge in artificial intelligence is lifelong reinforcement learning, where learners are given many tasks in sequence and must transfer knowledge between tasks while avoiding catastrophic forgetting. Policy reuse and other multi-policy reinforcement learning techniques can learn multiple tasks but may generate many policies. This paper presents two novel contributions, namely 1) Lifetime Policy Reuse, a model-agnostic policy reuse algorithm that avoids generating many policies by optimising a fixed number of near-optimal policies through a combination of policy optimisation and adaptive policy selection; and 2) the task capacity,  a measure for the maximal number of tasks that a policy can accurately solve. Comparing two state-of-the-art base-learners, the results demonstrate the importance of Lifetime Policy Reuse and task capacity based pre-selection on an 18-task partially observable Pacman domain and a Cartpole domain of up to 125 tasks.
\end{abstract}


\end{frontmatter}

\section{The importance of efficient learning over multiple tasks}
During their lifetime, animals may be subjected to a large number of unknown tasks. In some environmental conditions, nutritious food sources may be readily available, while in others they may be sparse, hidden, or even poisonous, and dangerous predators may roam in their vicinity. To address these challenging conditions, various behaviours must be selectively combined, such as avoidance, reward-seeking, or even fleeing.  When direct perception provides limited or no cues about the current task, animals have to infer the task or use a strategy that works for many different tasks it may encounter. If task sequences are long and diverse, the animal may need to find  different strategies, each of which apply to a large sub-domain of the tasks it encounters. Selecting how many such strategies are required represents a trade-off between optimality and the animal's limited cognitive resources.

In artificial intelligence, variants of the above problem have been formulated within the field of lifelong learning \cite{Thrun1995,Silver2013,Chen2016}, in which new tasks may appear at any time and where a learned task might be forgotten when it has not been seen regularly. The lifelong learning setting combines the problems of transfer learning and catastrophic forgetting into a single scenario. In transfer learning, learners should leverage the knowledge gained from a set of previously learned tasks with similar characteristics to a new task, whilst avoiding transferring knowledge that is not relevant \cite{Taylor2009,Pan2010a,Lazaric2013b}. In catastrophic forgetting, knowledge learned on one task removes or deteriorates knowledge learned on some previous tasks \cite{French1992,Hasselmo2017}. Due to combining both necessary aspects of learning as well as the emphasis on long-term problem solving, the lifelong learning setting is an exciting frontier for advancing modern artificial intelligence systems. 

Lifelong reinforcement learning represents the intersection of reinforcement learning (RL) \cite{Sutton2017} and lifelong learning in the above sense. Lifelong RL requires representing the commonalities among tasks which may be done by considering a single policy with task-specific features or through multiple policies. Single policy learning emphasises feature sharing which is beneficial when it is known which aspects are task-specific and which are shared. Multi-policy learning has improved applicability to larger task sets with more widely varying tasks and it allows greater task specialisation within the task set. A key challenge is that to store solutions to different tasks, multi-policy RL (e.g. \cite{Finn2017,Hernandez-Leal2016}) generates more policies as more tasks are presented, which results in excessive memory consumption. Moreover, in both single- and multi-policy RL, there is a fundamental challenge regarding the \textit{task capacity} of policy representations: how many tasks can be represented or learned within a policy without a significant loss of performance? So far, this question has only been investigated indirectly by tracking forgetting and transfer metrics \cite{Taylor2009,Kemker2018} or by growing a policy library until there is no more benefit of adding policies (e.g. \cite{Fernandez2006}). 

To address both challenges, this paper focuses on two novel contributions:
\begin{itemize}
\item Lifetime Policy Reuse: a policy reuse algorithm which maintains a fixed-size policy library. The algorithm adaptively assigns policies to tasks based on bandit learning over the lifetime reinforcement on the different tasks. Contrasting to previous approaches, the algorithm is model-agnostic in the sense that it can take any base-learner as the underlying policy optimisation and exploration on the single-task level is preserved as is.
\item Measures for task capacity: the paper presents theoretical and empirical measures for task capacity, which can be used to determine how many policies to use in a lifelong RL problem. The theoretical task capacity applies arguments from representational capacity  \cite{Baldi2019,Folli2017,Friedland2017,Hornik1989,SCHAFER2007,VC1971} to the multi-task multi-policy context. To account for sequential effects and learning, the empirical task capacity is based on the relative performance compared to task-specific policies.
\end{itemize}
Using lifetime policy reuse with pre-selection based on task capacity allows high performance in lifelong learning domains, as demonstrated in a Cartpole domain and a partially observable Pacman domain, both of which consist of a long sequence of randomly ordered tasks.

The paper is structured as follows. Section \ref{sec: prelim} provides preliminaries to the lifelong RL task sequences. Section \ref{sec: LRL} provides an overview of lifelong reinforcement learning algorithms and their relation to Lifetime Policy Reuse. Section \ref{sec: LPR} develops the three lifetime principles behind Lifetime Policy Reuse. Section \ref{sec: taskcapacity} develops useful definitions of task capacity. Section \ref{sec: setup} provides the experimental setup including lifelong learning domains and the various experimental conditions. Section \ref{sec: performance} provides the performance evaluation of Lifetime Policy Reuse and analyses the impact of adaptive policy selection, the importance of task capacity, and the difference between the base-learners.

\section{Reinforcement learning setup}
\label{sec: prelim}
In RL, an agent interfaces with the environment in the following control cycle. First, it receives the state of the environment, a state $s$ from the state space $\mathcal{S}$ that directly corresponds to a unique environment state, or a more limited observation, an observation $o$ from the observation space $\mathcal{O}$ that is only a partial observation of the environment state. Then, based on the full state, if it is observable, or else the more limited observation,   it performs an action $a$ according to its policy $\pi: \mathcal{S} \to \mathcal{A}$ where $\mathcal{A}$ is the space of actions; and finally, it receives a real-valued reward $r$ from the environment. The policy can be formulated explicitly or implicitly; for example, an often-used implicit approach is to define an action-value function $Q$, which represents the utility $Q(s,a)$ of any state-action pair $\langle s, a \rangle$, and then select the optimal action for the given state probabilistically according to the action-value function.  To adaptively increase the reward intake over time, the learning agent will perform periodic updates to its value-function or directly to its policy. The policy and the value-function are often fitted through a function approximator, such as a deep neural network, to be able to solve larger state-action spaces.

The most common task-modelling frameworks in RL are Markov Decision Processes (MDPs) and Partially Observable Markov Decision Processes (POMDPs). 

An MDP is defined by a tuple $\langle \mathcal{S}, \mathcal{A}, \mathcal{T}, r, \gamma  \rangle$, where: $\mathcal{S}$ is the state space; $\mathcal{A}$ is the action space; $\mathcal{T}$ is the transition function that samples the next state $s_{t+1}$ conditioned on the current state $s_t$ and the current action $a_t$, i.e. $s_{t+1} \sim \mathcal{T}(s_t,a_t)$;  $r_t = r(s,a)$ denotes the real-valued reward at time $t$ after performing action $a_t$ in state $s_t$; and $\gamma \in [0,1]$ denotes the discount parameter that represents the patience in the value function
\begin{equation}
V(s_t) = \mathbb{E}_{\pi, \mathcal{T}}\left[\sum_{j=t}^{t+H} \gamma^{j-t} r(s_j,a_j)\right] \,,
\end{equation}
where $H$ is the horizon. An action-specific $Q$-function is often formulated as $V(s_t) = \max_{a \in \mathcal{A}} Q(s_t,a_t)$. Due to the transition function being state-action dependent, the MDP follows the Markov property such that $\mathbb{P}(s_{t+1} | s_{t}a_{t}) = \mathbb{P}(s_{t+1} | s_{t}a_{t} \dots s_{1}a_{1})$.

The POMDP generalises the MDP to account for partial observability of the current environment state, making it more widely applicable. The POMDP $\langle \mathcal{S}, \mathcal{A}, \mathcal{T}, r, \gamma  , \mathcal{O}, \mathcal{T}_{o} \rangle$ assumes an underlying MDP but additionally incorporates an observation space $\mathcal{O}$ and an observation transition function $\mathcal{T}_{o}$, such that $o_{t+1} \sim  \mathcal{T}_{o}(s_{t+1},a_t)$ is the observation following action $a_t \in \mathcal{A}$ in the reached state $s_{t+1} \in \mathcal{S}$. In POMDPs, the observations are input to the learning agent, rather than the states, and the Markov property does not hold for the observations. That is, it is not required that $\mathbb{P}(o_{t+1} | o_{t} a_{t} \dots o_1 a_t) = \mathbb{P}(o_{t+1} | o_{t} a_{t})$. However, the states of the underlying MDP still follow the Markov property. The MDP can therefore be seen as a special-case POMDP in which observations and states are always equal such that $\mathcal{O} = \mathcal{S}$ and $\mathcal{T}_o = \mathcal{T}$.

While traditional RL solves a single MDP or POMDP, this paper will focus on how to solve lengthy sequences of randomly presented MDPs and POMDPs.

\section{Lifelong reinforcement learning: a review}
\label{sec: LRL}
Lifelong reinforcement learning (LRL) is one of the branches of lifelong learning which focuses on the problem setting of RL. This section surveys the field, identifying scalability to a large number of widely varying tasks and applicability to different base-learners as key gaps in the current state-of-the-art.

\subsection{Single-policy methods}
Most LRL approaches consider how to learn with a single policy and provide some task-specific aspects. The main aim of this approach is that it allows transferable features to be efficiently learned within a single representation while adding some additional task-specificity. Specificity may be represented in architectural components or within the training algorithm itself to improve performance on multiple tasks  \citep{Kirkpatrick2017,Jung2017,Li2018,Rusu2016,Isele2018,Schaula,Deisenroth2014}. While a single policy is compact and the neural network's hidden layers close to the input layer are able to represent transferable features, it can bring disadvantages. The number of neurons may grow with the number of tasks \citep{Rusu2016}, or expert knowledge of the task's relevant features may be needed as inputs to the learning system \citep{Schaula,Deisenroth2014}. Further, there may be no common representation to all tasks; for example, one lifelong learning approach, which reused a single Deep Q-Network policy on all of the tasks, has also been outperformed by task-specific DQN policies \citep{Mnih2015,Kirkpatrick2017}. More generally, these approaches typically have a few limiting assumptions on the relation between tasks. For example, \cite{Cheung2020} and \cite{Lecarpentier2019} require some bounds on the change in reward function and transition function, implying it is more suited to continuously changing scenarios rather than a sequence of tasks from a wide task domain. Techniques and insights relevant for single-policy RL may also be found in related fields; for example, parameter sharing for multi-agent RL  \cite{Terry2020,Gupta2017}, which uses a single-network to represent all policies of the agents but which may add agent indicators to the observation to account for heterogeneity of different agents.
\subsection{Policy reuse}
To avoid some of the above-mentioned challenges, some have recognised the need for learning with multiple policies. In particular, policy reuse methods learn a limited number of policies each specialised to their own subset of tasks as a means to improve transfer to a new unseen task \citep{Wang2018a,Li2018a,Li2018b,Rosman2016,Fernandez2006}. The approach represents a step forward in improving the scalability of multiple policy approaches as some empirical results have shown the viability of transfer learning. For example, 50 tasks in an office domain were learned with 14 Q-learning policies; the results show this was possible by learning a common policy for all of the tasks which have their goal locations in the same room \citep{Fernandez2006,Watkins1992}. While there have been recent advances in policy reuse which demonstrate improved transfer on a smaller number of tasks (e.g. \cite{Li2018a}), the high number of 50 tasks in \cite{Fernandez2006} has not been replicated in other LRL studies (policy reuse or otherwise). Of primary interest here, the high number of tasks indicates policy reuse as a promising approach for scalable LRL.

To scale the policy reuse approach to memory-consuming deep reinforcement learners or to even more policies than is currently the case, current methods have their limitations. In particular,  online learning of the policy library without generating too many temporary or permanent policies remains a challenge in policy reuse. Most works have focused on how to select the new policy for a new, unlabeled task (see e.g. \cite{Rosman2016}). Other works have provided tools to build the policy library. In \cite{Fernandez2006}, new, temporary policies are formed for new tasks and these may be added to the library if after convergence on the task the resulting policy performs significantly better than the existing policies in the library. A downside of this method is that when convergence is not allowed, and many tasks are randomly presented, memory issues may become apparent. In \cite{Hernandez-Leal2016}, one also forms new policies for new tasks but all policies are added to the library, which implies memory issues will be apparent even more rapidly. 

Policy reuse has been limited in its applicability as base-learners other than Q-learning methods have yet to be incorporated with policy reuse. This would be particularly desirable after results suggesting Q-learning may not be optimal for reuse, since (i) policy reuse with Q-learning policies may result in negative transfer, being outperformed by task-specific Q-learning \citep{Li2018a}; and (ii) a single-policy lifelong learning approach with DQN has been outperformed by task-specific DQN policies \citep{Mnih2015,Kirkpatrick2017}.

\subsection{Alternative multi-policy methods}
A variety of alternative approaches have been taken to multi-policy LRL. These methods assume some similarity across all tasks and are therefore not the method of choice to learn large sets of tasks with widely varying dynamics and reward functions.

One approach is to provide an initial policy based on the prior tasks and then generate a new policy for each additional task encountered during operation \citep{Abel2018,Finn2017,Wilson2007,Konidaris2012,Thrun1995b}. However, each task requires a new policy to be stored, causing memory problems when a long sequence of tasks is considered. Moreover, finding an initial policy may be difficult to locate in parameter space when tasks require widely differing policies. 

Hierarchical approaches are proposed to improve transfer learning by learning sub-policies that achieve a subgoal or search for a rewarding high-level state  \citep{Thrun1995,Brunskill2014,Konidaris2012,Tessler2016,Li2019a}. Sub-policies used in this way represent solutions to sub-tasks which occur across a number of different tasks, reducing the number of required policies. While there are a few exceptions (e.g.,  \citep{Levy2019}), a downside of this approach is that (i) there is a need for two separate phases, one to learn sub-policies, and one to combine them; and (ii) that tasks may come in clusters in which similar tasks are more efficiently solved by a common policy.

\subsection{Meta-learning}
Meta-learning, in which a meta-level algorithm optimises the underlying learning algorithm itself by means of a meta-level objective, has been investigated in domains relevant for lifelong learning. In general, such methods share key aims, such as learning over an extended lifetime and improving generalisation but do not consider the same learning setting or the same goal to learn widely varying task sequences.

Meta-learning methods are particularly prevalent in supervised learning. For example, hyper-networks, in which a top-level super-ordinate network performs gradient descent to optimise a sub-ordinate network, can solve a variety of tasks from within a similar class \citep{Hochreiter2001,Andrychowicz2016}. 

In meta-RL, the literature focuses on generic aspects of RL such as improved exploration (e.g., \cite{Xu2018}). Other meta-RL approaches have focused specifically on settings relevant to lifelong learning: multi-task learning, in which multiple tasks are learned in batch (e.g., \cite{Finn2017}); continual learning, in which a limited number of similar tasks of increasing difficulty are presented (e.g., \cite{Riemer2018}); and lifetime RL methods that are guided by the lifetime average reward to improve the system in long-term environments (e.g., \cite{Bossens2019}). The lifetime policy reuse method is partly inspired by the latter approach.

\section{Lifetime Policy Reuse}
\label{sec: LPR}
Lifetime Policy Reuse avoids excessive memory consumption by using a number of policies fixed at design-time  which are continually refined based on incoming tasks across the lifetime. Contrasting to earlier policy reuse systems \citep{Fernandez2006,Rosman2016,Wang2018a}, Lifetime Policy Reuse has the following features (to which it owes its name):
\begin{itemize}
\item Rather than assuming convergence on the task, the successive presentation of one task may be a much more limited number of episodes as part of a much longer randomised sequence, and the policies in the library are adjusted throughout the entire lifetime of the RL agent based on the subset of tasks they are assigned to.
\item Policies are assigned to subsets of tasks by comparing their lifetime average reward compared to the other policies. 
\item The number of policies is fixed at all times, and thereby it scales better to lengthy lifetimes with many tasks.
\end{itemize}
The following provides a concrete description of Lifetime Policy Reuse, including the problem setting (see Section~\ref{sec: problem}), the policy selector, details on the adaptive mechanism for policy selection (Section~\ref{sec: policy-selector}), and the model-agnostic algorithm (see Section~\ref{sec: algorithm}). The full algorithm of Lifetime Policy Reuse is described with pseudo-code in Algorithm~\ref{alg: LPR}.

\begin{algorithm}
\caption{Lifetime Policy Reuse algorithm with conservative $\epsilon$-greedy policy selector.} \label{alg: LPR}
\begin{minipage}[c]{0.85 \textwidth}
\begin{algorithmic}[1]
{\small
\Require Number of policies $N_{\pi}$, Task set $\mathcal{F}$, Number of episodes per task block $N_b$.
\State Initialise policy library $\Lambda = \{\pi_1, \dots, \pi_{N_{\pi}}\}$.
\State Initialise greedy policy selector $\mu$ randomly.
\State Start the lifetime: $t \gets 0$.
\While{$t < T$}
\State Sample task from unknown task distribution: $\tau \sim \mathbb{P}[\mathcal{F}]$.
\For{Episode $i = 1, \dots, N_b$}
\State Select policy: with probability $1 - \epsilon$, set $\pi \gets \mu(\tau)$; otherwise select $\pi \in \Lambda$ randomly.
\State $j=0$
\State Sample initial state and observation from starting distribution: $s_0 \sim P_0$, $o_0 \sim O_0(s_0)$.
\While{Episode not done}
\State Select action: $a_j \sim \pi(o_j)$
\State Get reward: $r_j \sim r(s_j,a_j)$
\State Sample next state: $s_{j+1} \sim \mathcal{T}(s_j,a_j)$.
\State Sample next observation: $o_{j+1} \sim \mathcal{T}_{o}(s_{j+1},a_j)$.
\State $j \gets j + 1$.
\State $t \gets t + 1$.
\If{$j$ is training time}
\State Update base-learner: $\pi \gets \texttt{update-base-learner}(\pi,(o_j,a_j,r_j,o_{j+1}))$.
\EndIf
\EndWhile
\State Update greedy policy selector: $\mu(\tau) \gets \argmax_{\pi} \mathcal{R}_{\pi,\tau}$.
\EndFor
\EndWhile
}
\end{algorithmic}
\end{minipage}
\end{algorithm}

\begin{figure*}[htbp!]
\centering 
\tikzstyle{whiteblock} = [draw, rectangle,
    minimum height=5em, minimum width=8em]
\tikzstyle{container} = [draw, fill=blue!20,rectangle, inner sep=0.3cm]
\tikzstyle{block} = [draw, fill=blue!20, rectangle,
    minimum height=0.1\textwidth, minimum width=0.1\textwidth]
\tikzstyle{redcontainer} = [draw, fill=red!20,rectangle, inner sep=0.3cm]
\tikzstyle{redblock} = [draw, fill=red!20, rectangle,
    minimum height=0.1\textwidth, minimum width=0.1\textwidth]
    \tikzstyle{yellowblock} = [draw, fill=yellow!20, rectangle,
    minimum height=0.1\textwidth, minimum width=0.1\textwidth]
\tikzstyle{greencircle} = [draw, fill=green!20, circle,
    minimum height=0.1\textwidth, minimum width=0.1\textwidth]
\begin{tikzpicture}[scale=0.6, every node/.style={scale=0.6}]
    \node[block,name=P1,text width=0.2\textwidth, align=right,node distance=0.25\textwidth,text height=0.3\textwidth]{Policy 1};
    \node[block,name=P2, right of= P1,text width=0.2\textwidth, align=right,node distance=0.35\textwidth,text height=0.3\textwidth]{Policy 2};
     \node[name=dots, right of= P2,text width=0.2\textwidth, align=center, node distance=0.2\textwidth,text height=0.1\textwidth]{\mbox{\Huge $\dots$}};
     \node[block,name=P3, right of= dots,text width=0.2\textwidth, align=right,node distance=0.2\textwidth,text height=0.3\textwidth]{Policy N};
       \node [yellowblock, name=task drift, above of=P2,align=center,node distance=1.6in] {Policy selector};
       
       \node [greencircle, name=env, left of=task drift,align=center,node distance=1.9in] {Environment};
    \draw [->] (task drift.270) -- node[left]{select} (P2.90);
   \draw [->] (P2.80) -- node[right]{evaluate} (P2.80 |- task drift.290);
   \draw [->] (env.0) -- node[above]{task index}(task drift.180);
   
   \draw [<->] (env.290) -- node[sloped,below]{interface}(P2.120);
   
\end{tikzpicture}
\caption[Lifetime policy reuse]{Lifetime policy reuse. The following cycle is repeated: first, a policy selector selects a policy to be used; second, the policy interfaces with the environment for a prolonged amount of time (e.g. one episode), during which time it not only acts but also learns in the environment. When the policy has finished, it feeds back its performance to the policy selector.} \label{fig: multiplepolicies}
\end{figure*}

\subsection{Problem setting}
\label{sec: problem}
Let $\mathcal{F} = \{\tau_1,\tau_2,\dots,\tau_{N_{\tau}}\}$ be a finite set of unique RL tasks, either MDPs or POMDPs. The tasks share the same state-action space and discount factor but come with their own reward functions $\{r^{\tau_1}, \dots, r^{\tau_{N_{\tau}}}\}$ as well as their own transition dynamics $\{\mathcal{T}^{\tau_1}, \dots, \mathcal{T}^{\tau_{N_{\tau}}}\}$. 

The task sequence consists of consequent task-blocks, each of which is a time interval during which the same task is repeated for a number of episodes (or a particular time interval for non-episodic environments). At the start of each task-block, the current task $\tau$ is sampled with replacement from an unknown distribution $\mathbb{P}[\mathcal{F}]$ over the set of tasks $\mathcal{F}$. Then $\tau$ is presented until the end of the task-block. Task-blocks may be relatively short so that full convergence is not guaranteed on the task leading to the potential for catastrophic forgetting (i.e. a performance reduction upon revisiting the same task) as well as negative transfer (i.e. a performance reduction after learning other tasks).

The goal is to develop a fixed number of policies by assigning to each policy their best task specialisation based on their task-specific lifetime average $\mathcal{R}_{ij} = \frac{1}{t_{ij}} \sum_{k=0}^{t_{ij}} r_{ij}(k)$, where $i$ is the policy, $j$ is the task index, $k=0$ is the start of the lifetime, $t_{ij}$ is the number of time units that policy $i$ spent in task $\tau_j$, and $r_{ij}(k)$ is the reward obtained at time step $k$ of applying policy $i$ in task $\tau_j$.

The task-specific lifetime average reward is favoured over a few alternatives as the objective of the adaptive policy selector. First, the jumpstart, which is the performance difference at the start of the new task, and the most rapid improvement, which considers the improvement in performance over an initial time interval in the new task, may lead to rapid initial improvement but may be followed by stagnation. Second, while the asymptotic performance on the task is especially important for long-term scenarios, it is unknown during the learning process and therefore cannot provide online learning; instead, the lifetime average performance on the task allows online, incremental learning of which policy to select and if it converges, it converges to the asymptotic performance. Third, while the discounted cumulative reward is used to optimise the policies, this objective is not used in the policy selection because discounting defines the solution method (i.e., low-patience vs high-patience) rather than the goal. To give a practical example, in Pacman or Atari games, the RL community evaluates the performance of algorithms not on the objective of the learner but on the actual score, be it asymptotic or lifetime average performance. Discounting in continuing environments does not optimise the lifetime average but the lifetime average does (see e.g., \cite{Naik2019}).

\subsection{Policy selector}
\label{sec: policy-selector}
In Lifetime Policy Reuse, there is a natural trade-off between exploration, selecting an untested policy for the task, and exploitation, selecting the well-tested policy for the task. Bandit algorithms such as Upper Confidence Bound (UCB) and $\epsilon$-greedy are therefore prime candidates to function as policy selectors. The experiments in this paper will use $\epsilon$-greedy as policy selector. At the start of each episode, $\epsilon$-greedy selects the best policy with a probability $1-\epsilon$ and a randomly chosen policy in $\Lambda$ with probability $\epsilon$, where $\epsilon$ is a small probability, here set to $\epsilon=0.10$ to give sufficient opportunities to train the current best policy. The initial policy is chosen randomly and as long as a policy has not been selected for a task yet it cannot be the best policy and therefore can only be chosen with a probability of $\epsilon / N_{\pi}$; this conservative bias aims to first train the best policy repeatedly before selecting another policy, which avoids being inefficient when there are many policies and when reliable evaluation of a policy requires many episodes. $\epsilon$-greedy provides a higher performance compared to UCB (see Section~\ref{sec: cartpole} vs Appendix B), which is due to it providing a stable specialisation compared to UCB's frequent alternation between policies observed even at the end of the lifetime.
\subsection{Model-agnostic algorithm}
\label{sec: algorithm}
The algorithm first initialises a library of policies $\Lambda = \{\pi_1, \dots, \pi_{N_{\pi}}\}$ randomly in parameter space. At the start of each episode, the policy selector decides on the policy to be used for the current task based on the lifetime average reward. At any given time step within an episode, the selected policy interfaces with the environment as the base-learner would do in a traditional single-policy approach, including action selection as well as training and other parameter updates. After the episode finishes, a new policy is chosen by the policy selection module and the cycle repeats. The algorithm is illustrated in Figure \ref{fig: multiplepolicies} and described in pseudo-code in Algorithm~\ref{alg: LPR}. While further details on the algorithmic behaviour are base-learner and policy selector dependent, the model-agnostic algorithm results in specialising policies to selected subsets of tasks.

\section{Task capacity}
\label{sec: taskcapacity}
Lifelong reinforcement learning algorithms would greatly benefit from measures that help to determine how many tasks they can learn. In Lifelong Policy Reuse, for example, this would help to select a suitable number of policies, $N_{\pi}$. This of critical importance since, if the number of policies is too low, then the learning agent may experience negative transfer and catastrophic forgetting; if the number of policies is too high, this may lead to excessive memory consumption or slower learning on new tasks that are similar to previously seen tasks. This section proposes two measures for the task capacity: (i) the theoretical task capacity, the number of tasks that a policy can \textit{represent} to sufficient accuracy; and (ii) the empirical task capacity, the number of tasks that a policy can empirically learn to sufficient accuracy. The theoretical task capacity being purely representational does not require any preliminary runs while the empirical task capacity does but additionally accounts for learning properties and can serve as a benchmarking tool.

\subsection{Solving multiple tasks accurately}
A near-optimal lifetime policy reuse system yields for each task $\tau \in \mathcal{F}$ a lifetime average reward, $\mathcal{R}^{\tau} = \mathbb{E}_{\pi,\mathcal{T}^{\tau}} \left[\frac{1}{T} \sum_{t=0}^{T} r^{\tau}(s_t,\pi(s))\right]$, close to or equal to the optimal lifetime average reward, $\mathcal{R}^{\tau*} = \mathbb{E}_{\pi^{\tau*},\mathcal{T}^{\tau}} \left[ \frac{1}{T} \sum_{t=0}^{T} r^{\tau}(s_t,\pi^{\tau*}(s)) \right]$, where $\pi$ is the policy chosen for task $\tau$ and $\pi^{\tau*}$ is the optimal policy for task $\tau$. Defining the lifetime regret for a policy $\pi_{\theta}$ on task $\tau$ as $\mathcal{L}(\theta; \tau) = \mathcal{R}^{\tau*} - \mathcal{R}^{\tau}$, a near-optimal reinforcement learner has a negligible lifetime regret on all tasks in $\mathcal{F}$, which is formalised below through the notion of an epsilon-optimal partition (see Lemma 1). 

In addition to providing a notion of near-optimality, Lemma 1 below provides a sufficiency condition that illustrates that to obtain near-optimal policies, it is sufficient to analyse the states in which the policy performs sub-optimally. The lemma will make the following assumptions. First, rewards and expected rewards are normalised in $[0,1]$, which is not a limiting assumption as these can be scaled to obtain a more general case. Second, the tasks are continuing MDPs. The tasks are continuing in the sense that either the task is non-episodic or the different episodes are converted into a single MDP. In the latter case, one can re-design the reward function to penalise shorter episodes and define the transition from the terminal state as the initial state distribution. With the above assumptions, the lemma will make use of a task-specific occupation measure $\mathbb{P}[s |\pi_{\theta}, \tau]  = \frac{1}{T_{\tau}} \sum_{t=1}^{T_{\tau}} \mathbb{I}(s_t = s | \tau)$, which, based on the indicator $\mathbb{I}$, defines for a task $\tau$ the expected fraction of time spent in state $s \in \mathcal{S}$ when compared to the total time $T_{\tau}$.

\begin{lemma}
An \textbf{epsilon-optimal partition} defines for a given task $\tau \in \mathcal{F}$  a subset of parameter space $ \Theta_{\epsilon} \subset \Theta$ such that for each $\mathbf{\theta} \in \Theta_{\epsilon}$, the policy $\pi_{\theta}$ parametrised by  $\mathbf{\theta}$ satisfies $\mathcal{L}(\theta; \tau) \leq \epsilon \mathcal{R}^{\tau*}$. Define the optimal policy $\pi^{\tau*}$, arbitrary $\epsilon \geq 0$, $\mathcal{S}^{-} =  \{ s \in \mathcal{S} :  r^{\tau}(s,\pi_{\theta}(s)) \neq r^{\tau}(s,\pi^{\tau*}(s)) \}$. Then $\theta \in \Theta_{\epsilon}$ if $\sum_{s' \in \mathcal{S}^{-}} \mathbb{P}[s'| \pi_{\theta},\tau]r^{\tau}(s',\pi_{\theta}(s')) \leq \epsilon \mathcal{R}^{\tau*}$. \\
\end{lemma}
\textbf{Proof:} For arbitrary $\mathbf{\theta} \in \Theta$, its corresponding policy $\pi_{\theta}$ will have a set of states $\mathcal{S}^{+} \subset \mathcal{S}$ for which its chosen action yields $r^{\tau}(s,\pi_{\theta}(s)) = r^{\tau}(s,\pi^{\tau*}(s))$ for task $\tau \in \mathcal{F}$, while for the remainder of the states $\mathcal{S}^{-} = \mathcal{S} \setminus \mathcal{S}^{+}$ its action chosen yields  $r^{\tau}(s,\pi_{\theta}(s)) < r^{\tau}(s,\pi^{\tau*}(s))$. Setting the expected reward for any $s \in \mathcal{S}^{-}$ minimally, $0$, for $\pi_{\theta}$ allows dropping the summation over $\mathcal{S}^-$ in the equality $\mathcal{R}^{\tau} = \sum_{s \in \mathcal{S}^{-}} \mathbb{P}[s |\pi_{\theta}, \tau] r^{\tau}(s,\pi_{\theta}(s)) + \sum_{s' \in \mathcal{S}^{+}} \mathbb{P}[s' |\pi_{\theta}] r^{\tau}(s',\pi^{\tau*}(s'))$, and yields the following upper bound to the regret:
\begin{align*}
\mathcal{L}(\theta; \tau) \leq \mathcal{R}^{\tau*}  -  \sum_{s' \in \mathcal{S}^{+}} \mathbb{P}[s' |\pi_{\theta}, \tau] r^{\tau}(s',\pi_{\theta}(s')) \,.
\end{align*}
Therefore, $\theta \in \Theta_{\epsilon} \subset \Theta$ if
\begin{align*}
&\mathcal{L}(\theta; \tau) \leq  \epsilon \mathcal{R}^{\tau*} \\
&\sum_{s' \in \mathcal{S}^{+}} \mathbb{P}[s'| \pi_{\theta}, \tau]r^{\tau}(s',\pi_{\theta}(s')) \leq (1 -\epsilon) \mathcal{R}^{\tau*}  \\
&\sum_{s' \in \mathcal{S}^{-}} \mathbb{P}[s'| \pi_{\theta}, \tau]r^{\tau}(s',\pi_{\theta}(s')) \leq \epsilon \mathcal{R}^{\tau*} \,, 
\end{align*} 
which concludes the proof.

Lemma 1 implies that to find an $\epsilon$-optimal policy, it is sufficient to focus on those states $s \in \mathcal{S}^-$ in which its reward differs from that of the optimal policy. This result is especially useful for those MDPs where there is only a small subset of the states that contributes to the regret. The Cartpole MDP is an instance of such MDPs as the only sub-optimal states are those near termination; therefore the theoretical task capacity within a Cartpole MDP domain can be assessed by analysing states just before termination (see Section \ref{sec: theoretical-taskcapacity}).

Further, Lemma 1 comes with two corollaries.
\begin{corollary}\textbf{Changing transition dynamics limits transferability.} Let $\tau, \tau' \in \mathcal{F}$, $r^{\tau'} = r^{\tau}$, and $\mathcal{T}^{\tau} \neq \mathcal{T}^{\tau'}$. Let $\pi_{\theta}$ be a policy that satisfies $\sum_{s' \in \mathcal{S}^{-}} \mathbb{P}[s'| \pi_{\theta},\tau]r^{\tau}(s',\pi_{\theta}(s')) \leq \epsilon \mathcal{R}^{\tau*}$ as in Lemma 1. Then the analogous result for $\tau'$, namely 
\begin{align*}
\sum_{s' \in \mathcal{S}^{-}} \mathbb{P}[s'| \pi_{\theta},\tau']r^{\tau'}(s',\pi_{\theta}(s')) \leq \epsilon \mathcal{R}^{\tau*} \,,
\end{align*}
holds if 
\begin{align*}\sum_{s \in \mathcal{S}^{-}(\tau)} \Delta \mathbb{P}[s | \pi_{\theta}] \mathcal{L}(\theta,s) \leq \epsilon \Delta \mathcal{R}^* \,,
\end{align*}
where $\Delta \mathcal{R}^* :=  \mathcal{R}^{\tau'*} - \mathcal{R}^{\tau*}$, $\Delta \mathbb{P}[s | \pi_{\theta}] := \mathbb{P}[s | \pi_{\theta}, \tau'] - \mathbb{P}[s | \pi_{\theta}, \tau]$, and $\mathcal{L}(\theta,s) := r^{\tau*} - r^{\tau}(s,\pi_{\theta}(s)) = r^{\tau'*} - r^{\tau'}(s,\pi_{\theta}(s))$.
\end{corollary}
\textbf{Proof:} Let $\theta \subset \Theta$ for which
\begin{align*}
\sum_{s' \in \mathcal{S}^{-}} \mathbb{P}[s'| \pi_{\theta}, \tau]r^{\tau}(s',\pi_{\theta}(s')) \leq \epsilon \mathcal{R}^{\tau*} \,.
\end{align*}
Then $\pi_{\theta}$ is an epsilon-optimal policy for task $\tau$. However, for another task $\tau'\neq \tau$ with different transition dynamics model $\mathcal{T}$, the policy $\pi$ yields a different occupation measure $\mathbb{P}[s | \pi_{\theta}, \tau']$.
Since $r^{\tau'} = r^{\tau}$ and therefore also $\mathcal{S}^{-}(\tau) = \mathcal{S}^{-}(\tau')$, any increase in regret can only come from more visitations of $\mathcal{S}^{-}(\tau)$. Therefore, the added regret $\delta(\tau,\tau')$ can be upper-bounded by
\begin{align*}
\delta(\tau',\tau) &\leq \sum_{s \in \mathcal{S}^{-}(\tau)} \Delta \mathbb{P}[s | \pi_{\theta}] \mathcal{L}(\theta,s) \,.
\end{align*}
Consequently, $\epsilon$-optimality on $\tau'$ requires that
\begin{align*}
\mathcal{L}(\theta; \tau') &= \mathcal{L}(\theta; \tau) + \delta(\tau',\tau)  \\
						   &\leq \epsilon \mathcal{R}^{\tau*}  + \delta(\tau',\tau) \\
						   &\leq \epsilon \mathcal{R}^{\tau*} + \sum_{s \in \mathcal{S}^{-}(\tau)} \Delta \mathbb{P}[s | \pi_{\theta}] \mathcal{L}(\theta,s) \\
						   &\leq \epsilon \mathcal{R}^{\tau'*}
\end{align*}
implying 
\begin{align*}
\sum_{s \in \mathcal{S}^{-}(\tau)} \Delta \mathbb{P}[s | \pi_{\theta}] \mathcal{L}(\theta,s) \leq \epsilon \Delta \mathcal{R}^*  \,,
\end{align*}
concluding the proof.

When a given a policy $\pi_{\theta}$ satisfies an $\epsilon$-optimality guarantee of the type in Lemma 1, but the transition dynamics change, Corollary 1 provides a guarantee for transferring tasks if their expected change in regret due to visiting sub-optimal states is lower than a fraction of the change in optimal regret. 

\begin{corollary}
\textbf{Changing reward functions limits transferability.} Let $\tau, \tau' \in \mathcal{F}$, $r^{\tau'} \neq r^{\tau}$, and $\mathcal{T}^{\tau} = \mathcal{T}^{\tau'}$. Let $\pi_{\theta}$ be a policy that satisfies $\sum_{s' \in \mathcal{S}^{-}} \mathbb{P}[s'| \pi_{\theta},\tau]r^{\tau}(s',\pi_{\theta}(s')) \leq \epsilon \mathcal{R}^{\tau*}$ as in Lemma 1. Then the analogous result for $\tau'$, namely 
\begin{align*}
\sum_{s' \in \mathcal{S}^{-}} \mathbb{P}[s'| \pi_{\theta},\tau']r^{\tau'}(s',\pi_{\theta}(s')) \leq \epsilon \mathcal{R}^{\tau*}
\end{align*}
 holds if 
 \begin{align*}
 \sum_{s \in \mathcal{S}^{c}} \mathbb{P}[s | \pi_{\theta}, \tau] \Delta \mathcal{L}(\theta,s) \leq \epsilon \Delta \mathcal{R}^*
 \end{align*}
 for $\mathcal{S}^c := \{ s \in \mathcal{S} : r^{\tau}(s) \geq r^{\tau'}(s) \land s \in \mathcal{S}^{-}(\tau') \}$, $\Delta \mathcal{L}(\theta,s) := r^{\tau}(s,\pi_{\theta}(s)) - r^{\tau'}(s,\pi_{\theta}(s))$, and $\Delta \mathcal{R}^* :=  \mathcal{R}^{\tau'*} - \mathcal{R}^{\tau*}$.
\end{corollary}
\textbf{Proof:} Since $\mathbb{P}[s | \pi_{\theta}, \tau] = \mathbb{P}[s | \pi_{\theta}, \tau']$, any increase in regret can only come from more visitations of states  in $\mathcal{S}^c = \{ s \in \mathcal{S} : r^{\tau}(s) \geq r^{\tau'}(s) \land s \in \mathcal{S}^{-}(\tau') \}$, because these have their reward reduced in task $\tau'$.  The added regret $\delta(\tau,\tau')$ can therefore be upper-bounded by 
\begin{align*}
\delta(\tau',\tau) &\leq \sum_{s \in \mathcal{S}^{c}} \mathbb{P}[s | \pi_{\theta}, \tau] \Delta \mathcal{L}(\theta,s) \,.
\end{align*}
Therefore, 
\begin{align*}
\mathcal{L}(\theta; \tau') &= \mathcal{L}(\theta; \tau) + \delta(\tau',\tau)  \\
						   &\leq \epsilon \mathcal{R}^{\tau*}  + \delta(\tau',\tau) \\
						   &\leq \epsilon \mathcal{R}^{\tau*} + \sum_{s \in \mathcal{S}^{c}} \mathbb{P}[s | \pi_{\theta}, \tau] \Delta \mathcal{L}(\theta,s) \\
						   &\leq \epsilon \mathcal{R}^{\tau'*}
\end{align*}
implying 
\begin{align*}
\sum_{s \in \mathcal{S}^{c}} \mathbb{P}[s | \pi_{\theta}, \tau] \Delta \mathcal{L}(\theta,s) \leq \epsilon \Delta \mathcal{R}^*  \,,
\end{align*}
where $\Delta \mathcal{R}^* =  \mathcal{R}^{\tau'*} - \mathcal{R}^{\tau*}$, concluding the proof.

When a given a policy $\pi_{\theta}$ satisfies an $\epsilon$-optimality guarantee of the type in Lemma 1, but the reward function changes,  Corollary 2 provides a guarantee for transferring if the changed states (i.e. those for which the reward function is different) contribute at most an $\epsilon$-fraction of the optimal regret. 

When both reward function and transition dynamics are dissimilar, transferability of one task to another is low, resulting in scalability issues. This effect can be observed in a sequence of POcman tasks, where many policies are required for epsilon-optimality across the task set (see Section \ref{sec: performance}). By contrast, the Cartpole problem maintains the same reward function across tasks, improving the overlap between tasks.

\subsection{Theoretical task capacity}
\label{sec: theoretical-taskcapacity}
Epsilon-optimality introduced in Lemma 1 can be used to define how many policies are needed or equivalently, how many tasks a single policy can represent accurately. This quantity, called the \textit{theoretical task capacity}, is defined below and illustrated on simple examples as well as the 27-task cartpole domain, which will be revisited in Section \ref{sec: domains}.

The theoretical task capacity is defined representationally based on (i) the hypothesis class $\mathcal{H}$, the set of allowed policy representations (for example, the set of functions that can be fitted by the function approximator); (ii) the task set $\mathcal{F}$; and (iii) the tolerance, $\epsilon$, specified for epsilon-optimality.
\begin{definition}
The \textbf{theoretical task capacity} $C$ of a hypothesis class $\mathcal{H}$ with regard to a task set $\mathcal{F}$ is the average number of tasks per policy when selecting the minimal number of policies, $N_{\pi}$, that still $\epsilon$-optimally represents all the tasks,
\begin{equation}
\begin{split}
C(\mathcal{H},\mathcal{F},\epsilon) = \max_{N_{\pi}} N_{\tau} / N_{\pi} : \exists \Lambda = \{\pi^1,\dots,\pi^{N_{\pi}}\} \subset \mathcal{H}: \forall \tau \in \mathcal{F}: \\\exists \pi_{\theta} \in \Lambda: \mathcal{L}(\theta; \tau) \leq \epsilon \mathcal{R}^{\tau*} \,,
\end{split}
\end{equation}
where $N_{\tau}$ is the number of tasks and $N_{\pi}$ is the number of policies.
\end{definition}

An important aspect of the theoretical task capacity is its dependency on the task set $\mathcal{F}$. A higher number of tasks in the set leads to a higher upper bound on the task capacity. Whether the task capacity is actually increased will depend on the task similarity. To illustrate this principle, consider what happens when formulating a task space based on a few defining task dimensions, similar to the construction of the Cartpole domain (see Section \ref{sec: domains}). If the number of tasks increases and the task bounds are expanded, then the task capacity could potentially be reduced. If the number of tasks increases but the task bounds stay the same, then the theoretical task capacity is increased due to tasks in the set being similar; this feature is convenient when one has analysed the number of policies needed for a coarse-grained version of a larger task domain (see the 125-task domain in Section \ref{sec: analysis}).

\paragraph{Simple examples}
To illustrate the representational argument made for selecting the number of policies, below are two examples on how the representation of the base-learner limits its theoretical task capacity for a given task set.
\begin{example}
\textbf{Linear approximators for changing reward functions.} In this example, a set of 4 polynomial functions must be approximated, providing a simple example in which transition dynamics do not play a role but the reward function changes depending on the task. A policy here simply outputs the estimated function output, with 100\% probability. States are sampled independently and identically from a uniform distribution over the state space $\mathcal{S} = [0,1]$. Consider a task set $\mathcal{F} = \{ \tau_1,\tau_2,\tau_3,\tau_4  \}$. Each of its four tasks are defined by the corresponding functions to approximate: $f_1(s) = -s^2 + 2s$, $f_2(s) = s^2$, $f_3(s) = -0.01s^2 + 0.02s$, and $f_4(s) = 0.01s^2$; in other words, $\pi_{\tau_i}^{*}(s) = f_i$ for all $i \in  1,\dots,4$. Their reward function is the complement of the absolute error, that is, $r^{\tau_i}(s,\pi(s)) = 1 - |\pi(s) - \tau_i(s)|$ for all $i \in \{1,\dots,4\}$. Consider the hypothesis class to be linear classifiers of the form $\pi(s) = As$ where $A \in \mathbb{R}$ is a constant. 
\end{example}

Since the minimal number of linear functions that can be selected while maintaining expected error smaller than $\epsilon=0.25$ is $N_{\pi}=2$, the theoretical task capacity on this task set of size $N_{\tau}=4$ is given by $C = N_{\tau} / N_{\pi} = 2$. An example of such a function set is $\{ \pi^1(s) = s, \pi^2(s) = 0.01s \}$. $\pi^1(s)$ can approximate $f_1$ and $f_2$ with an expected error of $\int_{0}^{1} s - s^2 \, ds = 1/6$. However, $\pi^1$ cannot fit $f_3$ or $f_4$ to $\epsilon$-precision, since for $f_3$ the expected error is $\int_{0}^{1} s + 0.01s^2 - 0.02s \, ds = 0.493 > \epsilon$ and the expected error for $f_4$ is even larger. The function $\pi^2(s) = 0.01s$ can fit both $f_3$ and $f_4$ with an error of $0.01 \int_{0}^{1} s - s^2 \, ds = 1/600$. 

\begin{example}
\textbf{State-action pair representations for changing transition dynamics.} In this example, a set of 6 RL tasks have different transition dynamics but are otherwise the same, with state space $\mathcal{S}=\{0, 1, 2, 3, 4 \}$, action space $\mathcal{A}=\{-1,+1\}$, the initial state $s_0=0$, and the reward function $r$ gives a reward equal to the number of successive steps going successfully in one direction. Each task in $ \mathcal{F} = \{\tau_{1,2},\tau_{2,3},\tau_{3,4}, \tau_{3,2}, \tau_{2,1}, \tau_{1,0} \}$ is identified by its unique transition model. For all $\tau_{i,j} \in \mathcal{F}$, the transition model $\mathcal{T}^{\tau_{i,j}}$ defines $\mathcal{T}^{\tau_{i,j}}(s,a) = \max(0,\min(s+a,4))$ except that transitioning from state $i$ to state $j$ is not allowed.
\end{example}

Each task $\tau \in \mathcal{F}$ has reachable states, $\mathscr{R}^{+}(\tau)$, states that can be repeatedly visited until the end of the task, as well as unreachable states, $\mathscr{R}^{-}(\tau)$, and the expected reward varies across tasks. For example, in $\tau_{2,3}$ the optimal task-specific strategy is to take $a=+1$ until $s=2$, take $a=-1$ until $s=0$, etc. because state $s=3$ is not reachable. The optimal expected reward is $(0+1)/2=0.5$, $(0+1+2)/3=1$, or $(0+1+2+3)/4=1.5$, respectively, depending on the number of reachable states. Executing a policy optimal for one task $\tau$ will be sub-optimal for another task $\tau'$, with an expected reward of $0$ when the agent will attempt to keep moving either forward or backward but get stuck, or an expected reward of $\sum_{i=0}^{N - 1} \frac{i}{N}$, where $N=|\mathscr{R}^{+}(\tau)|$, as the agent will go back and forth over the smaller path spanned by $\mathscr{R}^{+}(\tau)$. A more expressive reinforcement learner with memory of the previous state could solve multiple tasks satisfactorily by switching direction if it observes its state is the same as before. Due to extending the path with one additional time step of standing still, the expected reward would be reduced by a factor $\frac{N}{N+1}$ compared to optimal, where $N$ is the number of reachable states for the task to be solved. Therefore, a single policy would be able to solve tasks for which $N\geq 3$ with a regret of $\mathcal{L}(\theta; \tau) \leq \epsilon \mathcal{R}^{\tau*}$ where $\epsilon=0.25$, but not other tasks with $N < 3$, which would come with greater regret. The number of policies needed is therefore lower-bounded by $N_{\pi} \geq 2$. However, by defining policies which move back and forth in $[0,1]$, $[0,2]$, $[0,3]$, $[1,4]$, $[2,4]$,and $[3,4]$, the setting $N_{\pi} = 6$ provides an upper bound to the required policies as this solves all the tasks optimally. Therefore, the corresponding task capacity is bounded by $1 \leq C \leq 3$.

\paragraph{Task cluster analysis of the 27-task cartpole domain}
When exact arguments are not possible, as is the case for deep-reinforcement learners in relatively complex tasks, the theoretical task capacity can be roughly approximated by identifying distinct clusters of tasks. This cluster analysis is now exemplified for deep reinforcement learners on the 27-task Cartpole domain, a domain that will be revisited in the empirical experiments of this paper.

The Cartpole domain includes different MDPs characterised by different transition dynamics. In this domain, a cart must balance a pole placed on its top by moving left or right with fixed force of \SI{1}{N}. Each time step is rewarded with $+1$ but the episode terminates when \textbf{(a)} the pole has an angle of more than $15^{\circ}$ from vertical; \textbf{(b)} when the distance of the cart to the centre is greater than $\SI{2.4}{m}$; or \textbf{(c)} 200 time steps are completed, in which case the solution is considered optimal. 27 different tasks are formed in this domain by varying pole length, pole mass, and cart mass. In general, the optimal strategy is to move left when the pole turns left and to move right when the pole turns right. Many actions will come with equal reward of $+1$, i.e. all those that do not finish the episode.

A single deep RL policy can represent a single task accurately since it can map arbitrary states onto arbitrary low-entropy probability distributions over actions. However, by Corollary 1, when different MDPs have different transition dynamics, a subset of such tasks will make the policy reach states in which its action is suboptimal. In the Cartpole domain, such suboptimal states can only be the states preceding the above-mentioned terminal states of the type \textbf{(a)} and \textbf{(b)}; these states are followed by a reward of 0 while all others are followed by a reward of 1. Therefore, relevant differences in policy requirements across tasks can be identified by observing for each task which states lead to cases (a) and (b) and how frequently.

A computationally cheap experiment is performed, requiring only  a few minutes runtime. The experiment tests a random uniform policy on all 27 tasks for 600,000 time steps each, which amounts to 20,000 to 60,000 episodes depending on task difficulty. Since case (b) is observed rarely (at most 15 times), while case (a) happens  between 17,000 and 30,000 times, depending on the length of the cart, the focus is on which states precede event (a) and how frequently. In this analysis, clear differences in the angular velocity and the episode length before event (a) are observed across tasks. The cartpole problem is symmetric, so the lowest absolute angular velocity is computed as a boundary between the safe and unsafe regions. With this methodology, 8 clusters of nearby points are observed, with means $\{(0.026, 17.5), (0.026, 30),(0.0125,30),(0.013,17.5),(0.006,17.5),(0.006,30),(0.004,17.5),(0.004,30)\}$. Consequently, for epsilon-optimality on each task, a rough estimate of the number of policies required is $N^{\pi}=8$, for a theoretical task capacity of $C=N_{\tau}/N_{\pi} = 3.4$.

\subsection{Empirical task capacity}
\label{sec: empirical-taskcapacity}
A near-optimal set of policies may not necessarily be found, even if it is possible to represent them. The theoretical task capacity ignores various learnability issues within the hypothesis class $\mathcal{H}$, such as (i) the expressive efficiency which implies higher capacity methods require more data \citep{Sontag1998,Cohen2017}; (ii) the exploration technique providing representative samples; (iii) the inductive bias of the optimisation (e.g. gradient-descent methods \citep{Rumelhart1986} cannot easily escape local optima in non-convex landscapes); and (iv) the effects of learning without convergence and updates cancelling each other out. For these reasons, an empirical task capacity is defined that accounts for learnability. In addition to $\mathcal{H}$, $\mathcal{F}$, and $\epsilon$, the empirical task capacity is also dependent on the temporal structure of the environment and the training algorithm of the base-learner. The empirical task capacity is based on a notion of \emph{relative} epsilon-optimality for a base-learner, where its one-to-one mapping of policies to tasks represents optimality for that base-learner.

\paragraph{Definition}
Base-learners with high empirical task capacity are those for which a low number of policies results in a performance close to, or even better than, a one-to-one mapping of tasks to policies. The \textbf{empirical task capacity} $C_{\text{emp}}$ is defined in Equation \ref{eq: capacity},
\begin{equation}
\label{eq: capacity}
C_{\text{emp}} = N_{\tau}/N_{\pi}^{*} \,,
\end{equation}
where $N_{\tau}$ is the total number of tasks and $N_{\pi}^{*}$ is the lowest number of policies that yields \begin{equation}
\label{eq: tolerance}
 \mathcal{R} \geq (1-\epsilon_c) \mathcal{R}^{\text{1-to-1}} \, ,
 \end{equation}
where $\mathcal{R}^{\text{1-to-1}}$ is the lifetime average performance of the one-to-one mapping with task-specific policies. Learners with high $C_{\text{emp}}$ and lower $\epsilon_c$ can solve on average a number of $C_{\text{emp}}$ tasks with a regret of at most $\epsilon_c$ percentage of $\mathcal{R}^{\text{1-to-1}}$. 

The choice of $\epsilon_c$ is ultimately up to the end-user, based on the trade-off between resources used and performance. However, to provide a complete evaluation of a base-learner, as a benchmarking tool, Equation \ref{eq: ITC} defines the \textbf{Integrated Task Capacity} (ITC),
 \begin{equation}
\label{eq: ITC}
\text{ITC} = \int_{0}^{1} C_{\text{emp}}(\epsilon_c) d\epsilon_c\, ,
 \end{equation}
which integrates the task capacity over the possible values of $\epsilon_c \in [0,1]$.

\paragraph{Suggested usage}
The empirical task capacity is proposed for two use cases. First, it can be used to compare base-learners over a domain and observe which base-learner scales best over tasks. Second, it can be used, analogous to the theoretical task capacity, as a tool for pre-selecting the number of policies. In this case, the user selects a task sequence that represents the application of interest and then estimates the optimal number of policies. 

\section{Experiment setup}
\label{sec: setup}
\subsection{Base-learners}
\label{sec: base-learners}
Two base-learners, Deep Q-Network and Proximal Policy Optimisation, are compared in the study. They represent two classes of state-of-the-art deep RL methods, value-based learners and actor-critic learners. In both cases, a single actor is chosen instead of distributed RL with many actors, because in a realistic, continual lifelong learning setting the learners are not able to perfectly simulate copies of their environment. Their architectures are chosen to be as similar as possible, as depicted in Figure \ref{fig: architecture-diagram}.

\begin{figure*}[htbp!]
\centering
\hspace{1cm}
\subfigure[DRQN]{
\begin{tikzpicture}[scale=0.6, every node/.style={scale=0.6}]
\tikzstyle{whiteblock} = [draw, rectangle,
    minimum height=5em, minimum width=8em]
\tikzstyle{container} = [draw, fill=blue!20,rectangle, inner sep=0.3cm]
\tikzstyle{block} = [draw, fill=blue!20, rectangle,
    minimum height=0.1\textwidth, minimum width=0.1\textwidth]
\tikzstyle{redcontainer} = [draw, fill=red!20,rectangle, inner sep=0.3cm]
\tikzstyle{redblock} = [draw, fill=red!20, rectangle,
    minimum height=0.1\textwidth, minimum width=0.1\textwidth]
    \tikzstyle{yellowblock} = [draw, fill=yellow!20, rectangle,
    minimum height=0.1\textwidth, minimum width=0.1\textwidth]
\tikzstyle{greencircle} = [draw, fill=green!20, circle,
    minimum height=0.1\textwidth, minimum width=0.1\textwidth]
	    \node[block,name=L0,text width=0.12\textwidth, align=center,node distance=0.15\textwidth,text height=0.10\textwidth]{Observation \\ $o_t$};
    \node[block,name=L1,right of=L0,text width=0.1\textwidth, align=center,node distance=0.15\textwidth,text height=0.10\textwidth]{Dense RELU (80)};
    \node[block,name=L2, right of= L1,text width=0.1\textwidth, align=center,node distance=0.15\textwidth,text height=0.10\textwidth]{LSTM tanh (80)};
     \node[block,name=L3, right of= L2,text width=0.1\textwidth, align=center,node distance=0.15\textwidth,text height=0.10\textwidth]{$Q(o_t,a)$ $\forall a \in \mathcal{A}$};
    \draw [->] (L0.0) -- (L1.180);
    \draw [->] (L1.0) -- (L2.180);
    \draw [->] (L2.0) -- (L3.180);

\end{tikzpicture}
}
\hspace{1cm}
\subfigure[PPO]{
\begin{tikzpicture}[scale=0.6, every node/.style={scale=0.6}]
\tikzstyle{whiteblock} = [draw, rectangle,
    minimum height=5em, minimum width=8em]
\tikzstyle{container} = [draw, fill=blue!20,rectangle, inner sep=0.3cm]
\tikzstyle{block} = [draw, fill=blue!20, rectangle,
    minimum height=0.1\textwidth, minimum width=0.1\textwidth]
\tikzstyle{redcontainer} = [draw, fill=red!20,rectangle, inner sep=0.3cm]
\tikzstyle{redblock} = [draw, fill=red!20, rectangle,
    minimum height=0.1\textwidth, minimum width=0.1\textwidth]
    \tikzstyle{yellowblock} = [draw, fill=yellow!20, rectangle,
    minimum height=0.1\textwidth, minimum width=0.1\textwidth]
\tikzstyle{greencircle} = [draw, fill=green!20, circle,
    minimum height=0.1\textwidth, minimum width=0.1\textwidth]
	    \node[block,name=L0,text width=0.12\textwidth, align=center,node distance=0.15\textwidth,text height=0.10\textwidth]{Observation \\ $o_t$};
    \node[block,name=L1,right of= L0,text width=0.1\textwidth, align=center,node distance=0.15\textwidth,text height=0.10\textwidth]{Dense RELU (80)};
    \node[block,name=L2, right of= L1,text width=0.1\textwidth, align=center,node distance=0.15\textwidth,text height=0.10\textwidth]{LSTM tanh (80)};
     \node[block,name=L3a, right=0.5cm of L2.north east,text width=0.10\textwidth, align=center,text height=0.05\textwidth]{Actor $\pi(s,a)$ \\ $\forall a \in \mathcal{A}$};
          \node[block,name=L3b, right=0.5cm of L2.south east,text width=0.1\textwidth, align=center,text height=0.05\textwidth]{Critic $V(o_t)$};
    \draw [->] (L0.0) -- (L1.180);
    \draw [->] (L1.0) -- (L2.180);
    \draw [->] (L2.0) -- (L3a.180);
    \draw [->] (L2.0) -- (L3b.180);

\end{tikzpicture}
}
\hspace{3cm}
\caption[]{Architectures used in the experiments. The observation $o_t$ at time $t$ is input to a feedforward layer with RELU activations, which is further processed by an LSTM layer with $tanh$ activations, and then further processed by a linear layer which outputs the $Q(o_t,a)$ for all $a \in \mathcal{A}$. A single arrow represents connections between each unit in the outgoing layer and each unit in the subsequent incoming layer. For MDP tasks, the LSTM layer is replaced by a second Dense RELU layer, also of 80 neurons.} \label{fig: architecture-diagram}
\end{figure*}

\paragraph{Deep (Recurrent) Q-Network}
For MDP tasks, we investigate Deep Q-Network \citep{Mnih2015}, a state-of-the-art value-based reinforcement learner suitable for discrete action spaces and complex state spaces, as is the case for example in Atari game environments. Its policy depends on a state-action value, and its updates are done in an off-policy style, allowing the training to be independent of the current policy and to repeat events in the distant past. For POMDP tasks, Deep Recurrent Q-Network (DRQN) is chosen due to its track record in partially observable video games \citep{Hausknecht2015,Schulze2018,Lample2016,Chaplot2017a}. DRQN extends Deep Q-Network with a different experience replay method that samples  traces of observations rather than a single observation. This allows it to learn from sequences of observations with a Long Short-Term Memory \citep{Hochreiter1997} layer, making it suitable for partially observable environments. A key question in both learners is whether negative transfer results in DQN-based, and other Q-learning-based lifelong learning systems \citep{Li2018a,Kirkpatrick2017}, can be replicated in other types of tasks and whether these results also hold for DRQN.

DQN repeatedly performs control cycles in which first it outputs for each action its Q-value, $Q(s,a)$ and then selects an action based on these Q-values. To select an action, $\epsilon$-greedy is used, such that with probability $1 - \epsilon$ the action with the highest Q-value, $a^* = \argmax_{a \in \mathcal{A}} Q(s,a)$,  is taken while with probability $\epsilon$ a random action is taken. Then periodically, updates are performed using experience replay. In DQN, updates are based on Q-learning \citep{Watkins1992} and the loss function is defined in Equation \ref{eq: DRQN_loss_function},
\begin{equation}
\label{eq: DRQN_loss_function}
L(\theta) = \mathbb{E}_{(s,a,r,s') \sim U(\mathcal{B})} [( r + \gamma \max_{a'} Q(s',a';\hat{\theta}) - Q(s,a; \theta))^2 ] \,,
\end{equation}
where experience tuples $(s,a,r,s')$ of state, action, reward and next state, are sampled uniformly from a buffer $\mathcal{B}$, and  where a factor $\gamma$ discounts future rewards. The policy's parameters are $\theta$, which are updated frequently, whilst another set of policy parameters $\hat{\theta}$ are synchronised infrequently on a periodic basis to match the policy's parameters and are used as the target for the neural network. 

For initialising DRQN, burn-in initialisation \cite{Kapturowski2019} is chosen, where the agent selects a number of random actions to initialise the trace of the recurrent network; for DQN this step is not taken. In DRQN, the updates and the loss function are the same as in DQN except that the sampled states are traces of observations which are passed through an LSTM layer such that the system remembers previous time steps. 

In the present paper's experiments, the buffer of D(R)QN is treated as part of the policy, and therefore each policy will have a separate experience buffer; this ensures that experiences are sampled only from the subset of tasks on which the policy specialises. The convolutional layers are replaced by a single densely connected layer due to the small observation without spatial correlations. Further details of its implementation, including the hyper-parameters, are given in Appendix A.
\paragraph{Proximal Policy Optimisation with/without LSTM network}
Proximal Policy Optimisation (PPO) \citep{Schulman2017a} is a policy optimisation algorithm which is often applied as an actor-critic reinforcement learner. PPO is chosen because of its robustness and transfer learning benefits \citep{Burda,Heess2017,Nichol2018}. It applies a clipped loss function, shown in Equation \ref{eq: PPO_loss_function}, that takes into account that the policy updates should not be too large, to allow monotonic improvements,
\begin{equation}
\label{eq: PPO_loss_function}
L(\theta) = \min(g_t \hat{\mathbb{A}}_t, clip(g_t; 1-\epsilon,1+\epsilon) \hat{\mathbb{A}}_t) \,,
\end{equation}
where $\hat{\mathbb{A}}_t$ is the advantage estimated at time $t$ by Generalised Advantage Estimation
\citep{Schulman2016}, $g_t=\pi(a_t|o_t, \theta)/\pi(a_t|o_t,\theta_{old})$ is the ratio of the
probability of the chosen action $a_t$ according to the new parameters $\theta$ and the old parameters $
\theta_{old}$, and $clip(x; y,z)$ clips the number $x$ to the interval $[y,z]$. This method is applied in an actor-critic style, in which a critic learns the value of a given observation according to $V(o_t)=\mathbb{E} [ \sum_{i=0}^{\infty} \gamma^{i} r_{t+i} ]$, with $r_t$ the reward at time $t$, and in which the actor learns $\pi(a_t|o_t, \theta)$, the probability of the chosen action in the observation $o_t$, based on the current policy parameters $\theta$. 

Analogous to the D(R)QN setting, for POMDP tasks the neural network architecture is supplemented with a Long Short-Term Memory layer \citep{Hochreiter1997} and burn-in initialisation \citep{Kapturowski2019} is the chosen initialisation at the start of the episode. The architecture differs from DRQN in the sense that there are two output layers, one for the actor and one for the critic, that share their connections to the LSTM layer. The actor's output, $\pi(a|s) \, \forall a \in \mathcal{A}$, provides a probabilistic policy over actions which is directly used for action selection. Further details of its implementation, including the hyper-parameters, are given in Appendix A.

\subsection{Learning metrics}
\label{sec: othermetrics}
To assess forgetting and transfer, we now construct a forgetting ratio and a transfer ratio, which define forgetting by a negative forgetting ratio and negative transfer using a negative transfer ratio. The ratios are similar to transfer learning metrics and are directly based on the performance after transitioning from earlier tasks to the task of interest \citep{Steunebrinka,Taylor2009}, and in particular, the area ratio \citep{Taylor2009}, which compares the performance of a policy with experience on other tasks to a policy that is randomly initialised. Like the area ratio, our metrics integrate the performance across the entire task-block to capture the complete learning behaviour, contrasting to jumpstart, time to threshold, and asymptotic performance. Due to the existence of many task-blocks of the same task, the proposed transfer and forgetting ratios are first computed for all task-block transitions across the lifetime. Forgetting ratios are then aggregated for different bins of the number of interfering task-blocks while transfer ratios are aggregated for different bins of the number of prior task-blocks.

The \textbf{forgetting ratio} is defined in Equation \ref{eq: forgetting ratio} as
\begin{equation}
\label{eq: forgetting ratio}
\text{forgetting ratio}= \dfrac{\Delta - \Delta^{\text{1-to-1}}}{\text{area-uniform-random}} \,,
\end{equation}
where $\Delta = \text{area-after} - \text{area-before}$ is the improvement in performance area under the curve comparing the performance on the current task-block to that of the prior task-block of the same task, $\Delta^{\text{1-to-1}}$ is the analogous improvement for the 1-to-1 policy selection, and area-uniform-random is the performance area under the curve for a uniform random policy, i.e. a policy that selects actions randomly from the action set.

The \textbf{transfer ratio} is defined in Equation \ref{eq: transfer ratio} as
\begin{equation}
\label{eq: transfer ratio}
\text{transfer ratio}= \dfrac{\text{area-after} - \text{1-to-1-area}}{\text{area-uniform-random}} \,,
\end{equation}
where area-after, or area-with-transfer, is the performance area under the curve for the current task-block and 1-to-1-area is the performance area under the curve for the 1-to-1 policy. The 1-to-1 policy performance on the task-block is equal to the area-without-transfer used in the area ratio since the transfer ratio is only computed for the first presentation of the target task.

The transfer ratio differs from the area ratio in the use of the uniform random policy performance as the denominator; this compares different base-learners in the same units and avoids the sensitivity to scale mentioned in \cite{Taylor2009}. The forgetting ratio also is formulated using the uniform random policy in the denominator; dividing by the uniform random policy as opposed to the performance of the base-learner on a previous block avoids sequential effects (e.g. consequent task-blocks after a completely forgotten task could potentially have a performance near zero, giving a high score and dominating the aggregated forgetting ratio). The correction for the 1-to-1 policy's improvement $\Delta^{\text{1-to-1}}$ represents the expected transition improvement with no forgetting rather than the improvement in performance over the previous task-block which is subject to sequential effects; for instance, in late stages of training, only minor improvements in performance can be gained over the previous task-block. After aggregation of task transitions for bins of the number of prior task-blocks, the forgetting ratio determines the effect of interfering task-blocks on performance. The transfer ratio has no sequential effects, as only the first task-block is being assessed for any given task. Therefore, the transfer ratio can also be interpreted as is, without aggregation. 

\paragraph{Policy spread} An additional metric, called the \textbf{policy spread}, estimates the variability in the policies during the lifetime.  The policy is a function of the parameter space and the observations, and therefore, when different policies are spread widely in parameter space, this does not necessarily imply a wide spread in the policies. Therefore, the variability in the policies is assessed empirically by randomly sampling viable observations and determining the selected action probabilities. In PPO, this is directly based on the output of the actor-network which outputs a probability for each action $a \in \mathcal{A}$. In DRQN, this is based on the $\epsilon$-greedy exploration, which first obtains the action $a_{max}=\argmax_{a'} Q(o_t,a_t)$ and then assigns the probability $1-\epsilon + \epsilon/|\mathcal{A}|$ for $a_{max}$ and the probability $\epsilon/|\mathcal{A}|$ for actions other than $a_{max}$. The calculation of the spread is based on the total variation distance, a distance metric for probability distributions, applied to all pair-wise combinations of the policies' action probabilities in an experimental condition.
\subsection{Lifelong learning domains}
\label{sec: domains}
The experimental setup tests the impact of the number of policies on performance, probing in particular the effects of changing reward functions and transition dynamics as these limit the scalability (see Corollary 1 and 2). The experiments are conducted in a 27-task Cartpole MDP domain, in which a feedforward DQN and PPO are used, and an 18-task POcman POMDP domain, in which case an LSTM layer is added for a recurrent version of PPO and DRQN. 

These domains are chosen to illustrate the challenge of lifetime policy reuse, involving randomly chosen tasks presented in rapid succession. In both domains, the learner does not have time to converge its parameters, memories of earlier tasks can be lost by the time they are presented again, and policies learned on one task may transfer negatively to other tasks. The differences between these two domains help to analyse critical factors underlying its success: (i) changing reward functions; (ii) changing transition dynamics; (iii) partial observability; and (iv) task similarity. The MDP domain has a relatively large task-similarity and has full observability. The POMDP domain has lower task-similarity and has changes in reward functions, which combined with the partial observability makes it challenging to infer the task from within the policy; this contrasts to Atari and other video games where a single observation is sufficient to recognise the task. Such partial observability is often true in the real world, where opposite policies are often required but this is unknown to the learner due to the effect of hidden variables. Therefore, the POMDP domain allows to assess  strategies for lifetime policy reuse in low task-capacity scenarios.

\paragraph{27-task Cartpole MDP domain}
In the 27-task Cartpole MDP domain, reinforcement learners are tasked with moving a cart back and forth to balance a pole for as long as possible. The domain includes varying transition dynamics but not varying reward functions. 
27 unique tasks are based on varying properties of the cart and the pole: (i) the mass of the cart varies in $\{0.5,1.0,2.0\}\SI{}{kg}$; (ii) the mass of the pole varies in $\{0.05,0.1,0.2\}\SI{}{kg}$; and (iii) the length of the pole varies in $\{ 0.5,1.0,2.0\}\SI{}{m}$. Due to the tasks being relatively similar, this domain represents an optimistic evaluation of task capacity. 

The lifetime of the learner starts at $t=0$ and ends at $t=T$, where $T$ is chosen to be 40.5 million time steps. At each time step, the learner receives an observation $\langle x, \theta, \dot{x}, \dot{\theta} \rangle$, where $x$ and $\dot{x}$ are the position and velocity of the cart, and $\theta$ and $\dot{\theta}$ are the angle of the pole to vertical and its rate of change. The learner is equipped with a fixed set of 2 actions, $\mathcal{A}=\{\texttt{left},\texttt{right}\}$, moving the cart left or right by applying a force of $-\SI{1}{N}$ or $+\SI{1}{N}$. The observations and actions follow the Markov property such that $\mathbb{P}(o_t | o_{t-1} a_{t-1} \dots o_1 a_t) = \mathbb{P}(o_t | o_{t-1} a_{t-1})$. Each time step is rewarded with $+1$ but the episode terminates when either the pole has an angle of more than $15^{\circ}$ from vertical or when the distance of the cart to the centre is greater than $\SI{2.4}{m}$. The episode is also terminated after 200 time steps of balancing as in this case the learner is considered to be successful.

In this experiment, tasks are presented in a random sequence from a uniform random distribution, with each task sequence providing a unique order. To account for the variability in the experiments due to the stochasticity in the actions selected by the reinforcement learner, as well as the random task presentation, the learners are evaluated based on 27 independent runs. Each independent run consists of 40.5 million time steps and corresponds to a unique task sequence. Each task sequence consists of 675 task-blocks. Each task-block consists of 60,000 time steps, or at least 300 episodes, of the same given task. With 27 unique tasks, this allows on average 25 blocks per task. Each episode takes at most 200 time steps, depending on the success of the learner. To minimise the effects of task order, the tasks are spread evenly between the different task sequences, as is illustrated in Figure \ref{fig: experiment-sets}.

\begin{figure*}[htbp!]
\centering
\resizebox{0.5\textwidth}{!}{
\begin{tikzpicture}[circle dotted/.style={dash pattern=on .05mm off 4mm,
                                         line cap=round}]
	
    \draw[->] (0,0) -- (12 ,0) node[below]{time};
    \foreach \x/\xtext in {
    0/$\tau_0$,
    1/$\tau_5$,
    2/$\tau_0$,
    3/$\tau_6$,
    4/$\tau_{10}$,
    5/$\dots$,
    6/$\tau_{1}$,
    7/$\tau_2$,
    8/$\tau_2$,
    9/$\tau_{9}$,
    10/$\tau_{17}$,
    }{
      \draw (\x,-.2) -- (\x,.2) node[above] {\xtext};
    }

    \draw[->] (0,-1.5) -- (12,-1.5) node[below](arrow2){time};
    \foreach \x/\xtext in {
    0/$\tau_1$,
    1/$\tau_6$,
    2/$\tau_1$,
    3/$\tau_7$,
    4/$\tau_{11}$,
    5/$\dots$,
    6/$\tau_{2}$,
    7/$\tau_3$,
    8/$\tau_3$,
    9/$\tau_{10}$,
    10/$\tau_{18}$,
    }{
      \draw (\x ,-1.7 ) -- (\x ,-1.3) node[above] {\xtext};
    }
    
  \draw[line width = 1mm,circle dotted] (4.5,-2.6) -- (4.5,-3.8);
    \draw[->] (0,-5) -- (12 ,-5) node[below]{time};
    \foreach \x/\xtext in {
    0/$\tau_{17}$,
    1/$\tau_4$,
    2/$\tau_{17}$,
    3/$\tau_5$,
    4/$\tau_{9}$,
    5/$\dots$,
    6/$\tau_{0}$,
    7/$\tau_1$,
    8/$\tau_1$,
    9/$\tau_{8}$,
    10/$\tau_{16}$,
    }{
      \draw (\x,-5.2) -- (\x,-4.8) node[above] {\xtext};
    }
\end{tikzpicture}
}
\caption[Illustration of the task sequences]{Illustration of the task sequences assuming a number of tasks $N_{\tau}=18$. To form the zeroth task sequence, task indices are sampled randomly in $\{0,1,\dots,N_{\tau} - 1\}$ for each successive task-block. To form the $n$'th task sequence, where $n > 0$, the task index is chosen for each task-block as $j=(i+n) \mod N_{\tau}$, where $i$ is the task index in the zeroth task-sequence. Thus all tasks are represented exactly once when taking a vertical slice across sequences at any given time-point; apart from varying the task-order this also facilitates analysis due to a smooth aggregate development curve over time. $N_{\tau}=18$ represents the 18-task POcman POMDP domain. For the 27-task Cartpole MDP domain, the above illustration applies after setting $N_{\tau}=27$.} \label{fig: experiment-sets}
\end{figure*}

\paragraph{18-task POcman POMDP domain} 
In the 18-task POcman POMDP domain, reinforcement learners interact with 18 tasks each of which contains an object of interest and has an unknown grid-world topology. While the smaller state space reduces the number experiences required, the lifelong learning domain is extremely  challenging due to including varying transition dynamics, varying reward functions, and partial observability.

\begin{figure}[htbp!]
\centering
\subfigure[positive, dynamic, cheese maze]{\includegraphics[width=0.2\textwidth]{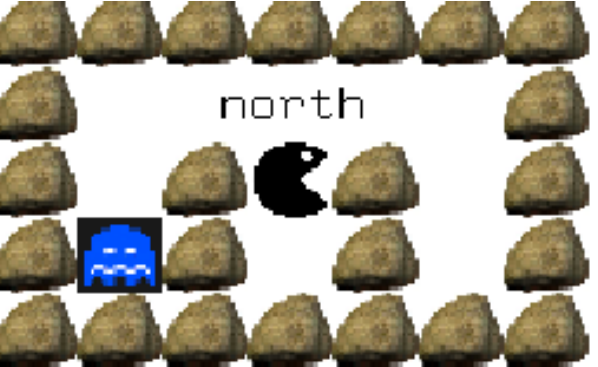}}
\quad
\subfigure[negative, dynamic, POcman maze]{\includegraphics[width=0.2\textwidth]{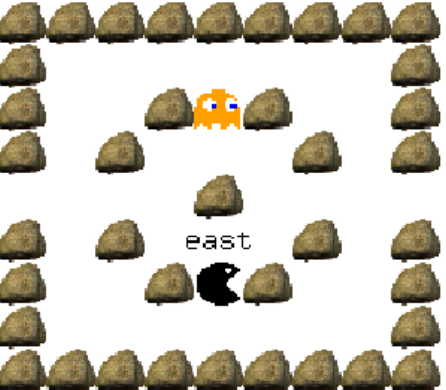}} \\
\subfigure[positive, static, Sutton maze]{\includegraphics[width=0.2\textwidth]{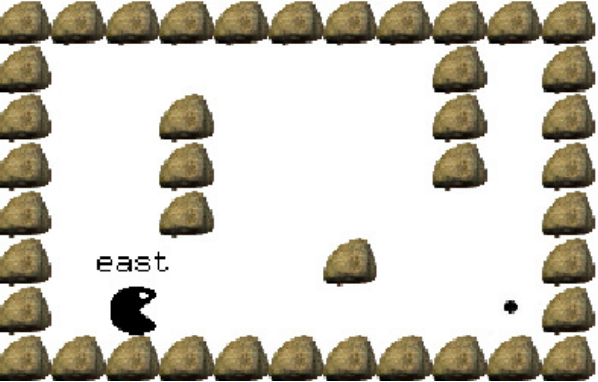}}
\quad
\subfigure[negative, static, Sutton maze]{\includegraphics[width=0.2\textwidth]{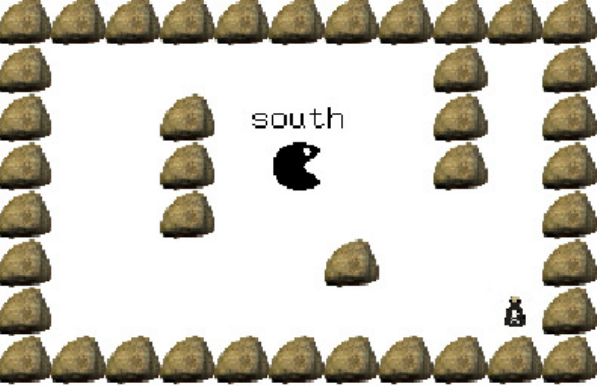}}  \\
\caption[Illustration of POcman tasks based on three defining task characteristics.]{Illustration of POcman tasks based on three defining task characteristics. (a) in a positive dynamic task, the learner must touch the defensively moving ghost; (b) in a negative dynamic task, the learner must avoid the aggressively moving ghost; (c) in a positive static task, the learner must touch the food pellet; (d) in a negative static task, the learner must avoid touching the poison bottle. Note that the objects are illustrated in this way purely for interpretation, whereas the learner perceives the objects as the same observation on all tasks.} \label{fig: taskpics}
\end{figure}

As illustrated in Figure \ref{fig: taskpics}, the environment's 18 distinct tasks are based on three dimensions, $(\texttt{R},\texttt{M},\texttt{T})$, where: $\texttt{R} \in \{-1.,1.\}$ is the reward for touching the main object in the task; $\texttt{M} \in \{0,1,2\}$ is a dimension specifying the movement of the object of interest, with 0 indicating static, 1 indicating a random step (north, east, south or west) once every 20 time steps, and 2 indicating a  strategy which reacts each time step to the RL agent, by moving away defensively  in case $\texttt{R}=1.$, or by aggressively moving towards the agent in case $\texttt{R}=-1$;\footnote{The defensive strategy is to move away with a 50\% probability and otherwise to stand still, and the aggressive strategy is to move towards the agent with a 50\% probability and otherwise to take a random action.} $\texttt{T} \in \{0,1,2\}$ is the grid-world topology corresponding to the cheese-maze \citep{McCallum1995}, Sutton's maze as mentioned in \citep{Schmidhuber1999}, and a 9-by-9 version of the partially observable pacman (POcman) \citep{Veness2011}. When the object is static,  the learner must find the location with the object, if $\texttt{R}=1$, or without the object, if $\texttt{R}=-1$. When the object is dynamic, the learner should either follow the object as closely as possible, if $\texttt{R}=1$, or stay away at a safe distance, if $\texttt{R}=-1$. Objects are never removed even when the agent touches, allowing more frequent contact between the agent and the object and the corresponding rewards until the elementary task ends. To ensure a certain difficulty level and emphasise a searching strategy rather than simple goal achievement, the initial coordinates of the object of interest are not fixed over time, but are randomly chosen from a set of $(x,y)$-coordinates; see Appendix A for starting coordinates.

The lifetime of the learner starts at $t=0$ and ends at $t=T$, where $T$ is 90 million time steps. The learner's actions are $\mathcal{A}=\{\texttt{north},\texttt{east},\texttt{south},\texttt{west},\texttt{stay}\}$. Observations are 11-bit vectors similar to those in POcman \citep{Veness2011}: the first four indicate for each cell in a Von Neumann-neighbourhood whether it contains an obstacle or not; the next four indicate for each cell in a Von Neumann-neighbourhood whether it contains the object of interest; and the final three bits indicate whether or not the object is within a Manhattan distance of 2, 3 or 4 steps from the learner. There are no terminal states but the episode ends after $1000$ time steps, after which the learner is reset to the starting location. In the first $15$ time steps after being reset, any external action is chosen randomly by the learner to allow the learner to form an initial trace of observations for its policy. 

As in the MDP task sequences, tasks are presented in a random sequence from a uniform random distribution and each task sequence provides a unique order. Independent runs account for the stochasticity in the learner and environment and task order effects are minimised by spreading tasks evenly between the different task sequences. In the POMDP domain, each task sequence consists of 450 task-blocks with 200 episodes of the same given task. 

\subsection{Experimental conditions and analyses}
To demonstrate the proposed approach, a first study compares the performance of Lifetime Policy Reuse across different settings of the number of policies as well as an ablation without the adaptive policy selector. For the unadaptive policy selector, the policy is selected based on a fixed, balanced partitioning of tasks. For unadaptive mappings, the experiments manipulate the number of policies $N_{\pi} \in \{1,2,4,9,14,27\}$ in the 27-task MDP domain and $N_{\pi} \in \{1,2,4,9,18\}$ in the 18-task POMDP domain. The adaptive policy selection conditions are the same, except that the 1-to-1 policy selection and the 1-policy condition are removed as adaptivity is not useful in these cases. 

The above-mentioned conditions are abbreviated based on three properties: (i) adaptivity, either \textbf{Adaptive} or \textbf{Unadaptive}; (ii) the base-learner, either \textbf{P(R)PO} or \textbf{D(R)QN}, where PRPO is used to indicate a PPO learner with LSTM policy; and (iii) the number of policies followed by \textbf{P}. For instance, AdaptivePPO14P indicates a 14-policy PPO learner with adaptive policy selection.

The performance evaluation experiments are conducted on the IRIDIS4 supercomputer \citep{IRIDIS} using a single Intel Xeon E5-2670 CPU (2.60GHz) with a varying upper limit to RAM proportional to the number of policies (approximately $2N_{\pi}\SI{}{GB}$). Each run lasts for 2 days for the 27-task MDP domain and 6-9 days for the 18-task POMDP domain. The code for the experiments is available at \url{https://github.com/bossdm/LifelongRL}.

After the performance demonstration, a further analysis then investigates three hypotheses to support the understanding of the Lifetime Policy Reuse and the importance of task capacity.

The first hypothesis is that adaptivity in the policy selection is beneficial if a particular policy is situated in a bad region of parameter space, in which case selecting one of the alternative policies may rapidly find a favourable location of the parameter space. This hypothesis is assessed by making use of the policy spread of learners and relating it to the performance benefit of adaptivity (if any).

The second hypothesis is that task capacity pre-selection is important in the sense that it can be used together with Lifetime Policy Reuse with the following benefits: a) it improves performance and learning; b) it reduces the memory cost; and c) it generalises towards related task spaces. While aspect a) is assessed based on the performance evaluation and the corresponding learning metrics, aspect b) and c) require additional experiments. To assess aspect b), the proposed approach of using a fixed policy library is compared to the growing policy library used in traditional policy reuse methods across different settings of the task capacity, the probability of accepting a policy into the policy library, and the number of task-blocks until convergence. To assess aspect c), additional experiments are conducted on a 125-task Cartpole MDP domain which is based on the same three task-features within the same range as the 27-task domain but with more values included per task-feature.

The third hypothesis is that PPO and DQN are subject to the same representational limits of the theoretical task capacity but due to the more aggressive updates, DQN will push these limits more rapidly; thereby learning more rapidly but also forgetting more rapidly. Additional experiments on the 27-task Cartpole domain investigate the effect of increasing task-block sizes and relaxing the clipped objective as evidence for the hypothesis. To rule out an alternative hypotheses, experiments with alternative experience replay methods are conducted, including modifying the replay buffer through distribution matching, task-matching, and larger buffer sizes.

\section{Performance evaluation and analysis}
\label{sec: performance}
\begin{figure}[htbp!]
\centering
\subfigure[27-task Cartpole domain]
{
\includegraphics[width=0.40\textwidth]{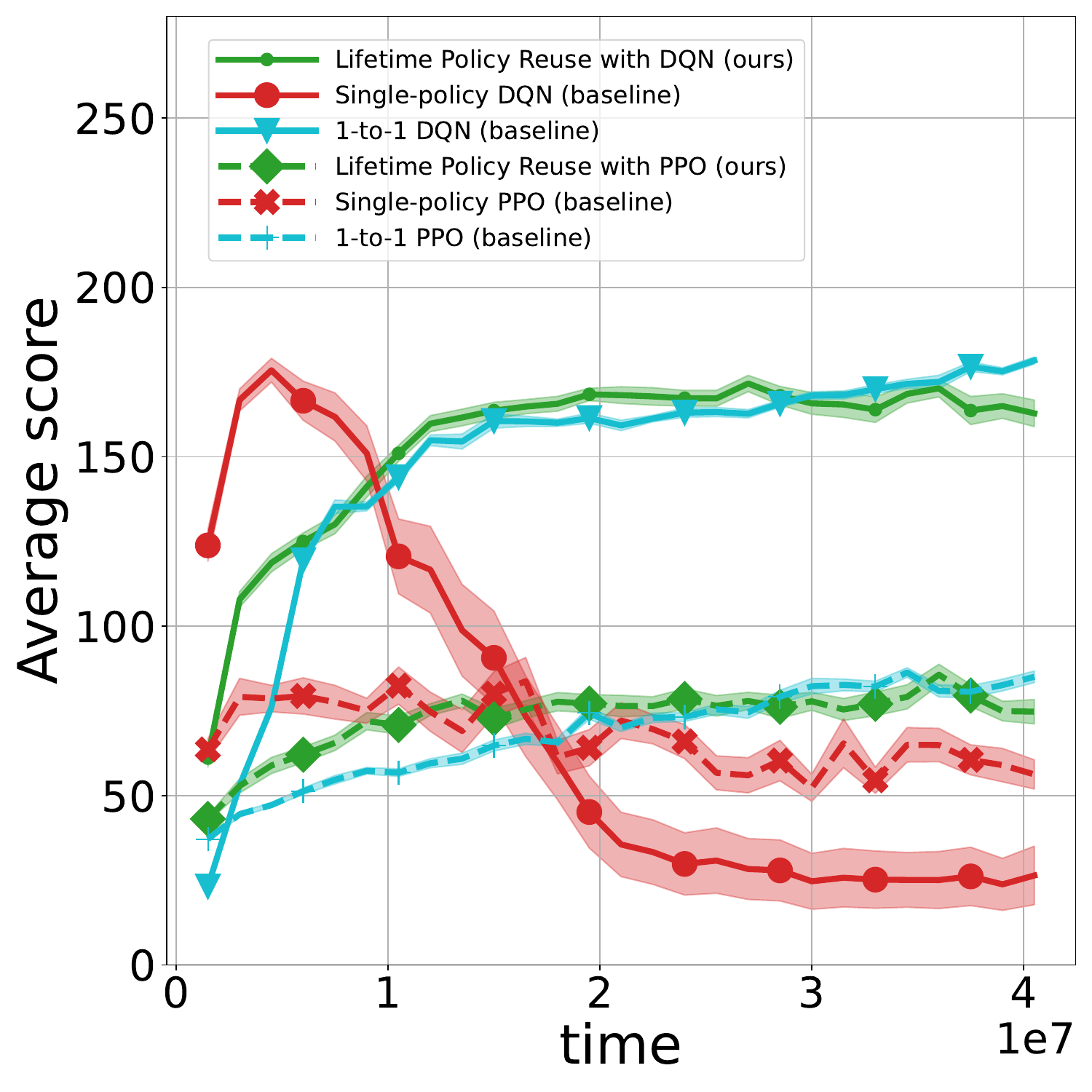}
}
\subfigure[18-task POcman domain]
{
\includegraphics[width=0.40\textwidth]{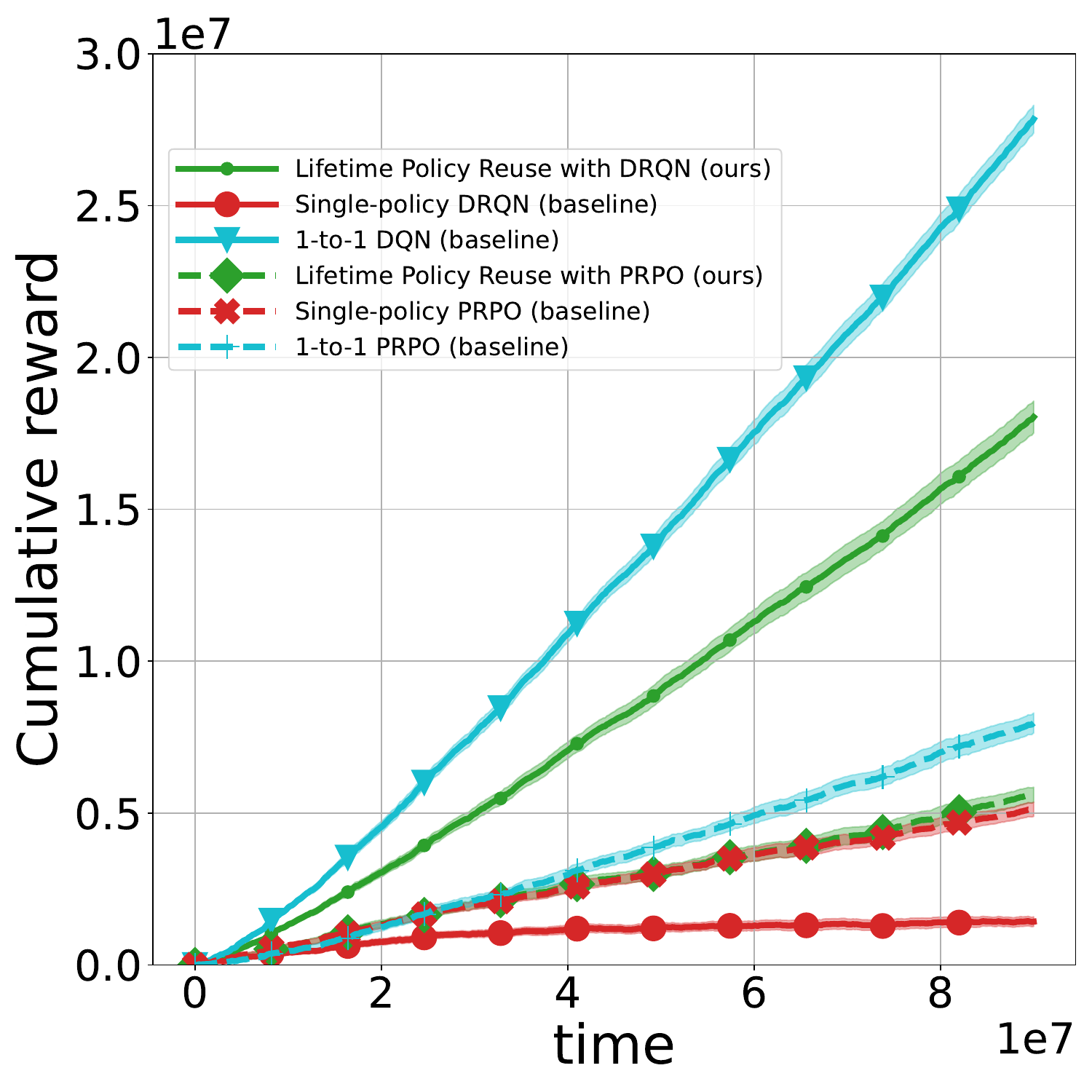}
}
\caption[Performance evaluation of Lifetime Policy Reuse]{Performance evaluation of Lifetime Policy Reuse compared to single-policy and 1-to-1 policy baselines. \textbf{a)} Results for the 27-task Cartpole domain, plotting the average score (Mean $\pm$ Standard Error) across 25 consecutive task-blocks, or every 1,500,000 time steps, as a function of the number of time steps. \textbf{b)} Results for the 18-task POcman domain, plotting the cumulative reward as a function of the number of time steps. The number of policies for Lifetime Policy Reuse is set based on the best available task capacity metric, namely: in the Cartpole domain, $N_{\pi}=9$ for PPO and DQN based on the theoretical task capacity; in the POcman domain, $N_{\pi}=9$ for DRQN and $N_{\pi}=2$ for PPO based on the empirical task capacity with $\epsilon_{c}=0.30$.} \label{fig: LPRcomp}
\end{figure}

Having defined the experimental conditions, this section provides the experimental results, including the performance evaluation for the 27-task Cartpole and 18-task POcman as well as a further analysis. Before providing detailed results with ablations and parametric studies, Figure~\ref{fig: LPRcomp} provides an initial demonstration of the effectiveness of Lifetime Policy Reuse: despite using significantly fewer policies, Lifetime Policy Reuse provides comparable performance levels to the 1-to-1 policy setting for both DQN and PPO base-learners. 

\subsection{27-task Cartpole MDP domain}
\label{sec: cartpole}
\begin{figure}[htbp!]
\centering
\subfigure[DQN]
{
\includegraphics[width=0.40\textwidth]{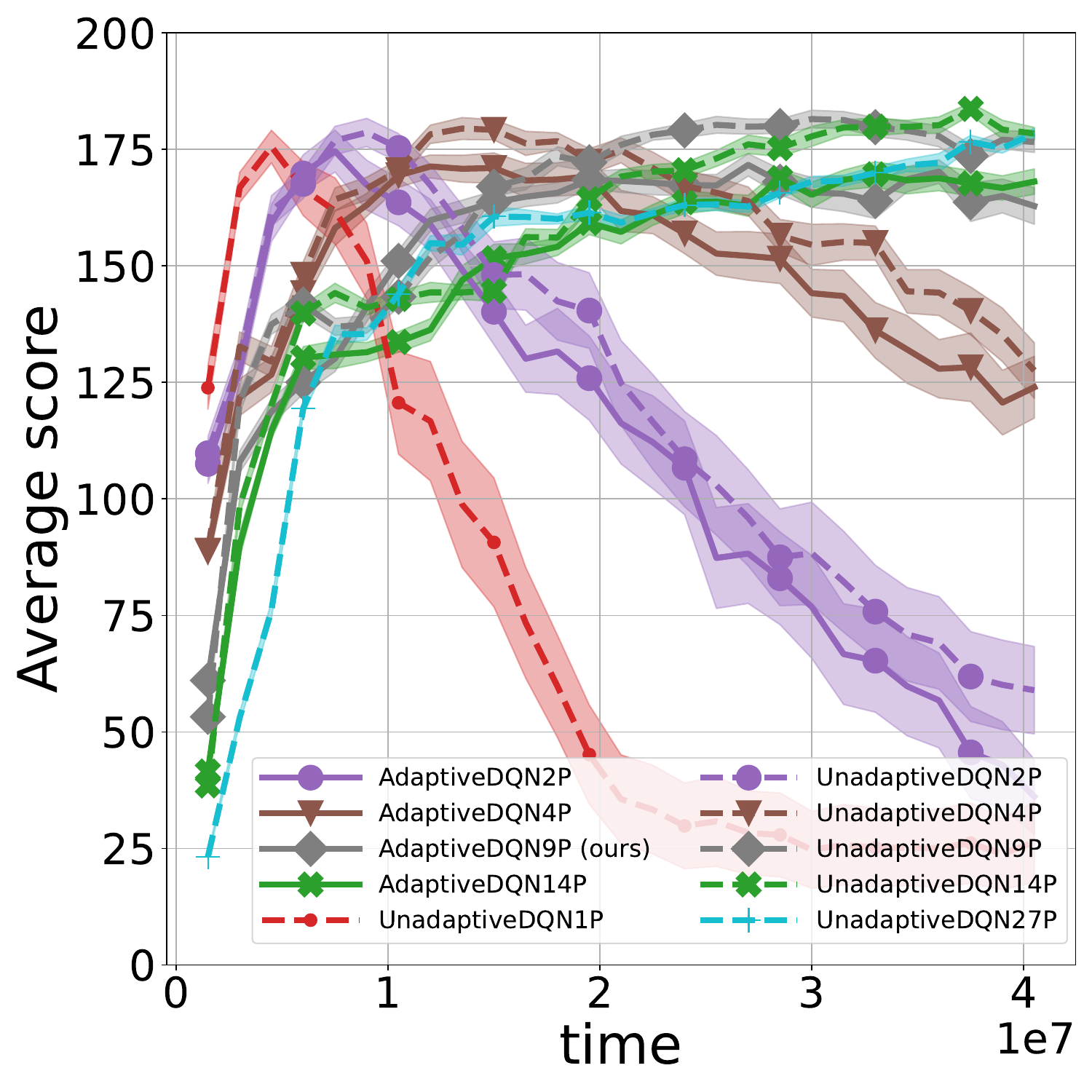}
}
\subfigure[PPO]
{
\includegraphics[width=0.40\textwidth]{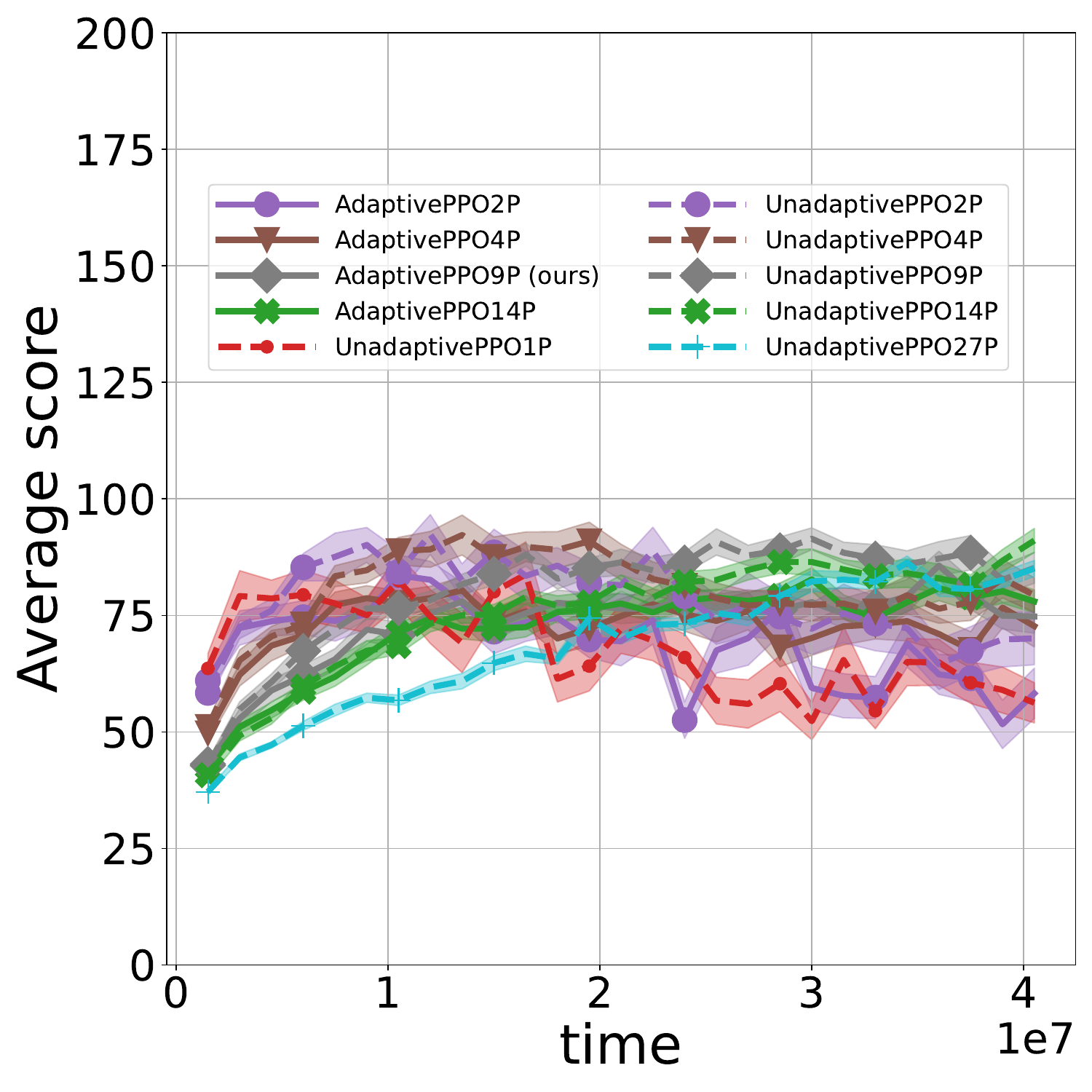}
}
\caption[Average score depending on the number of policies]{Average score (Mean $\pm$ Standard Error) in the 27-task Cartpole MDP domain for different multi-policy variants of the DQN and PPO base-learners. ``ours'' indicates the proposed Lifetime Policy Reuse algorithm with its $\epsilon$-greedy policy selection and a number of policies indicated by the task capacity. To demonstrate the importance of selecting a suitable number of policies and the effectiveness of the $\epsilon$-greedy policy selector, the number of policies is manipulated and the $\epsilon$-greedy policy selector is compared to a fixed policy selector which balances task assignments equally across policies. The inclusion vs exclusion of $\epsilon$-greedy is indicated by Adaptive vs Unadaptive while the number of policies is indicated by the suffix (e.g. \textbf{4P} for 4 policies). The average score is averaged every 25 consecutive task-blocks, or every 1,500,000 time steps. } \label{fig: policiesscore}
\end{figure}

The development of the average score in the 27-task Cartpole MDP domain is shown in Figure \ref{fig: policiesscore} for different settings of the number of policies. The following main trends are observed:
\begin{itemize}
\item For DQN, settings of the number of policies $N_{\pi} \geq 9$ confer a stable upward curve over time, while settings $N_{\pi} \leq 4$ demonstrate a rapid increase at first but then have  degrading  performance over time. With the optimal performance being 200, settings of $N_{\pi} \geq 9$ converge to an optimal or near-optimal performance, accounting for the need for exploratory actions.
\item For PPO, settings of the number of policies $N_{\pi} \geq 4$ confer a stable upward curve over time towards a high but suboptimal performance between 75 and 100; within this group of settings, $N_{\pi} = 27$ reaches a similar final performance but is considerably slower to converge. Settings $N_{\pi} \leq 2$ demonstrate a degrading performance over time, although the effects are much less pronounced compared to DRQN. 
\item Overall DQN demonstrates rapid and pronounced performance improvements while PPO learns more slowly but more stably.
\item Adaptivity has a small negative effect on the performance.
\end{itemize}
The average and final performances are summarised in Table \ref{tab: significance-27task}. For DQN, the best performances are reached by 14- and 27-policy settings, with an average lifetime performance of around 150 and a final performance of 170--180.  The 1-to-1 27-policy significantly outperforms the conditions where $N_{\pi} \leq 2$.\footnote{All significance values reported in this paper are based on pair-wise $F$-tests between the conditions, and $p<0.01$ is considered as statistically significant.} However, UnadaptiveDQN4P and settings of 9 or 14 policies, except the AdaptiveDQN14P ($p=0.994$),  significantly outperform the 1-to-1 27-policy learner. For PPO, all the Unadaptive conditions outperform the 1-to-1 27-policy learner; this effect is significant for all Unadaptive conditions except UnadaptivePPO1P, but not significant for Adaptive conditions ($p$ ranging from 1.0 for $N_{\pi}=2$ to 0.02 for $N_{\pi}=14$). PPO's 1-policy approach performs lower than the 1-to-1 policy but not significantly so ($p=0.763$).

\begin{table}[htbp!]
\centering
\caption[Performance as a function of the number of policies]{Performance in the 27-task Cartpole MDP domain as a function of the number of policies and the inclusion of adaptivity with the $\epsilon$-greedy policy selection. The \textbf{lifetime} average score, and the \textbf{final} score, averaged across the last 10 task-blocks, or the last 600,000 time steps, are shown as the Mean $\pm$ Standard-deviation across the 27 independent runs. ``ours'' indicates the proposed Lifetime Policy Reuse algorithm with its $\epsilon$-greedy policy selection and a number of policies indicated by the task capacity. Scores are sampled once every 60,000 time steps.} \label{tab: significance-27task}
\subtable[DQN]
{
{\scriptsize
\begin{tabular}{l l l }
                          & \multicolumn{2}{l}{\textbf{Performance}} \\ \hline
\textbf{Method}& \textit{lifetime} & \textit{final}  \\ \hline
AdaptiveDQN2P& $109.2 \pm 31.9$ & $34.7 \pm 41.4$  \\ 
AdaptiveDQN4P& $147.5 \pm 16.1$ & $121.3 \pm 35.6$ \\ %
AdaptiveDQN9P (ours) & $153.9 \pm 7.5$ & $163.7 \pm 21.8$  \\ %
AdaptiveDQN14P& $147.9 \pm 5.6$ & $169.7 \pm 14.6$ \\ %
UnadaptiveDQN1P& $71.8 \pm 35.9$ & $25.1 \pm 44.1$ \\ %
UnadaptiveDQN2P& $118.8 \pm 31.1$ & $56.2 \pm 47.6$ \\ %
UnadaptiveDQN4P& $156.2 \pm 9.3$ & $123.2 \pm 35.3$ \\ %
UnadaptiveDQN9P& $161.5 \pm 3.3$ & $176.0 \pm 12.7$ \\ %
UnadaptiveDQN14P& $155.8 \pm 2.8$ & $179.5 \pm 9.3$\\ %
UnadaptiveDQN27P& $147.9 \pm 2.6$ & $179.5 \pm 8.4$\\ %
\hline
\end{tabular}
}
}
\subtable[PPO]
{
{\scriptsize
\begin{tabular}{l l l }
                          & \multicolumn{2}{l}{\textbf{Performance}} \\ \hline
\textbf{Method}& \textit{lifetime} & \textit{final} \\ \hline
AdaptivePPO2P& $68.6 \pm 6.6$&  $55.5 \pm 28.4$ \\
AdaptivePPO4P& $72.5 \pm 9.6$ & $71.0 \pm 23.4$\\
AdaptivePPO9P (ours)& $72.9 \pm 7.7$ & $73.5 \pm 23.2$\\
AdaptivePPO14P& $72.1 \pm 4.7$   & $80.1 \pm 13.7$ \\
UnadaptivePPO1P& $67.7 \pm 9.1$ & $58.4 \pm 38.8$\\
UnadaptivePPO2P& $78.1 \pm 9.9$  & $63.5 \pm 31.8$\\
UnadaptivePPO4P& $80.2 \pm 6.0$  & $77.1 \pm 26.5$\\
UnadaptivePPO9P& $80.5 \pm 4.3$ & $84.1 \pm 21.0$ \\
UnadaptivePPO14P& $75.5\pm 4.2$ & 	$90.5 \pm 19.0$\\
UnadaptivePPO27P& $68.5 \pm 2.6$ & $84.5 \pm 11.6$\\ \hline
\end{tabular}
}
}
\end{table}

\begin{figure}[htbp!]
\centering
\subfigure[DRQN]
{
\includegraphics[width=0.40\textwidth]{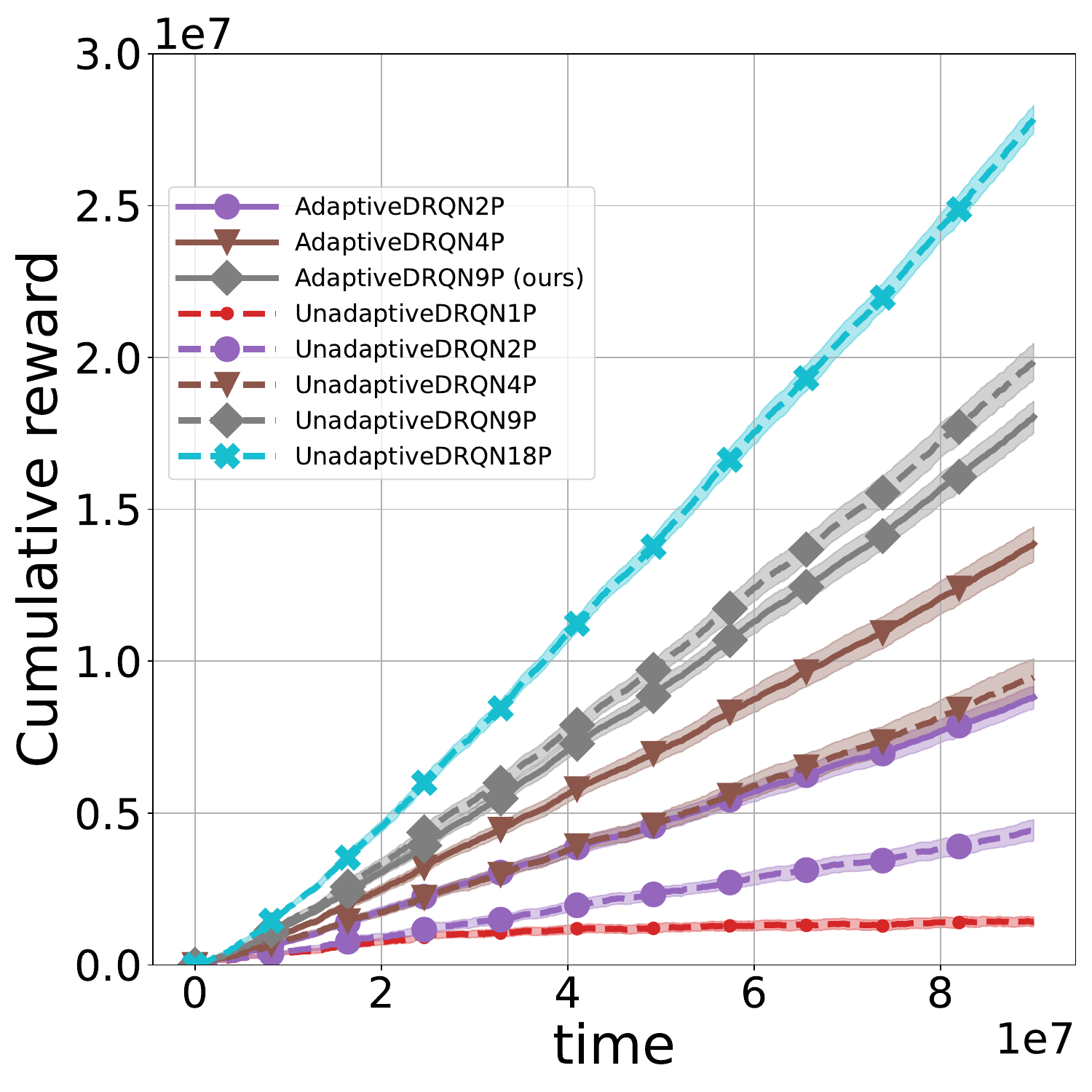}
}
\subfigure[PPO]
{
\includegraphics[width=0.40\textwidth]{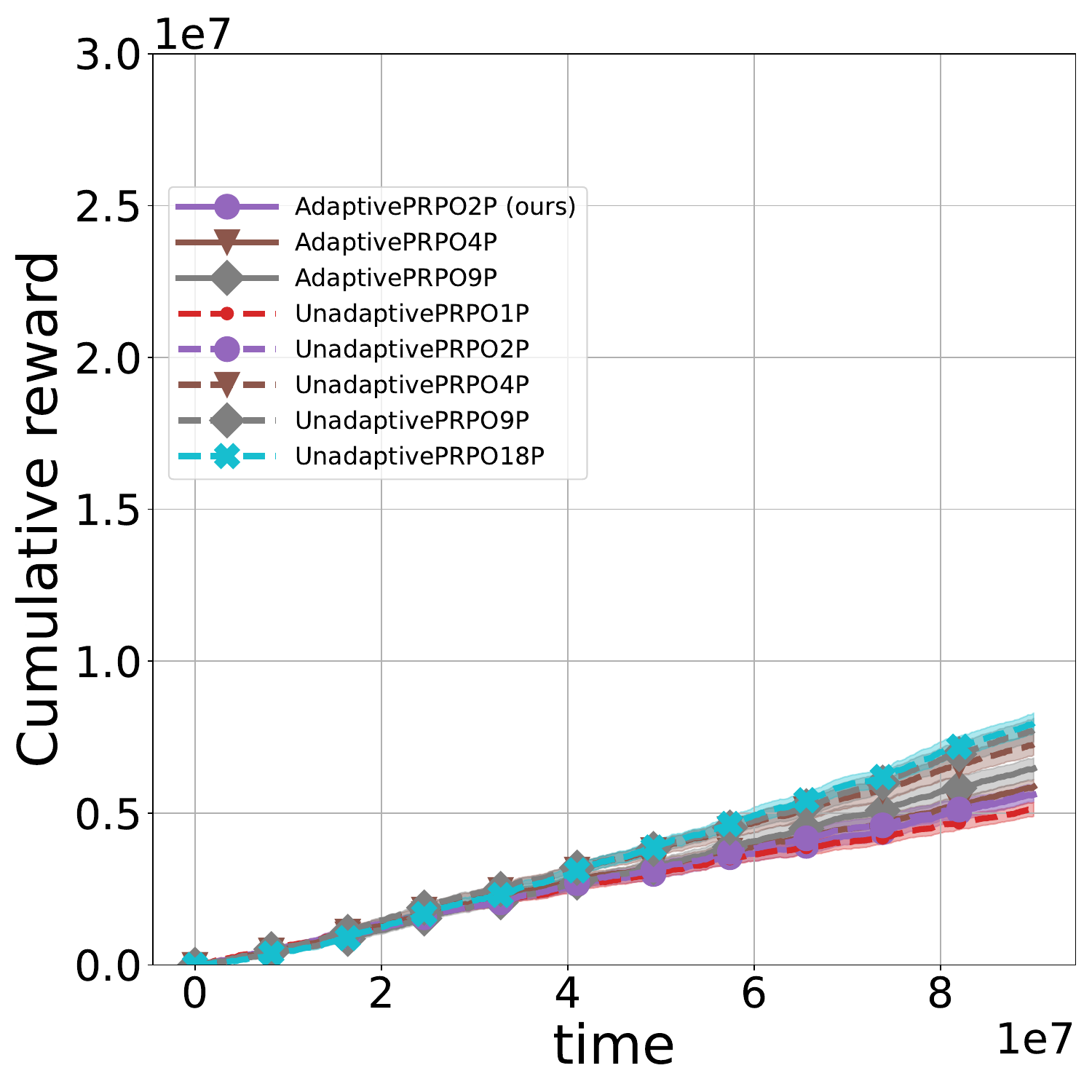}
}
\caption[Cumulative reward depending on the number of policies]{Cumulative reward (Mean $\pm$ Standard Error) in the 18-task POcman POMDP for different multi-policy variants of the DQN and PPO base-learners. ``ours'' indicates the proposed Lifetime Policy Reuse algorithm with its $\epsilon$-greedy policy selection and a number of policies indicated by the task capacity. To demonstrate the importance of selecting a suitable number of policies and the effectiveness of the $\epsilon$-greedy policy selector, the number of policies is manipulated and the $\epsilon$-greedy policy selector is compared to a fixed policy selector which balances task assignments equally across policies. The inclusion vs exclusion of $\epsilon$-greedy is indicated by Adaptive vs Unadaptive while the number of policies is indicated by the suffix (e.g. \textbf{4P} for 4 policies).} \label{fig: policiesscorePOMDP}
\end{figure}

\subsection{18-task POcman POMDP domain}
\label{sec: POcman}
The development of the cumulative reward in the 18-task POMDP POcman domain is shown in Figure \ref{fig: policiesscorePOMDP} for different settings of the number of policies. Two differences with the 27-task MDP domain are the positive effect of adaptivity and the 1-to-1 policy being much higher-performing than settings with 2 or 3 tasks per policy. However, the other trends are comparable to the previous domain:
\begin{itemize}
\item The cumulative reward intake improves as the number of policies is increased, with the 18-policy algorithm performing best in both DRQN and PPO.
\item For DRQN, increasing the number of policies substantially increases the performance and the single policy's performance even decreases over time.
\item For PPO, all learners are able to improve over the lifetime and the differences between conditions are less pronounced than for DRQN; PPO's single policy performance is better but its multi-policy is worse than the corresponding DRQN condition.
\item DRQN's best-performing setting, here $N_{\pi}=18$, is close to the optimal performance, here 45 million in lifetime cumulative reward, considering the need for exploratory actions.\footnote{An upper bound for the optimal performance is achieving a reward of 1 in tasks with $\texttt{R}=1$ and a reward of 0 in tasks with $reward=-1$. With tasks being equally split in $\texttt{R} \in \{-1,1\}$, this implies for a lifetime of 90 million steps that the theoretical upper bound on the lifetime cumulative reward is 45 million.}
\end{itemize}

The average and final performances are summarised in Table \ref{tab: significance}. For DRQN, the single policy has a lifetime average performance of $0.016$, and including multiple policies improves the lifetime average performance monotonically with increasing number of policies, with a factor ranging between 3, for the unadaptive 2-policy learner with its performance of $0.049$, and  20, for the 18-policy learner with its performance of $0.309$.  All learners with multiple policies, including both adaptive and unadaptive conditions and all settings of $N_{\pi} \in \{2,4,9,18\}$, are able to improve on the lifetime average performance with significant effects.
A higher number of policies is also beneficial for PPO but it has a smaller effect: all conditions have a performance ranging between $0.60$-$0.90$, which is 1--1.5 times the performance of the single policy condition, even though some effects are significant (e.g. compare unadaptive policy selection with high versus low number of policies). The single policy PPO has four times the lifetime average performance of the single policy DRQN but multi-policy DRQN conditions have 2--3 times the lifetime average performance of multi-policy PPO conditions.

The final performance yields results similar to the lifetime average, with two notable exceptions. First, the variability is much higher, resulting in higher $p$-values and a lower confidence in the results. Second, the single policy approach in DRQN deteriorates strongly, down to $0.002$, despite the other final performances being in the range of $0.072$, for the unadaptive 2-policy, to $0.323$, for the 18-policy DRQN approach. This means that the improvement obtained by including multiple policies is between 35 and 133 times in performance. By contrast, the single policy approach for PPO has not deteriorated and scores similarly to its lifetime average, $0.055$.

With 2--4 policies, adaptive DRQN significantly outperforms unadaptive DRQN; however, with 9 policies, there is no significant difference. Adaptivity can have a beneficial effect comparable to increasing the number of policies: although comparisons of a larger to a smaller setting of $N_{\pi}$ are significant, the AdaptiveDRQN2P, with its lifetime average of $0.098$, does not show a significant difference to the UnadaptiveDRQN4P, with its lifetime average of $0.105$ ($p = 0.532$); similarly, for the final performance, AdaptiveDRQN4P is not significantly different from UnadaptiveDRQN2P whereas UnadaptiveDRQN4P vs UnadaptiveDRQN9P does give a significant difference.

\begin{table}[htbp!]
\centering
\caption[Performance as a function of the number of policies]{Performance (Mean $\pm$ Standard-deviation) in the 18-task POcman POMDP domain as a function of the number of policies and the inclusion of adaptivity with the $\epsilon$-greedy policy selection. Two performance metrics, based on the cumulative reward function $R(t)=\sum_{\tau=0}^{t} r_{\tau}$,  are shown as the Mean $\pm$ Standard-deviation across the 18 independent runs: the \textbf{lifetime average reward}, defined as $R(T)/T$, where $T=9*10^{7}$ is the total lifetime of the learner; and the \textbf{final average reward}, defined as $\frac{R(T)-R(T-t)}{t}$, where $T=9*10^{7}$ and $t=2*10^{6}$. ``ours'' indicates the proposed Lifetime Policy Reuse algorithm with its $\epsilon$-greedy policy selection and a number of policies indicated by the task capacity. } \label{tab: significance}
\subtable[DRQN]
{
{\scriptsize
\begin{tabular}{l l l}
                          & \multicolumn{2}{l}{\textbf{Performance}} \\ \hline
\textbf{Method}& \textit{lifetime} & \textit{final}  \\ \hline
AdaptiveDRQN2P& $0.098 \pm 0.017$ & $0.104 \pm 0.092$\\
AdaptiveDRQN4P& $0.154 \pm 0.027$ & $0.160 \pm 0.104$\\
AdaptiveDRQN9P (ours) & $0.200 \pm 0.025$ & $0.221 \pm 0.105$\\
UnadaptiveDRQN1P& $0.016 \pm 0.007$ & $0.002 \pm 0.054$\\
UnadaptiveDRQN2P& $0.049 \pm 0.017$ & $0.072 \pm 0.104$\\
UnadaptiveDRQN4P& $0.105 \pm 0.029$ & $0.098 \pm 0.108$\\
UnadaptiveDRQN9P& $0.220 \pm 0.028$ & $0.236 \pm 0.135$\\
UnadaptiveDRQN18P& $0.309 \pm 0.021$ & $0.323 \pm 0.124$\\  \hline
\end{tabular}
}
}
\subtable[PPO]
{
{\scriptsize
\begin{tabular}{l l l}
                                 & \multicolumn{2}{l}{\textbf{Performance}} \\ \hline
\textbf{Method}& \textit{lifetime} & \textit{final}  \\ \hline
AdaptivePRPO2P (ours) & $0.062 \pm 0.012$ & $0.055 \pm 0.081$    \\
AdaptivePRPO4P& $0.065 \pm 0.011$ & $0.070 \pm 0.098$\\
AdaptivePRPO9P& $0.072 \pm 0.017$ & $0.070 \pm 0.088$\\
UnadaptivePRPO1P& $0.057 \pm 0.011$      & $0.055 \pm 0.090$ \\
UnadaptivePRPO2P& $0.063 \pm 0.013$ & $0.071 \pm 0.094$\\
UnadaptivePRPO4P& $0.081 \pm 0.017$  & $0.082 \pm 0.093$\\
UnadaptivePRPO9P& $0.086 \pm 0.018$  & $0.084 \pm 0.116$\\
UnadaptivePRPO18P& $0.088 \pm 0.015$ & $0.107 \pm 0.096$ \\ \hline
\end{tabular}
}
}
\end{table}

\subsection{Analysis}
\label{sec: analysis}
\begin{figure}[htb!]
\centering
\includegraphics[width=0.4\textwidth]{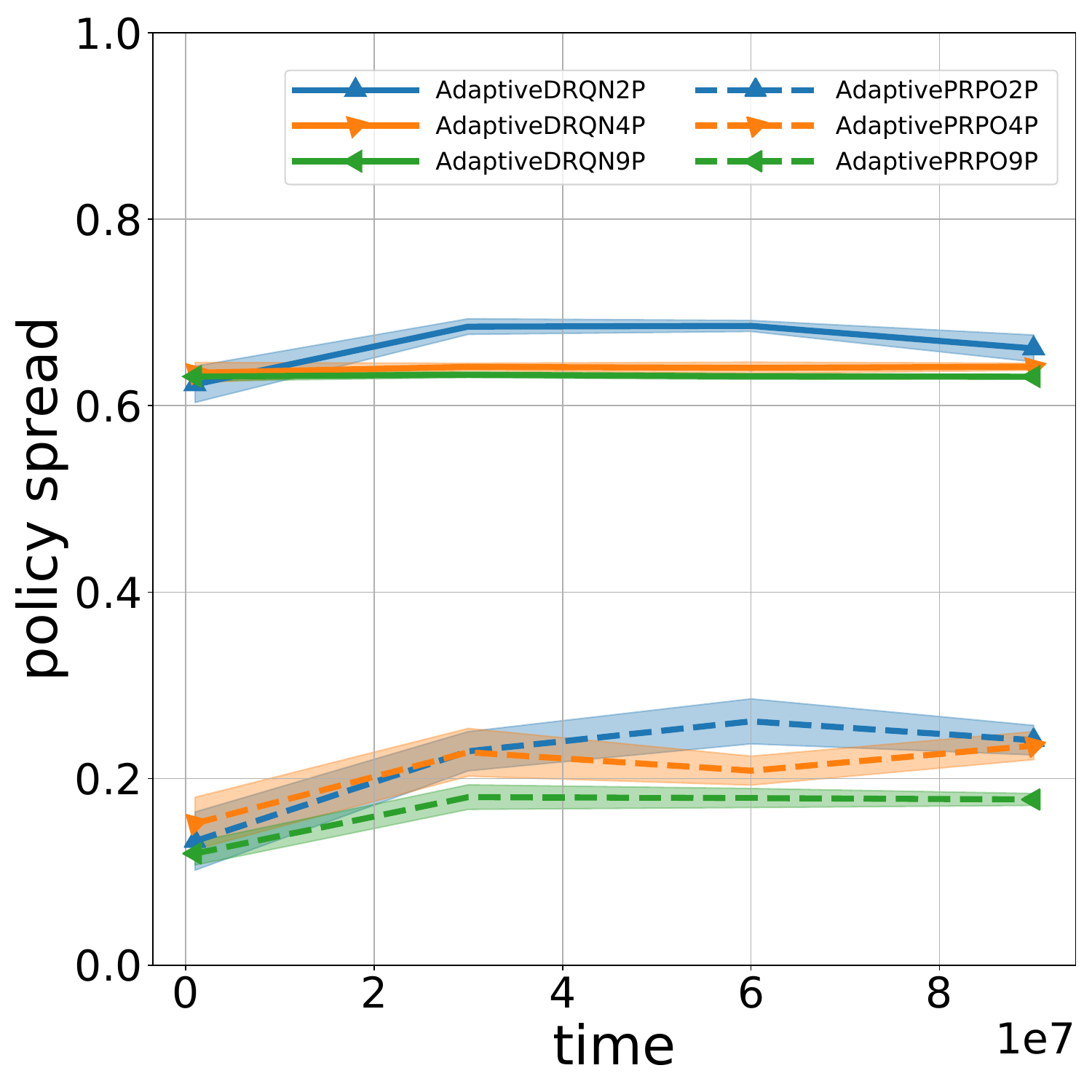}
\caption[Policy spread as a function of time, depending on the number of policies]{Policy spread (Mean $\pm$ Standard Error) in the 18-task POcman POMDP, representing the average variability between the stochastic policies in the policy library. The different base-learners are compared, showing that PPO has more limited variability in its policies compared to DRQN, explaining why DRQN gains more performance from adaptive policy selection.} \label{fig: paramdiversity}
\end{figure}

This section further analyses three core hypotheses to explain the above results, illustrating Lifetime Policy Reuse and the importance of task capacity.
\paragraph{Hypothesis 1: Adaptive policy selection provides escape from low-performing regions of parameter space} 
The first hypothesis is that adaptivity in the policy selection is beneficial if a particular policy is situated in a bad region of parameter space, in which case selecting one of the alternative policies may rapidly find a favourable location of the parameter space. Due to the positive effect of adaptivity only occurring in the DRQN conditions in the 18-task POMDP domain, this hypothesis is supported empirically if DRQN has a higher policy spread than PPO in the 18-task POMDP domain, but not in the 27-task MDP domain. Empirical results support the hypothesis: in the 18-task POMDP domain, adaptive DRQN conditions have a policy spread within $[0.6,0.7]$ while PPO has a policy spread within $[0.1,0.3]$ (see Figure~\ref{fig: paramdiversity}); in the 27-task MDP domain, adaptive DRQN conditions have a policy spread within $[0.3,0.4]$ while PPO has a policy spread within $[0.4,0.45]$. Comparing the two domains, the 18-task POMDP domain, which has opposing dynamics such as approach and avoid behaviours, may have a lower task-similarity compared to the 27-task MDP domain, which has more behavioural constants such as moving in the direction to which the pole tilts; this implies the required policies for 18-task POMDP are highly distinct, and continuing to train an initially unsuccessful policy will be less efficient than switching policy.

\paragraph{Hypothesis 2: the importance of task capacity}
The importance of task capacity is now demonstrated on three metrics, showing that task capacity pre-selection a) improves performance and learning metrics; b) reduces the memory cost; and c) generalises towards related task spaces, such that the same number of policies can be used with the same performance benefits. To this end, the analysis below compares settings of the number of policies above task capacity to settings below task capacity.

\textit{a) Improved performance and learning metrics:} In the Cartpole domain, the theoretical task capacity for deep reinforcement learners  has been estimated as $C=3.4$ (see final paragraphs of Section \ref{sec: theoretical-taskcapacity}), corresponding to a minimum of required policies $N^{\pi}=8$. The following benefits of pre-selection, using $N_{\pi} = 9$ just above the requirement, are found for DQN:\\
\begin{itemize}
\item Improved performance: with 9 or more policies, the performance converges to the highest level for the base-learner; with 1-, 2-, and 4-policy learners, a degrading performance is observed towards the end of the run. These findings are consistent with the hypothesis from the theoretical task capacity that the representational limit of the policies is being exceeded.
\item Improved DQN forgetting ratio: with 9 or more DQN policies, the forgetting ratio is near-zero; with 1-, 2-, and 4-policy DQN learners have a degrading forgetting ratio between $-0.5$ and $-1$ as more interfering task-blocks are presented, with $-1$ for the 1-policy DQN for 30 or more interfering task-blocks. This is consistent with forgetting being primarily due to representational overlap.
\item Marginally improved PPO forgetting ratio: for PPO, similar trends are demonstrated except that the differences are much smaller.
\end{itemize}
 A further observation is that while DQN displays a higher transfer ratio than PPO for low-policy settings, this effect does not improve performance because it is negated by DQN's stronger forgetting. In the POcman domain, a theoretical task capacity estimate is not available; however, it can be observed that the empirical task capacity is lower ( $C_{\text{emp}}=4.5$ for PPO and  $C_{\text{emp}}=2$ for DRQN) and strong forgetting and negative transfer trends are observed. Additional analyses of empirical task capacity can be found in Appendix B. 

\textit{b) Reduced memory cost:} The analysis of memory cost considers the growth of the number of policies as a function of (i) the task capacity; (ii) the number of task-blocks until convergence; and (iii) and the acceptance probability, defined as the probability that newly proposed policies are accepted into the policy library. The analysis fixes the number of tasks to $N_{\tau}=1000$ and makes a few simplifying assumptions, namely that the acceptance probability is fixed and that the number of policies required is equal to that in lifetime policy reuse. Figure~\ref{fig: memory} in Appendix B demonstrates that a higher task capacity, a larger number of task-blocks until convergence, and a higher acceptance probability are associated with larger increases in RAM when comparing traditional policy reuse to fixing the number of policies to the task capacity. Even though the maxima chosen, namely task capacity $C=5$ and task-blocks $T_c=4$, are still relatively low, the increase is between three- and five-fold, depending on how frequently new policies are accepted in the library. Due to the larger number of task-blocks until convergence, traditional policy reuse will store the temporary policy for one task while learning another task, whereas Lifetime Policy Reuse can maintain the fixed number of policies given by the task capacity. 

\begin{figure}[htbp!]
\centering
\subfigure[DQN]
{
\includegraphics[width=0.40\textwidth]{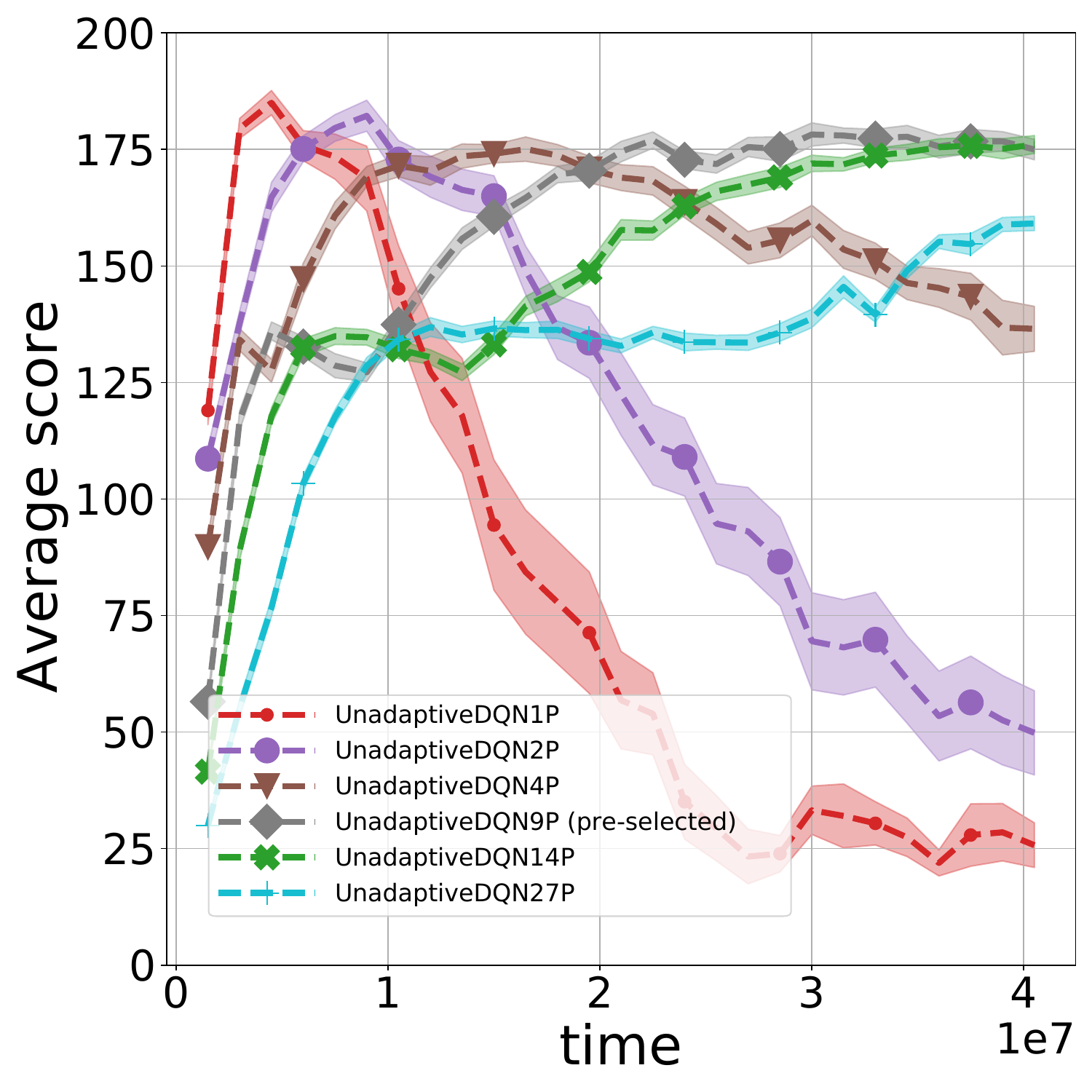}
}
\subfigure[PPO]
{
\includegraphics[width=0.40\textwidth]{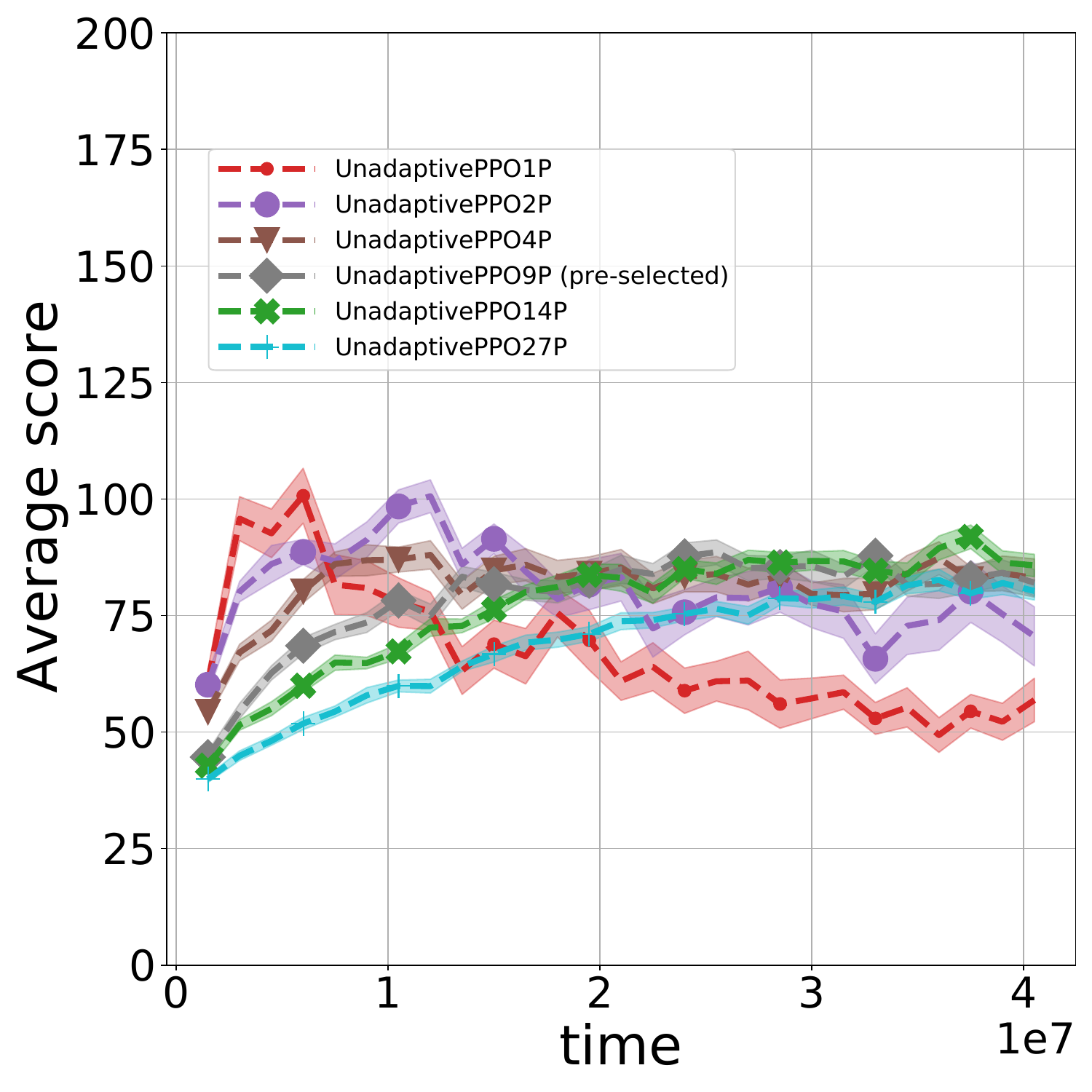}
}
\caption{Average score (Mean $\pm$ Standard Error) in the 125-task Cartpole MDP domain for different multi-policy variants of the DQN and PPO base-learners. The pre-selected number of policies, as indicated by \textbf{9P} for 9 policies and the ``pre-selected'' suffix, is among the top performers across different settings of the number of policies, indicating that one can set the number of policies based on smaller but related task domains. The average score is averaged every 25 consecutive task-blocks, or every 1,500,000 time steps.}\label{fig: 125-task}
\end{figure}

\textit{c) Generalisation to 125-task domain}
The hypothesis that the pre-selected number of policies generalises across related tasks spaces is investigated below by an empirical demonstration on a 125-task Cartpole MDP domain. The domain has the same three task-features within the same range as the 27-task domain but with more values included per task-feature. That is, (i) the mass of the cart varies in $\{0.5,0.75,1.0,1.5, 2.0\}\SI{}{kg}$; (ii) the mass of the pole varies in $\{0.05,0.075,0.1,0.15,0.2\}\SI{}{kg}$; and (iii) the length of the pole varies in $\{ 0.5,0.75,1.0,1.5,2.0\}\SI{}{m}$. The results for DQN and PPO confirm the above hypothesis. First, similar to the demonstration for the theoretical task capacity, the same performance trends are observed (see Figure \ref{fig: 125-task}), with 9 policies being sufficient for a smooth upward and converging curve and 4 or fewer policies appearing to induce representational overlap. Second, for $\epsilon_c = 0.05$ as in the empirical task capacity results in Appendix B, the relative $\epsilon_c$-optimality of the lifetime average performance with respect to the 27-policy learner is provided for both base-learners by the same setting as in the 27-task domain: for DQN, this is a setting of 4 or more policies, as the performance is $80.3 \pm 21.7$ for 1 policy, $116.3 \pm 26.8$ for 2 policies, $154.8 \pm 8.9$ for 4 policies, $158.0 \pm 3.6$ for 9 policies, $147.5 \pm 3.2$ for 14 policies, and $128.4 \pm 3.2$ for 27 policies; for PPO, this is a setting of 1 or more policies, as the performance is $67.0 \pm 8.0$ for 1 policy, $80.6 \pm 7.9$ for 2 policies, $81.4 \pm 5.7$ for 4 policies, $78.6 \pm 3.8$ for 9 policies, $76.8 \pm 3.3$ for 14 policies, and $68.6 \pm 2.5$ for 27 policies.

\paragraph{Hypothesis 3: DQN and PPO performance difference explained by unbounded updates vs monotonic improvement}
A third and last hypothesis is that the performance results are explained by the difference between unbounded updates of DQN compared to the clipped updates of PPO for monotonic improvement. That is, PPO and DQN are subject to the same representational limits of the theoretical task capacity but due to the more aggressive updates, DQN will push these limits more rapidly. The empirical results  (see section on DQN vs PPO in Appendix B) are consistent with the hypothesis. First, increasing task-block sizes and relaxing the clipped objective demonstrates that the monotonic improvement objective of PPO slows down the learning progress on a single task but helps to maintain performance across many sequentially presented tasks. Second, modifying the replay buffer, through distribution matching, task-matching, and larger buffer sizes, does not reduce DQN's forgetting metrics, thereby ruling out an alternative hypothesis.

Average score (Mean $\pm$ Standard Error) in the 27-task Cartpole MDP domain for different multi-policy variants of the DQN and PPO base-learners. ``ours'' indicates the proposed Lifetime Policy Reuse algorithm with its $\epsilon$-greedy policy selection and a number of policies indicated by the task capacity. To demonstrate the importance of selecting a suitable number of policies and the effectiveness of the $\epsilon$-greedy policy selector, the number of policies is manipulated and the $\epsilon$-greedy policy selector is compared to a fixed policy selector which balances task assignments equally across policies. The inclusion vs exclusion of $\epsilon$-greedy is indicated by Adaptive vs Unadaptive while the number of policies is indicated by the suffix (e.g. \textbf{4P} for 4 policies). The average score is averaged every 25 consecutive task-blocks, or every 1,500,000 time steps.

\section{Discussion}
\label{sec: discussion}
This paper presents an approach to policy reuse that scales to lifetime usage, in the sense that (a) its number of policies is fixed and does not result in memory issues; (b) its evaluation is based on the lifetime average performance; and (c) its policies are continually refined based on the lifetime of experience with newly incoming tasks. The approach is widely applicable to MDPs and POMDPs and various base-learners, as is illustrated empirically for DQN and PPO base-learners in cartpole and POcman domains. To help select the number of policies required for a domain, theoretical and empirical task capacity metrics are developed and illustrated for a few examples. These metrics are proposed for three purposes, each of which has been shown in this paper to have empirical support: 
\begin{itemize}
\item To pre-select, based on theoretical arguments, the number of policies required on a task sequence. This purpose has been demonstrated by the observation that violating the task capacity requirement leads to negative transfer and forgetting.
\item To select the number of policies required for a lengthy task sequence based on a limited but fairly representative task-sequence. This purpose has been demonstrated by the use of the 27-task Cartpole domain to select the number of policies for the 125-task Cartpole domain. 
\item To benchmark base-learners in terms of their scalability. Having a representation of comparable complexity, D(R)QN and PPO are found to have the same theoretical task capacity. However, they demonstrate observable differences in the empirical task capacity metric due to learning and forgetting at different rates.
\end{itemize}

Lifetime Policy Reuse comes with certain benefits but also certain limitations compared to existing alternatives. Compared to other algorithms, it can be applied when task
sequences involve (i) a rapid succession of randomly presented tasks; and (ii) distinct task clusters each of which allow significant positive transfer within clusters and negative transfer between clusters. Compared to traditional policy reuse \citep{Zheng2018,Wang2018a,Hernandez-Leal2016,Rosman2016,Li2018a,Li2018b,Fernandez2006}, the lifetime set-up allows online learning without growing the policy library, requiring policies to converge on the task, or modifications to the base-learner's usual learning algorithm. Lifetime Policy Reuse continually refines its existing policies, overcoming a downside of offline methods for policy reuse (e.g., \cite{Rosman2016}): since the simulation may not represent the real-world environment or as unexpected problems may come up during the application of interest, no suitable policy will be available in the library.  However, a challenge that follows is to avoid catastrophic forgetting; the results in this paper show that this can be avoided through appropriate selection of the number of policies guided by the task capacity. As in other policy reuse methods, Lifetime Policy Reuse avoids task-specific parameters and task-descriptive features (see e.g., \cite{Deisenroth2014,Li2018,Rostami2020}), allowing policies that are uniquely specialised to subsets of tasks and requiring less expert knowledge on the task domain. Lifetime Policy Reuse similar to other policy reuse algorithms and LRL algorithms using task-specific parameters requires that tasks can be identified with a task index. Various LRL algorithms (e.g. \cite{Abel2018,Brunskill2014}) focus only on learning a new task from various source tasks rather than also considering the problem of retaining knowledge of prior tasks; as such these do not require a task index but applying them as given would imply relearning any task re-occurring in a long task sequence. The requirement of the task index is not strong, since instead of being provided by the environment it could be identified automatically by the agent using task identification. For example, a Hidden Markov Model can predict the task from observations \citep{Kirkpatrick2017} or an observed change in reward distribution can indicate a new task is being presented \citep{Lomonaco2020}.

The performance and adaptivity findings in the Cartpole domain and the POcman domain have significance for RL in challenging lifelong learning domains. For the Cartpole domain, learners with one policy for every three tasks can similarly achieve a near-optimal performance despite the challenging domains that may perhaps be paraphrased as an ``anti-curriculum'': tasks are presented in an unordered sequence, follow in quick succession, and vary in orthogonal dimensions. For the POcman domain, partial observability further has the detrimental effect of making the different tasks more difficult to learn but also more difficult to distinguish from each other.\footnote{Although anti-curricula do not appear too often in structured man-made environments, they tend to appear often in the animal kingdom or in robotic missions, where environmental changes require rapid adaptation with limited if any cues from the environment.} Moreover, due to the presentation of a variety of objects of interest, POcman is challenging since a recent benchmarking study within OpenAI Gym \citep{Brockman2016} and Multiworld \citep{Nair2018} environments concluded that a variety of meta-/multi-task-RL approaches do not generalise well when the object of interest is varied \citep{Yu2019}. Scaling up lifelong RL systems in such domains therefore presents a significant but exciting challenge, and the results provided by Lifetime Policy Reuse indicate a positive step in this direction. Adaptivity-wise, the study shows that bandit learning can provide a performance improvement comparable to a doubling of the number of policies, but only if task requirements and the policies in the library vary widely. While so far, studies incorporating adaptivity over policies have used static policy libraries which are only changed by adding newly converged policies and which are based on Markov Decision Processes, or have used a large amount of policies developed in an off-line stage of learning  \citep{Fernandez2006,Rosman2016,Cully2015}, this paper demonstrates the viability of adaptive policy selection with continual refinement based on the lifetime average performance to solve challenging lifelong learning sequences.

The theoretical task capacity provides a representational argument for the limited number of tasks that can be  solved by a single policy. While a single policy can represent the optimal behaviour required for any one particular task, it can only represent the optimal behaviours for a limited number of tasks due to the conflicting state-action pairs that are required for different tasks. The empirical results show that the theoretical task capacity is one of the key limiting factors. The difficulty of epsilon-optimality over many tasks is illustrated for D(R)QN and PPO as a low-policy PPO could learn all tasks with low accuracy while a high-policy D(R)QN could learn all tasks with high accuracy. PPO's objective function is clipped specifically to stimulate monotonic improvement over time \citep{Schulman2017a}. This slow learning allows time to learn new patterns without erasing previously learned patterns and rather than representing a few tasks optimally, the policy represents a wide set of tasks sub-optimally. By contrast, the D(R)QN objective allows more aggressive weight updates resulting in either a near-optimal performance if a sufficient number of policies allows all tasks to be represented near-optimally, or in strong catastrophic interference and negative transfer otherwise. Further evidence of the importance of theoretical task capacity is that selective experience replay \citep{Isele2018} or other modifications to experience replay do not improve single-policy DQN.

\section{Conclusion}
\label{sec: conclusion}
In lifelong reinforcement learning settings, a challenge is to avoid catastrophic forgetting and to stimulate selective transfer. This paper makes two novel contributions, namely 1) Lifetime Policy Reuse, a model-agnostic policy reuse algorithm that avoids generating many policies as tasks are being added across the lifetime; and 2) measures for task capacity, which help to pre-select a number of policies and analyse the performance of lifelong reinforcement learning systems in terms of the number of tasks a policy can learn or represent. Combining task capacity based pre-selection with Lifetime Policy Reuse demonstrates improved performance, transfer, forgetting, and memory consumption on 18-task POMDP sequences and 27- to 125-task MDP sequences with random temporal structure.

\begin{acks}
This research was funded by the Engineering and Physical Sciences Research Council and by Lloyd's Register Foundation. The authors thank Nicholas Townsend for comments on an early draft of this paper and acknowledge the use of the IRIDIS High Performance Computing Facility and associated support services at the University of Southampton.
\end{acks}

\begin{appendix}
\section{Implementation details}
\subsection{Starting coordinates of the 18-task POcman domain}
Taking $x$ increasing to the east and $y$ increasing to the south, static main objects were initialised randomly from a set of coordinates: $\{(3,3),(5,3)\}$ in the cheese maze; $\{(6,1),(9,1),(9,6)\}$ in Sutton's maze; $\{(1,1),(1,7),(7,1),(7,7)\}$ in the pacman topology. This implies a general search strategy should be used, rather than the memorisation of a single path. Similar to the original tasks, the learner's initial position is $(1,2)$ for the cheese-maze, $(1,3)$ in Sutton's maze, and $(4,7)$ for the pacman task. The initial location of dynamic objects, similar to pacman ghosts, is based on a single home position, which is the central bottom location $(3, 3)$ for the cheese maze, the top-right location $(9, 1)$ for Sutton's maze, and the above-centre location $(4,3)$ for the pacman topology.
\subsection{Base learners}
\sloppy
Code from \url{https://github.com/flyyufelix/VizDoom-Keras-RL/} and \url{https://github.com/magnusja/ppo/} are taken as a template, and then modified to use with multiple policies, and to match the original papers \citep{Hausknecht2015,Schulman2017a}: for DRQN, a slowly changing target value-function; for PPO, the set-up similar to the Atari experiments in \citep{Schulman2017a}, where actions are discrete, non-output layers are shared and the objective include additional terms for the value function and the entropy coefficient. Parameter settings mentioned in Table \ref{tab: parametersMDP} and Table \ref{tab: parameters} are chosen based on the original papers and any other modifications are chosen based on a limited tuning procedure based on partial runs of the lifelong learning domain; this is illustrated for the learning rate hyperparameter of PPO and D(R)QN in Table \ref{tab: learning-rateMDP} and Table \ref{tab: learning-rate}. No annealing of parameters was done to keep the various conditions comparable. To suit the domains where observations have no spatial correlations, convolutional layers are replaced by a single densely connected layer. In partially observable environments, an LSTM tanh layer \citep{Hochreiter1997} is used; otherwise this LSTM layer is replaced by a dense RELU layer with equal number of 80 neurons. The resulting code with all the experiments is available at \url{https://github.com/bossdm/LifelongRL}. 

\begin{table}[htbp!]
\centering
\caption{Parameter settings for the base-learners on the 27-task Cartpole MDP domain. ``All'' denotes settings common to all base-learners.}
\label{tab: parametersMDP}
{\scriptsize
\begin{tabular}{l | l l}
\textbf{Base-learner} & \textbf{Parameter} & \textbf{Setting} \\ \hline
All & hidden layers &  80 (Dense RELU) - 80 (Dense RELU)  \\
	& discount & 0.99 \\
	\hline
DQN & batch size & 10 \\
 & update frequency & once every 4 time steps \\
& replay memory size &  400,000 experiences\\
& optimisation algorithm &  AdaDelta \citep{Zeiler2012}\\
& momentum & 0.95 \\
& learning rate & 0.1 \\
& exploration rate & 0.2\\
& clip gradient & absolute value exceeding 10 \\
& replay start & 50,000 time steps\\
& target update frequency & once every 10,000 time steps \\
\hline
PPO & optimisation algorithm &  Adam \citep{Kingma2015}\\
& learning rate & 0.00025 \\
& batch size & 34\\
& update frequency & once at end of episode \\
& epochs & 10 \\
& clip gradient & norm exceeding 1.0 \\
& GAE parameter & 0.95\\
& clipped objective coefficient & 0.10\\
& value function coefficient & 1\\
& entropy coefficient& 0.01
\end{tabular}
}
\end{table}
\newpage
\begin{table}[htbp!]
\centering
\caption{Final performance (Mean $\pm$ SD of the final 10 episodes) of task-specific policies as a function of the learning rate hyperparameter, after running on each of the 27 tasks in the Cartpole MDP domain for the full length of 1.5 million time steps.} \label{tab: learning-rateMDP}
{\scriptsize
\begin{tabular}{l | l l l l l}
				 & \multicolumn{5}{l}{\textbf{Learning rate}} \\ \hline
\textbf{Learner} & $0.00025$ & $0.001$ & $0.005$ & $0.025$ & $0.1$ \\ \hline
PPO & $96.922 \pm 42.73$ & $67.224 \pm 62.21$ & $17.196 \pm 24.79$ & $12.539 \pm 4.07$ & $10.180 \pm 2.21$  \\
DQN & $11.457 \pm 2.65$ & $27.089 \pm 33.05$ & $78.791 \pm 44.01$ & $150.898 \pm 44.60$ & $176.906 \pm 15.26$ \\
\hline
\end{tabular}
}
\end{table}

\begin{table}[htbp!]
\centering
\caption{Parameter settings for the base-learners on the 18-task POcman POMDP domain. ``All'' denotes settings common to all base-learners.}
\label{tab: parameters}
{\scriptsize
\begin{tabular}{l | l l}
\textbf{Base-learner} & \textbf{Parameter} & \textbf{Setting} \\ \hline
All & unroll \& burn-in & 15 time steps \\
	& hidden layers &  80 (Dense RELU) - 80 (LSTM tanh)  \\
	& discount & 0.99 \\
	\hline
DRQN & batch size & 10\\
 & update frequency & once every 4 time steps \\
& replay memory size &  400,000 experiences\\
& optimisation algorithm &  AdaDelta \citep{Zeiler2012}\\
& momentum & 0.95 \\
& learning rate & 0.1 \\
& exploration rate & 0.2\\
& clip gradient & absolute value exceeding 10 \\
& replay start & 50,000 time steps\\
& target update frequency & once every 10,000 time steps\\
\hline
PRPO & optimisation algorithm &  Adam \citep{Kingma2015}\\
& learning rate & 0.00025 \\
& batch size & 34\\
& update frequency & once every 100 time steps \\
& epochs & 3 \\
& clip gradient & norm exceeding 1.0 \\
& GAE parameter & 0.95\\
& clipped objective coefficient & 0.10\\
& value function coefficient & 1\\
& entropy coefficient& 0.01
\end{tabular}
}
\end{table}

\begin{table}[htbp!]
\centering
\caption{Performance (Mean $\pm$ SD) of task-specific policies as a function of the learning rate hyperparameter, after running on each of the 18 tasks in the POcman POMDP domain for the full length of 5 million time steps.} \label{tab: learning-rate}
{\scriptsize
\begin{tabular}{l | l l l l l}
				 & \multicolumn{5}{l}{\textbf{Learning rate}} \\ \hline
\textbf{Learner} & $0.00025$ & $0.001$ & $0.005$ & $0.025$ & $0.1$ \\ \hline
PRPO & $0.034 \pm 0.03$ & $0.037 \pm 0.03$ & $-0.105 \pm 0.10$ & $-0.128 \pm 0.10$ & $-0.130 \pm 0.10$ \\
DRQN & $-0.047 \pm 0.04$ & $-0.025 \pm 0.03$ & $0.010 \pm 0.03$ & $0.056 \pm 0.04$ & $0.150 \pm 0.06$\\
\hline
\end{tabular}
}
\end{table}

\section{Additional analyses}
\subsection{UCB-based policy selection}
\begin{figure}[htbp!]
\subfigure[DQN]
{
\includegraphics[width=0.30\textwidth]{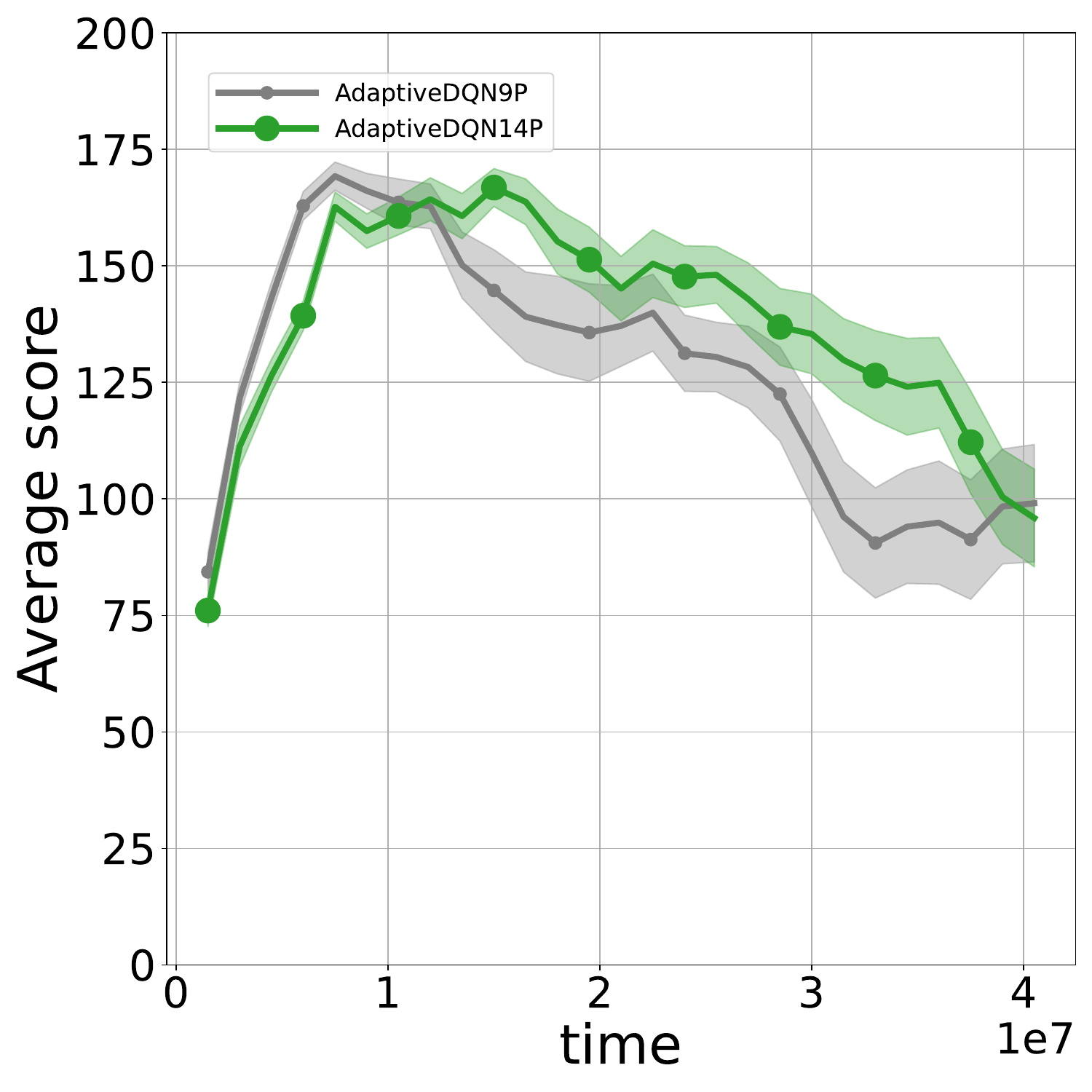}
}
\subfigure[PPO]
{
\includegraphics[width=0.30\textwidth]{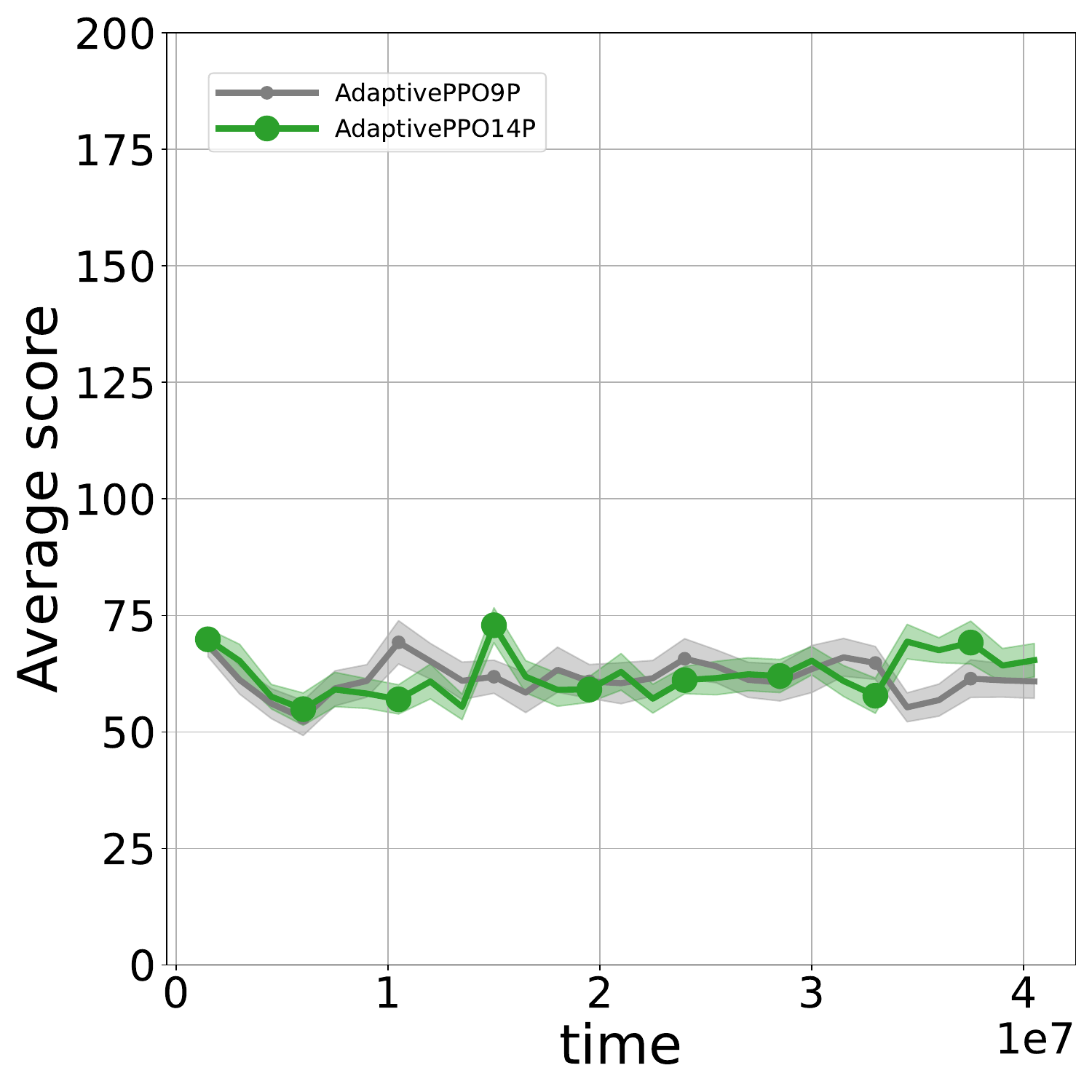}
}
\caption{Average score (Mean $\pm$ Standard Error) of a modified version of Lifetime Policy Reuse with UCB1-Tuned as policy selector in the 27-task Cartpole MDP domain. The average score is averaged every 25 consecutive task-blocks, or every 1,500,000 time steps.}
\label{fig: UCB}
\end{figure}

Additional experiments further investigate whether another bandit algorithm could potentially improve performance. For this purpose, UCB1-Tuned \cite{Auer2002} is selected due to its logarithmic regret, its automatic estimate of the variance upper bound, and the empirical results on stationary distributions demonstrating it to be comparable to or better than a  well-tuned $\epsilon$-greedy algorithm, depending on the distribution. The algorithm selects the policy that is highest-performing according to $\text{UCB1-tuned}(i,j) = \tilde{\mathcal{R}}{ij} + \sqrt{\frac{\ln(n_{\cdot,j})}{n_{i,j}} \min\left(1/4,\text{Var}(i,j)) \right)}$, where $\tilde{\mathcal{R}}{ij}$ is the lifetime average reward of policy $i$ on task $j$ normalised in $[0,1]$ such that $1$ corresponds to the optimal performance on the Cartpole task, $n_{\cdot,j}$ is the total number of policy selections on task $j$, $n_{i,j}$ is the number of policy selections of policy $i$ on task $j$, and $\text{Var}(i,j) = v + \sqrt{2 \ln\left(n_{\cdot,j}\right)}{n_{i,j}}$ based on sample variance $v$ of the normalised cumulative reward across episodes. Results are shown for the Cartpole domain in Figure~\ref{fig: UCB}. The UCB1 algorithm does not perform well and analysis shows that this is due to the algorithm frequently alternating between policies with comparable UCB scores. This result supports the use of epsilon-greedy algorithm with a focus on the controlled exploitation of the current best policy, which is being improved by training on the task, rather than continuing to explore different policies with a high frequency, which prevents specialisation and induces forgetting and negative transfer.

\subsection{Additional task capacity results}
For the 27-task Cartpole domain, the empirical estimates of task capacity based on relative $\epsilon_c$-optimality with $\epsilon_c = 0.05$ (see Equation \ref{eq: capacity}). The empirical task capacity is $C_{\text{emp}} = 27$ for PPO and $C_{\text{emp}}=6.75$ for DQN. In other words, $N_{\pi}=1$ policy is sufficient for near-optimal performance for PPO and $N_{\pi}=4$ is sufficient for near-optimal performance for DQN with an error tolerance of $\epsilon_c=0.05$. The integrated task capacity (ITC) based on integrating across choices of the error tolerance yields an integrated task capacity of  $ITC = 26.7$ for PPO and $ITC= 18.7$ for DQN, implying that lower performances are quite easy to obtain regardless of the number of policies.

For the POcman domain, with a setting of $\epsilon_c=0.30$ to account for the more limited transferability in the task set, PPO has an empirical task capacity of $C_{\text{emp}}=9.0$ and an integrated task capacity $ITC=13.2$, while DRQN has a task-capacity of $C_{\text{emp}}=2$ and an integrated task capacity of $ITC=3.8$, such that even low performances are difficult to obtain. 
DRQN demonstrates negative transfer ratios between $-0.5$ and $-1.5$ while PPO has a small but positive transfer ratio between $0$ and $1$. The degrading performance visible by the near-zero performance of the 1-policy DRQN in Figure \ref{fig: policiesscorePOMDP} further suggests forgetting. DRQN and PPO demonstrate forgetting increases linearly with the number of interfering blocks: for DRQN, from $0$--$0.5$ with non-interfering task-blocks to between $-0.5$--$-1.5$ with $30$ or more interfering task-blocks; for PPO, from $0$--$0.5$ with non-interfering task-blocks and $-0.5$--$0$ with $30$ or more interfering task-blocks. All the observed forgetting effects differ by many standard errors from zero and from each other. Manipulating the number of policies does not strongly affect the forgetting ratio, except for the following DRQN conditions with more than three standard errors difference: (a) for transitions with 1 to 9 interfering task-blocks, the 1-policy DRQN has a score of $-0.25$  while other DRQN conditions range from $0.10$ to $0.50$; and (b) for 10 to 19 interfering task-blocks, the 9-policy DRQN has a score of $0$ whereas other DRQN conditions have $-0.5$.
\begin{figure}[htbp!]
\centering
\includegraphics[width=0.40\textwidth]{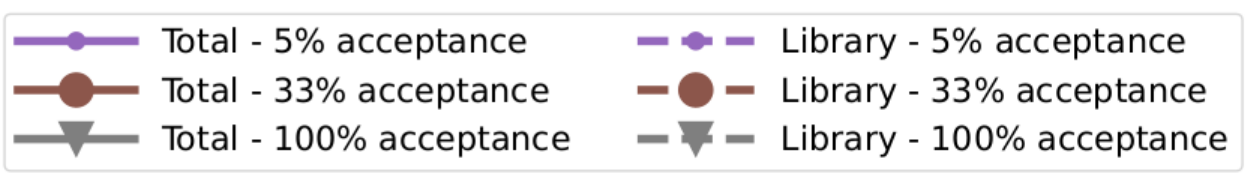}\\
\subfigure[$C=2$, $T_c=1$]
{
\includegraphics[width=0.25\textwidth]{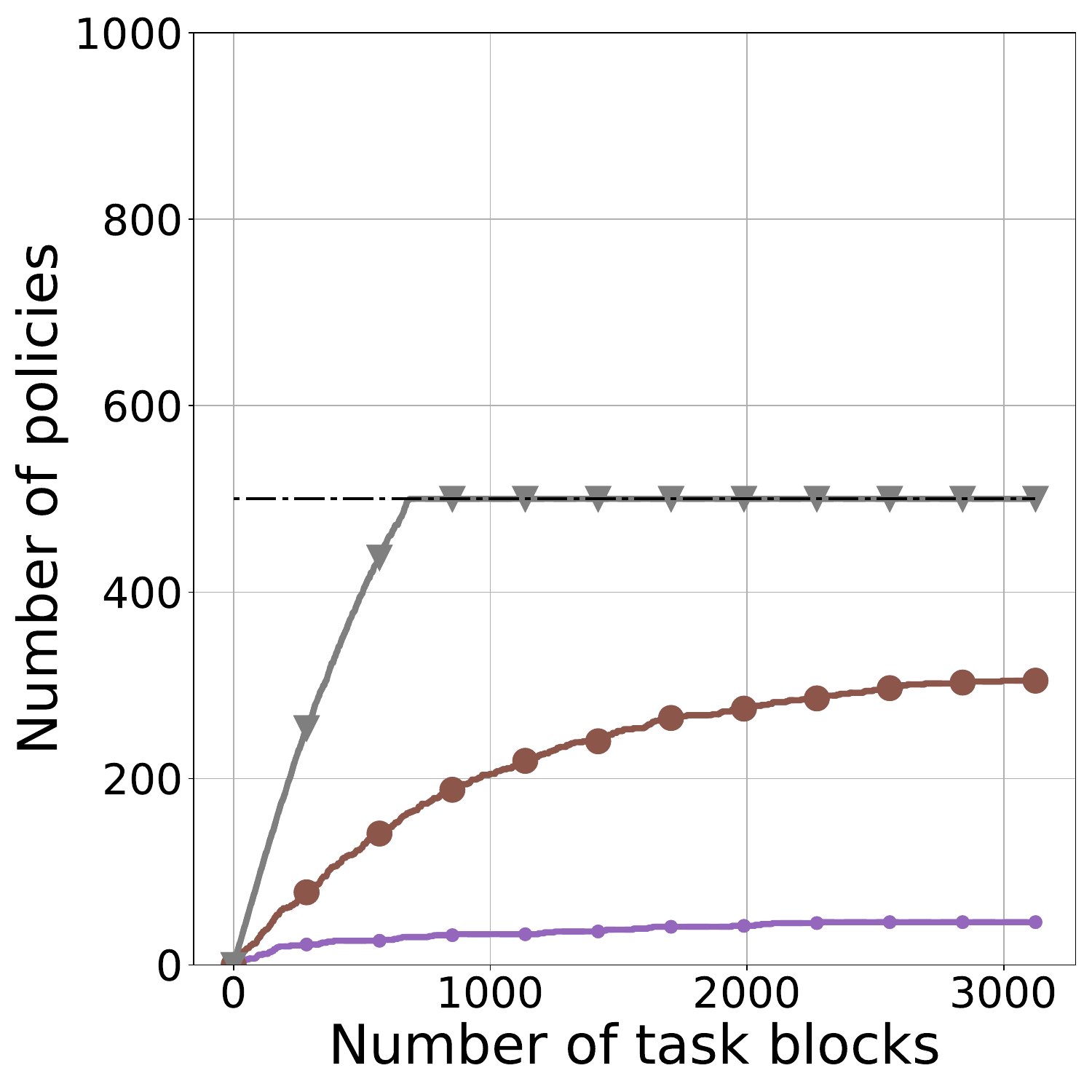}
}
\subfigure[$C=2$, $T_c=2$]
{
\includegraphics[width=0.25\textwidth]{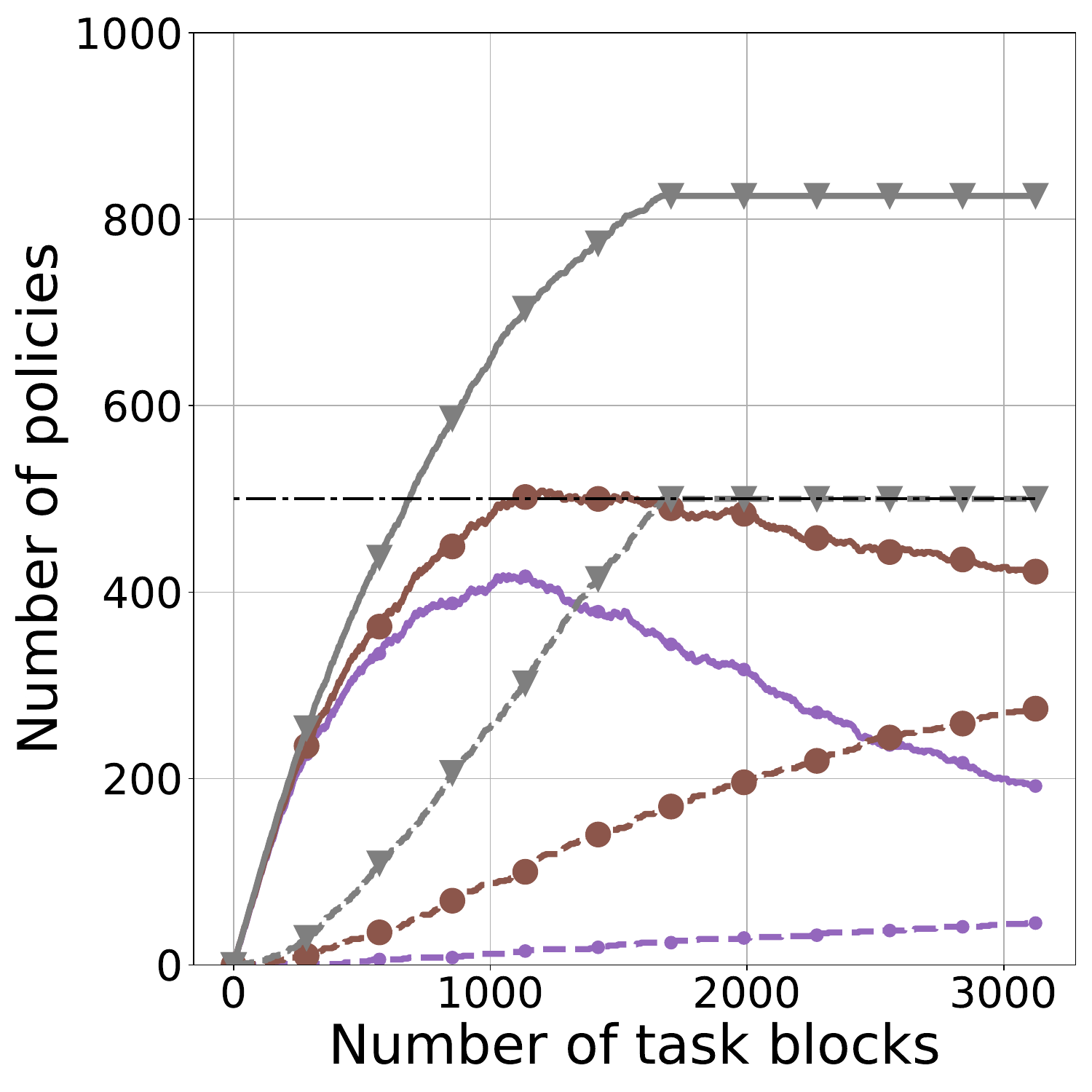}
}
\subfigure[$C=2$, $T_c=4$]
{
\includegraphics[width=0.25\textwidth]{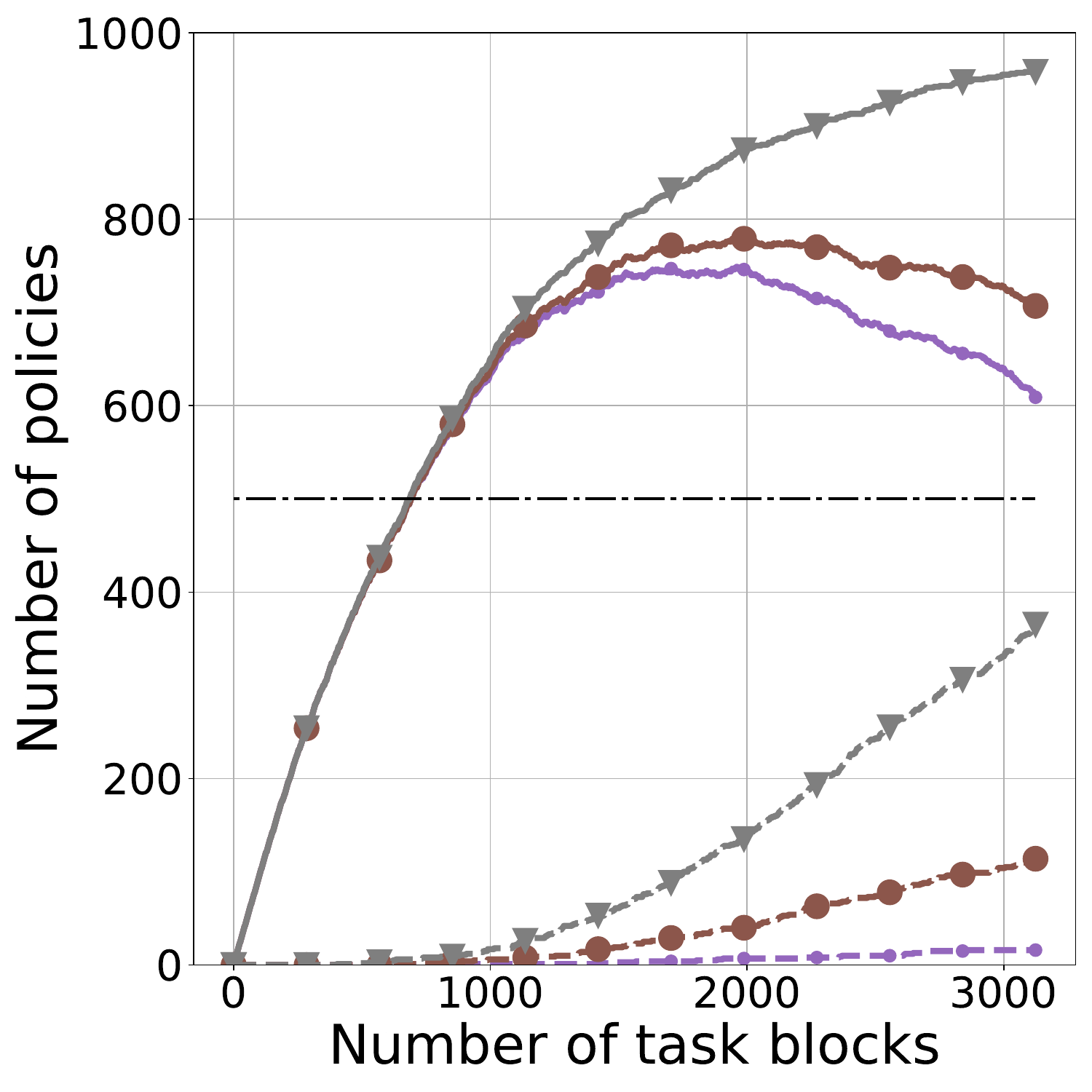}
}\\
\subfigure[$C=5$, $T_c=1$]
{
\includegraphics[width=0.25\textwidth]{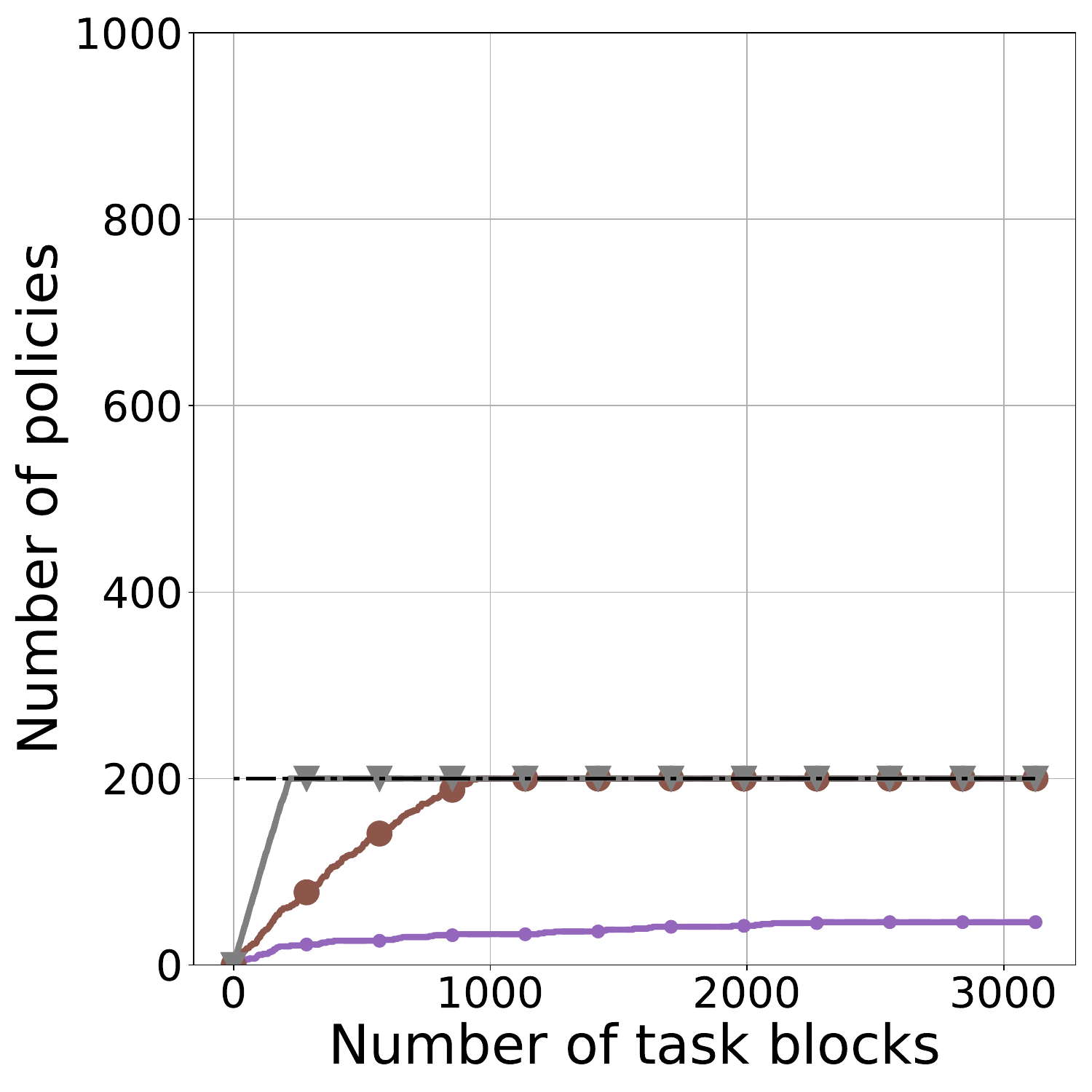}
}
\subfigure[$C=5$, $T_c=2$]
{
\includegraphics[width=0.25\textwidth]{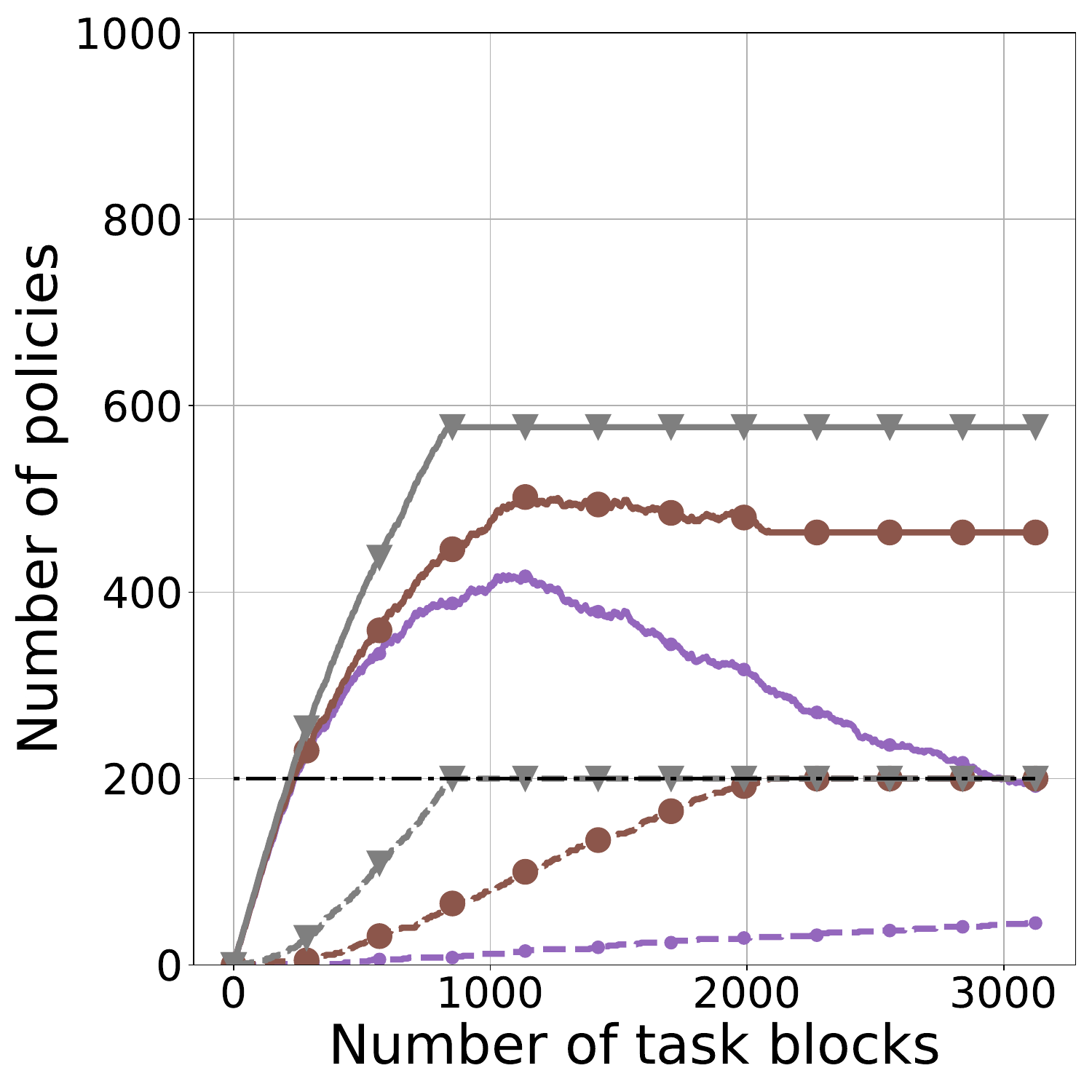}
}
\subfigure[$C=5$, $T_c=4$]
{
\includegraphics[width=0.25\textwidth]{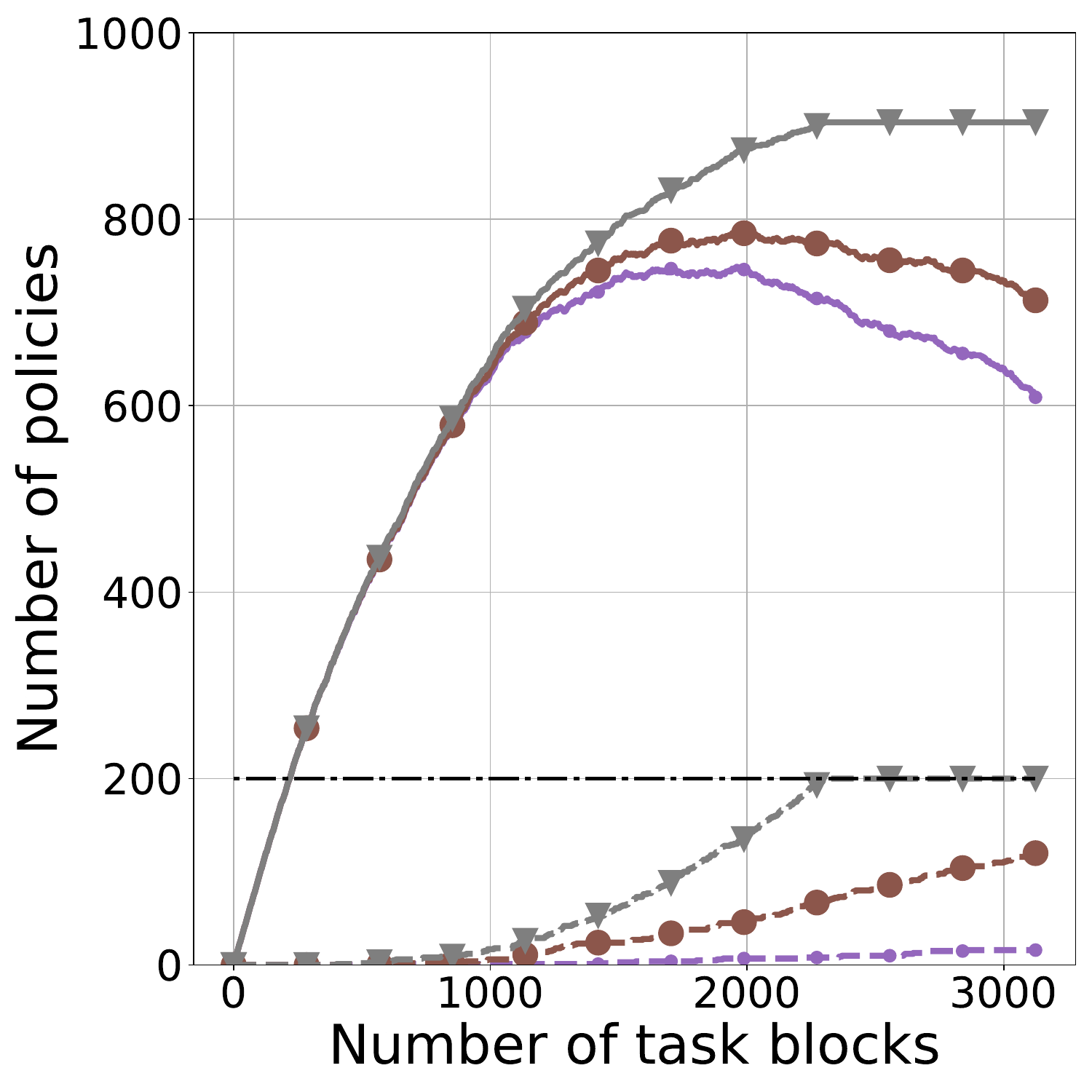}
}\\
\caption{Mean memory consumption of lifetime policy reuse (dashed line) compared to dynamically increasing the number of policies. Mean memory consumption is estimated as the number of policies averaged across 50 runs. \textbf{Library} denotes the number of policies in the policy library. \textbf{Total} denotes the total memory consumption, which combines the number of policies in the policy library with the temporary policies (i.e. those generated upon seeing a new task).  \textbf{Acceptance \%} denotes the probability of accepting a new policy into the policy library. Panels are organised based on the task capacity, $C$, and the number of task blocks until convergence, $T_c$.}
\label{fig: memory}
\end{figure}
\subsection*{Memory consumption results}
Figure~\ref{fig: memory} provides the detailed results on memory consumption.

\subsection{Understanding DQN vs PPO}
\label{sec: experience replay}
\begin{figure}[htbp!]
\centering
\subfigure[Reference setting]
{
\includegraphics[width=0.30\textwidth]{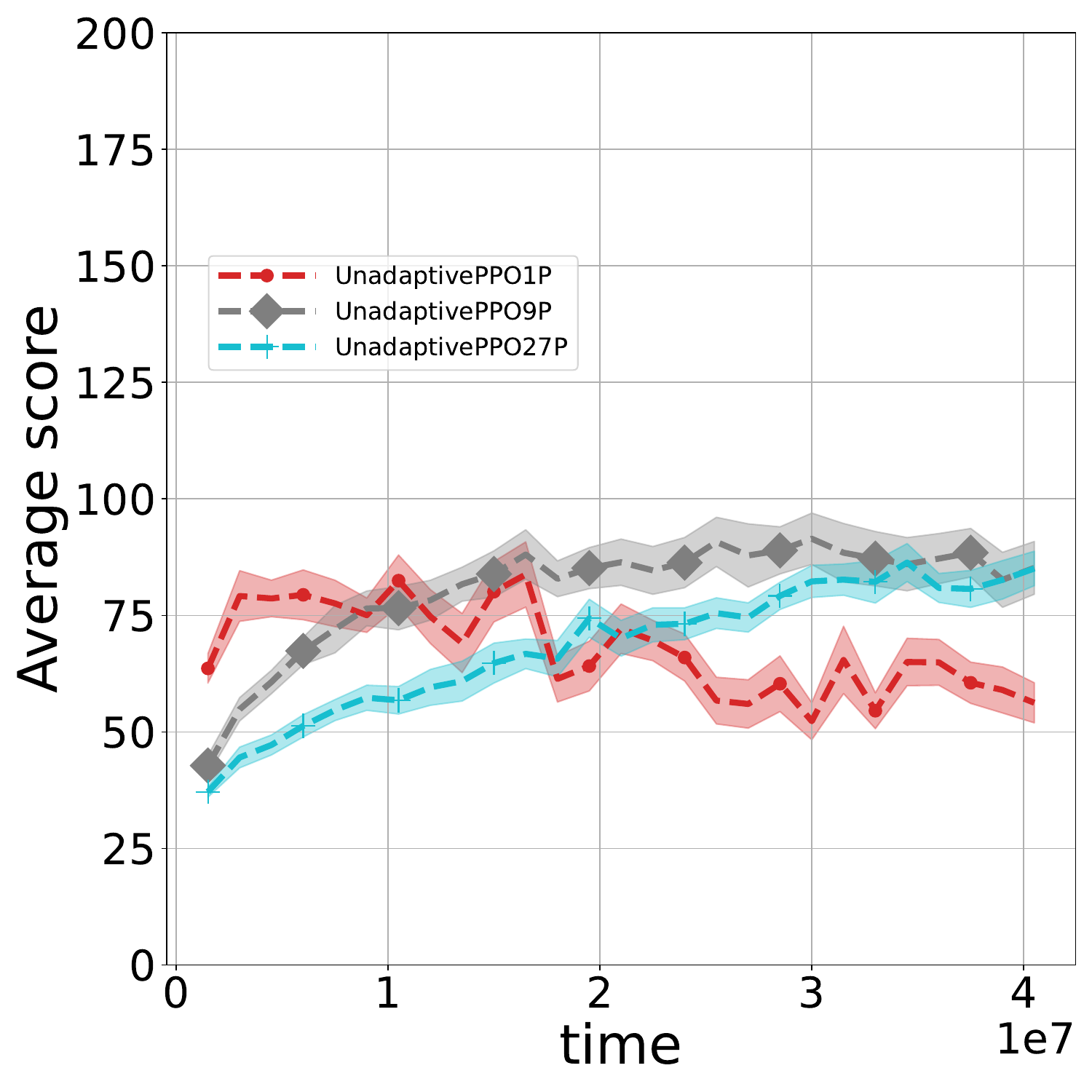}
}
\subfigure[Longer task block]
{
\includegraphics[width=0.30\textwidth]{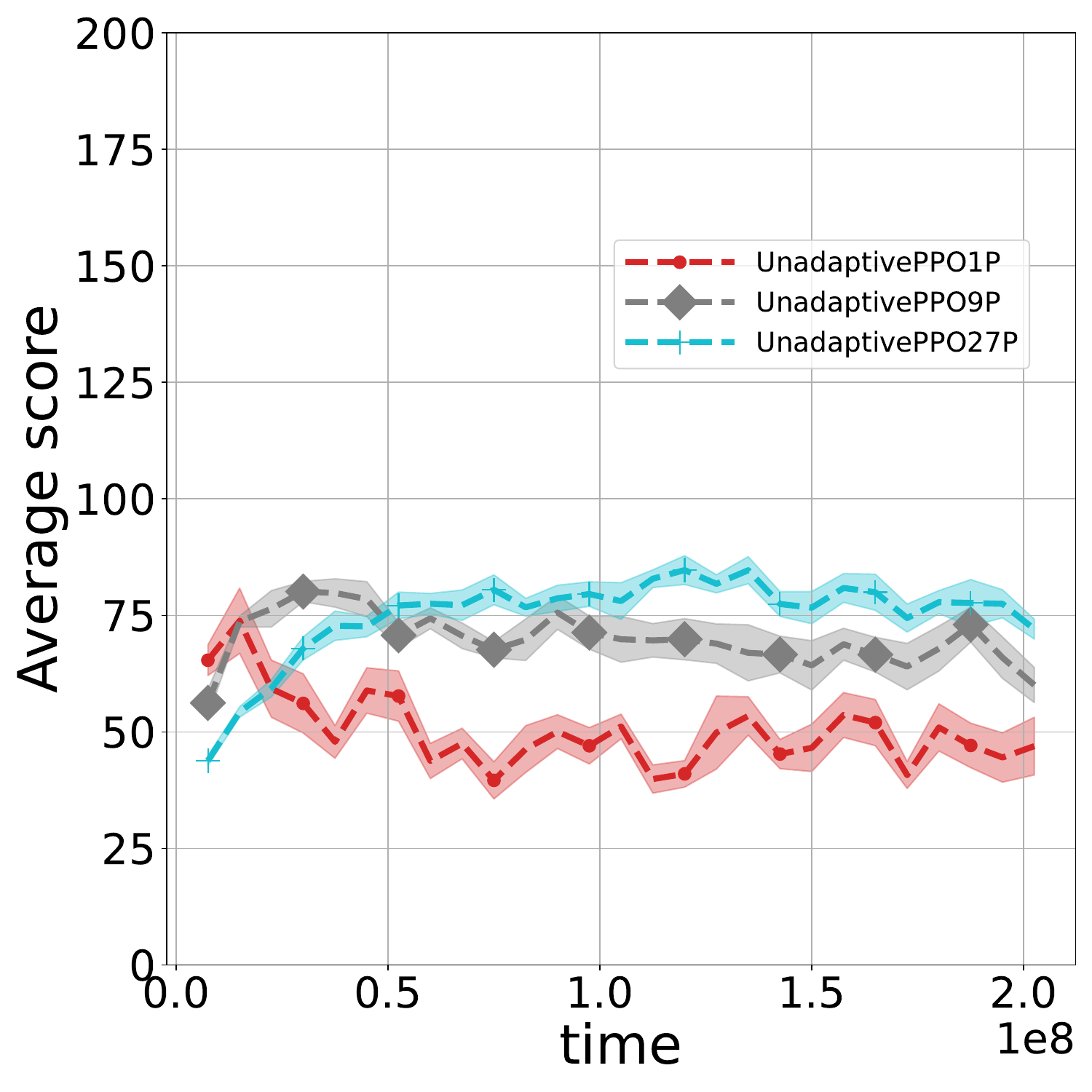}
}
\subfigure[Unbounded improvement]
{
\includegraphics[width=0.30\textwidth]{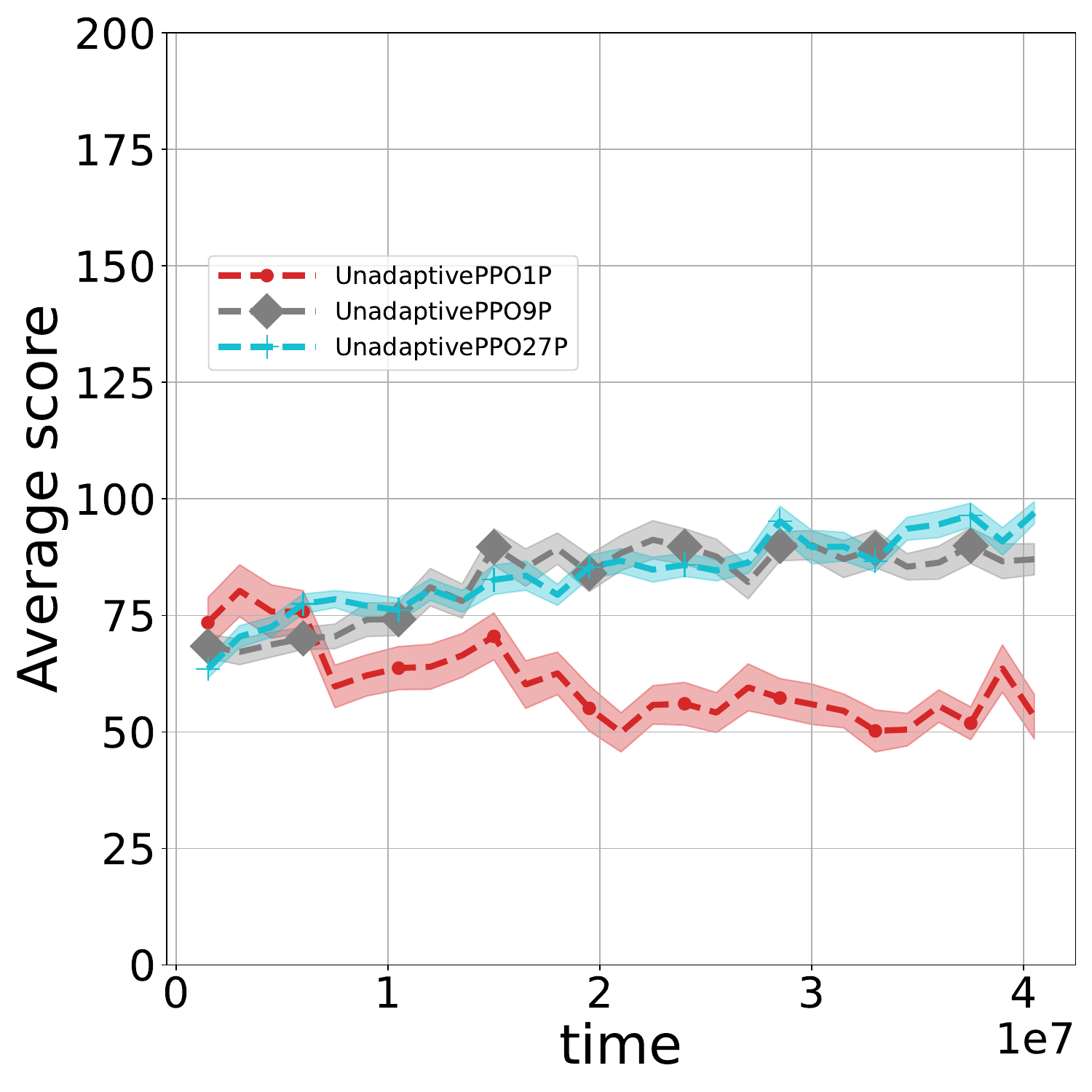}
}
\caption{Average score (Mean $\pm$ Standard Error) of PPO-based learners in the 27-task Cartpole MDP domain. \textbf{(a)} The original setting of Figure~\ref{fig: policiesscore} is replicated as a reference. \textbf{(b)} Learners are given the same ordered task sequence but task-blocks with 5 times the original duration (i.e. 300,000 rather than 60,000 time steps). To limit computational expense, results are aggregated across 10 rather than 27 sequences. \textbf{(c)} The clipped objective coefficient is modified to an extreme setting of 3.0 to more closely resemble the unbounded improvements in DQN. For both plots, the average score is averaged every 25 consecutive task-blocks.} \label{fig: PPOmonotonicimprovement}
\end{figure}

Below experiments support the performance difference analysis of DQN versus PPO. The first hypothesis is that PPO's focus on taking smaller steps for monotonic improvement increases the required number of samples but helps to maintain performance more evenly across different tasks. In the first experiment, 1-, 9-, and 27-policy PPO learners are run on a modified 27-task Cartpole MDP domain which has 300,000 instead of 60,000 time steps. Results in Figure \ref{fig: PPOmonotonicimprovement}a and b show the following: when the task capacity is exceeded, as is the case for the 1-policy learner, learning longer on a new task results in more rapidly erasing the knowledge on the tasks learned earlier in the lifetime; when the task capacity is not exceeded, as is the case for 9- and 27-policy learners, learning longer on the new task results in converging more early in the lifetime (relative to the total length of the lifetime). In the second experiment, PPO's clipped objective coefficient is modified from 0.10 to an extreme setting of 3.0, which more closely resembles the unbounded improvements allowed in DQN. Results comparing Figure \ref{fig: PPOmonotonicimprovement}a and c indicate that this makes the PPO system more similar to DQN, with a more extreme difference between the 1-policy versus the 9- and 27-policy learners; the 1-policy learner's performance dropped from 60 to just above 50 at the end of the lifetime while the 27-policy learner increased its performance from 75 to 100.  In short, the results are consistent with the hypothesis.

\begin{figure}[htbp!]
\centering
\includegraphics[width=0.4\textwidth]{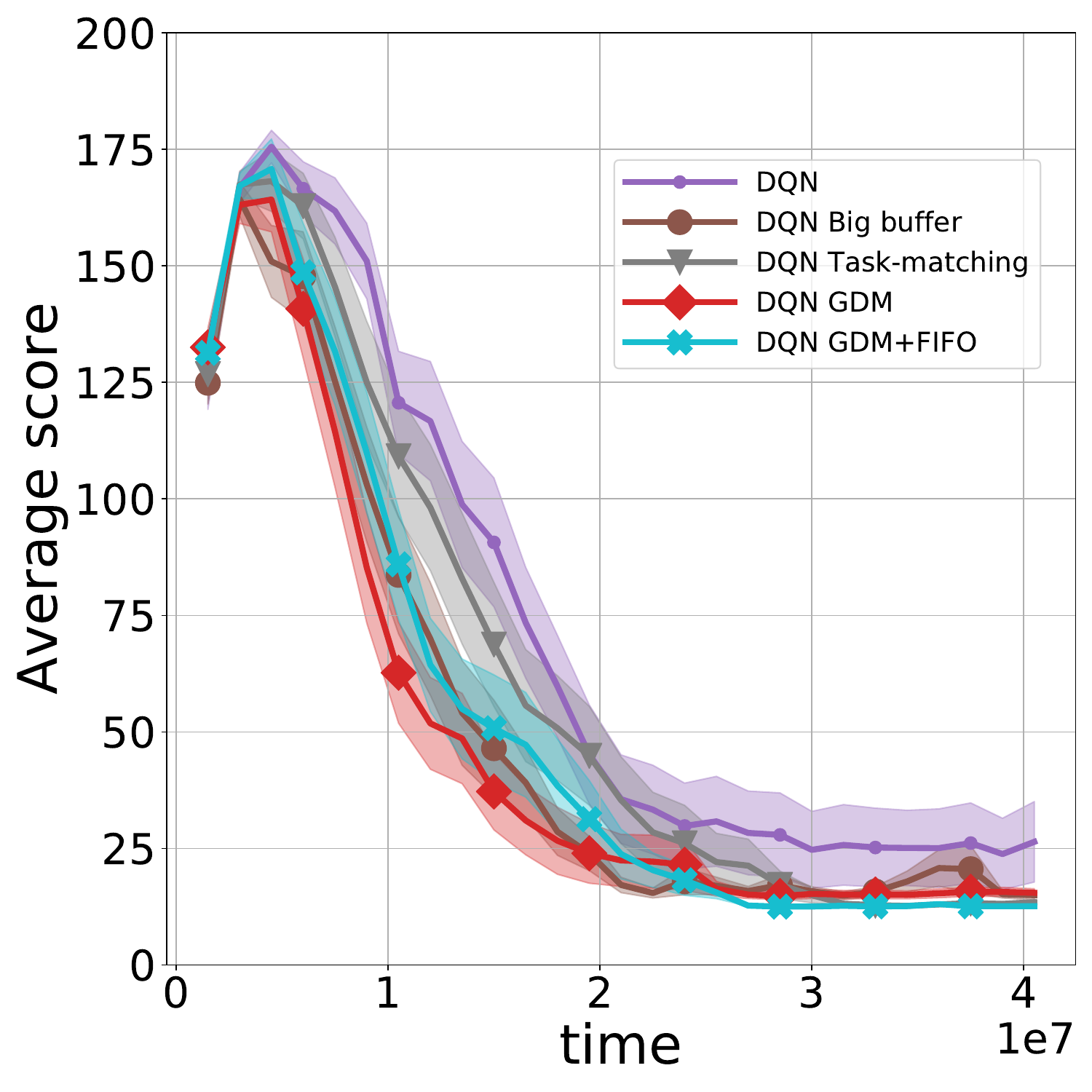}
\caption{Effect of experience replay methods on single-policy DQN performance (Mean $\pm$ Standard Error) in the 27-task Cartpole domain. GDM indicates global distribution matching; task-matching indicates replaying the current task from a task-specific buffer; and FIFO indicates First-In-First-Out buffer.} \label{fig: buffer}
\end{figure}

An alternative hypothesis is that the size and organisation of the replay buffer are the cause of DQN performance trends. If the size of the buffer is not large enough, then this may adversely affect performance. For example, the 400,000 experiences stored in the buffer allow training only on the most recent tasks and this may adversely impact being able to solve many tasks. If the organisation of the buffer does not match the overall distribution across the lifetime then DQN may overfit on the new task rather than consolidating previously seen tasks. The following experiments rule out this alternative hypothesis to further support the first hypothesis. A first experiment tests the impact of size of the buffer. The single-policy DQN is now run with a buffer size of 2,000,000 experiences, thereby spanning 33 task-blocks such that all tasks can be replayed. This does not improve performance (see Figure \ref{fig: buffer}). A second set of experiments tests the impact of the organisation of the buffer. The results are compared to the state-of-the-art selective experience replay method called global distribution matching (\textbf{GDM})  \citep{Isele2018}, which uses reservoir sampling to mimick the global distribution of experiences across the entire lifetime. In GDM, each experience in the replay buffer is associated with a random number from a normal distribution, $r \sim \mathcal{N}(0,1)$, and then only those experiences with the highest $r$ are maintained; this results in all experiences having an equal probability of $|\mathcal{B}|/t$ of being in the buffer $\mathcal{B}$ at time $t$. As observable in Figure \ref{fig: buffer}, GDM does not help to improve the performance. Adding a small additional First-In-First-Out (FIFO) buffer to the back of GDM’s buffer has no positive effect. An alternative strategy called task-matching, which allows separate buffers for each task and then replays only the experiences from the current task, also does not improve performance (see Figure \ref{fig: buffer}).

\end{appendix}

\nocite{*}
\bibliographystyle{ios1}           
\bibliography{library}
\newpage

\end{document}